\documentclass[onecolumn]{article}
\usepackage{amsmath,amssymb}
\usepackage{amsfonts,amsthm}
\usepackage{pgf,tikz}
\usepackage{graphicx}

\newtheorem{theorem}{Theorem}
\newtheorem{corollary}{Corollary}
\newtheorem{lemma}{Lemma}

\newtheorem{proposition}{Proposition}

\newtheorem{definition}{Definition}
\newtheorem{alg}{Algorithm}

\begin{document}

\title{\textbf{The Impact of Mutation Rate on the Computation Time of Evolutionary Dynamic Optimization}}
\author{Tianshi Chen$^{1,2,4}$,
Yunji Chen$^{1,4}$, Ke Tang$^2$\thanks{Corresponding author: Ke Tang (E-mail: ketang@ustc.edu.cn. Telephone: +86 551 360 0754)}, Guoliang Chen$^2$,
and Xin Yao$^{2,3}$\\ \quad\\
{\normalsize $^1$Institute of Computing Technology,}\\
{\normalsize Chinese Academy of Sciences,}\\
{\normalsize Beijing 100190, China}\\ \quad\\
{\normalsize $^2$Nature Inspired Computation and Applications Laboratory (NICAL),}\\
{\normalsize School of Computer Science and Technology,}\\
{\normalsize University of Science and Technology of China,}\\
{\normalsize Hefei, Anhui 230027, China}\\ \quad\\
{\normalsize $^3$Centre of Excellence for Research in Computational Intelligence and Applications (CERCIA),}\\
{\normalsize School of Computer Science,}\\
{\normalsize University of Birmingham,}\\
{\normalsize Edgbaston, Birmingham B15 2TT,UK}\\ \quad\\
{\normalsize $^4$Loongson Technologies Corporation Limited}\\
{\normalsize Beijing 100190, China}}

\tableofcontents

\maketitle

\begin{abstract}
Mutation has traditionally been regarded as an important operator in evolutionary algorithms. In particular, there have been many experimental studies which showed the effectiveness of adapting mutation rates for various static optimization problems. Given the perceived effectiveness of adaptive and self-adaptive mutation for static optimization problems, there have been speculations that adaptive and self-adaptive mutation can benefit dynamic optimization problems even more since adaptation and self-adaptation are capable of following a dynamic environment. However, few theoretical results are available in analyzing rigorously evolutionary algorithms for dynamic optimization problems. It is unclear when adaptive and self-adaptive mutation rates are likely to be useful for evolutionary algorithms in solving dynamic optimization problems. This paper provides the first rigorous analysis of adaptive mutation and its impact on the computation times of evolutionary algorithms in solving certain dynamic optimization problems. More specifically, for both individual-based and population-based EAs,
we have shown that any time-variable mutation rate scheme will not significantly outperform a fixed mutation rate on some dynamic optimization problem instances. The proofs also offer some insights into conditions under which any time-variable mutation scheme is unlikely to be useful and into the relationships between the problem characteristics and algorithmic features (e.g., different mutation schemes).
\end{abstract}


\section{Introduction}\label{sec:introduction}
Evolutionary Algorithms (EAs) are stochastic search algorithms, which have been used to solve many optimization problems in real-world applications \cite{liang02,TangYao07,yu03}. As one of the primary operators in the framework of EAs, the mutation operator has significant influence on the performance of an EA. During the past decades, various strategies of controlling the parameters of the mutation operator have been developed to promote the performance of EAs. Some of them were concerned with the mutation operators for binary search space (e.g., \cite{Back93,Eiben99,Fogarty89,jan00,Smith96,Stanhope98,Thierens02}), while some were dedicated to continuous search space \cite{Arnold06,Hansen01,Schwefel95,Yao99}. In this paper, we restrict our investigation to the former.

As the most commonly used mutation operator for binary search space, the so-called bitwise mutation operator flips each bit of an individual (solution) with a uniform probability $P_m$, where $P_m$ is called the mutation rate. Early investigations often employed a fixed mutation rate over the whole optimization process \cite{Eiben99}, but further studies have revealed that a time-variable mutation rate scheme might be better than a fixed mutation rate. According to B\"ack \cite{Back98}, Hinterding \emph{et al.} \cite{Hinterding97} and Thierens \cite{Thierens02}, there are three classes of \emph{time-variable mutation rate schemes}: dynamic mutation rate schemes, adaptive mutation rate schemes and self-adaptive mutation schemes. During the past decades, a large number of empirical investigations have been dedicated to show the advantages of various time-variable mutation rate schemes. Holland \cite{Holland75} proposed a time-variable mutation rate scheme for Genetic Algorithm (GA). After fourteen years, Fogarty \cite{Fogarty89} designed a number of dynamic mutation rate schemes with which the performance of GAs was significantly improved on some static optimization problems. Inspired by Evolution Strategies (ES), B\"ack \cite{Back92} proposed a self-adaptive mutation scheme for GAs for the continuous search space. Later he carried out both theoretical analysis and empirical study to show that for multimodal static optimization problems there might exist some ``Optimal Mutation Rate Schedule'' that can accelerate the search of EAs \cite{Back93}. Srinivas and Patnaik proposed a GA with adaptive mutation rate scheme and adaptive crossover scheme (Adaptive Genetic Algorithm, AGA) in \cite{Srinivas94adaptive}. They empirically showed that AGA outperformed Simple GA (SGA) on a number of static benchmark optimization problems. Smith and Fogarty \cite{Smith96} carried out empirical comparisons between an EA with self-adaptive mutation rate scheme and a number of EAs with different fixed mutation rates, and they found that the former outperforms the latter on a number of static optimization problems. Thierens \cite{Thierens02} designed two adaptive mutation rate schemes, whose advantages over a fixed mutation rate were demonstrated by experimental studies.

In comparison with empirical studies, the theoretical investigations on time-variable mutation rate schemes are much fewer. Droste, Jansen and Wegener \cite{Droste00,jan00} analyzed a $(1+1)$ EA
with a dynamic mutation rate scheme, and they proved that the expected number of generations spent by this $(1+1)$ EA on the so-called \textsc{PathToJump} problem is $O(n^2\log n)$. In contrast, with a probability that converges to $1$ (with respect to the problem size $n$), the $(1+1)$ EA with the fixed mutation rate $1/n$ will spend a super-polynomial number of generations to optimize the above \textsc{PathToJump} problem. On the basis of \cite{Droste00} and \cite{jan00}, Jansen and Wegener \cite{jan06} carried out further theoretical investigations for the above $(1+1)$ EA (with a dynamic mutation rate), and they presented a number of time complexity results on some well-known benchmark problems, such as \textsc{OneMax} \cite{Back93,Chen08a,Droste02} and \textsc{LeadingOnes} \cite{Chen08a,Rudolph98}. The investigations in \cite{jan06} show us that the specific dynamic mutation rate scheme outperforms (in terms of the runtime of $(1+1)$ EA) the fixed mutation rate $1/n$ significantly on \textsc{PathToJump}, while is slightly inferior on \textsc{OneMax} and \textsc{LeadingOnes}. As we might expect, there have also been a few studies demonstrating the potential weakness of some \emph{specific} time-variable mutation rate schemes (e.g., \cite{Rudolph01}), but it is a drop in the bucket compared with the positive results. Besides, these negative results only showed that there is no time-variable mutation rate scheme is universally good for an EA over a large variety of problems, which did not exclude the usefulness of time-variable mutation rate schemes. In other words, no general result is available so far demonstrating the complete failure of the time-variable mutation rate schemes under certain conditions.

In this paper, we prove for both individual-based and population-based EAs that time-variable mutation rate schemes (including any dynamic, adaptive and self-adaptive mutation schemes) cannot significantly outperform a well-chosen fixed mutation rate on some Dynamic Optimization Problems (DOPs) \cite{Branke01a,Jin05}. Unlike static optimization problems that maintain problem instances, stationary objective functions and constraints consistently, for DOPs, ``\emph{the objective function, the problem instance, or constraints may change over time}'' \cite{Jin05}. In most investigated DOPs, the above changes of DOPs will directly result in the movements of global optima \cite{Blackwell06dynamic,Parrott06dynamic,Yang08}. However, such movements of different DOPs can follow distinctive manners. To representatively characterize moving global optima in DOPs, we employ the so-called BDOP class in our theoretical investigations, which models the movements of global optima as random walks of binary strings in the solution space. The BDOP class is a representative DOP model for theoretical investigations, which consist of DOPs with different dynamic degrees. Concretely, the dynamic degree of a DOP belonging to the BDOP class (named BDOP in the paper) mainly depends on the value of the so-called shifting rate $\sigma\in(0,1/2]$, which is a parameter controlling the random walks of BDOPs' global optima. A larger $\sigma$ tends to lead to larger frequency and size of change for the moving global optimum.

Intuitively, the simplest way to cope with a DOP is to regard the DOP as a new static optimization problem after every change. In this way, a DOP can be solved by traditional EAs or other problem-specific algorithms designed for static optimization problems. However, this kind of approaches require vast computational resources (e.g., computation time), which is not practical for many real-world DOPs (e.g., dynamic vehicle routing problem \cite{Branke05} and online scheduling problem \cite{Chien07}). Under this circumstance, intensive investigations have been carried out to understand the behaviors of EAs on DOPs or to design better EAs for DOPs, among which the theoretical investigations are often restricted to the $(1+1)$ EA.
From a theoretical perspective, Rohlfshagen \emph{et al.} studied how the frequency and magnitude of the optimum's movement influence the performance of the $(1+1)$ EA for dynamic optimization \cite{Rohlfshagen09dynamic}. Surprisingly, on two specific problems called \textsc{Magnitude} and \textsc{Balance}, they showed that larger frequency and magnitude do not necessarily lead to a worse performance of the EA. Stanhope and Daida \cite{Stanhope99} studied the fitness dynamics of the $(1+1)$ EA on the so-called dynamic \textsc{BitMatching} problem that was frequently used in previous empirical investigations \cite{Droste03,Yang08,yuxin08}. By a similar approach, Branke and Wang \cite{Branke03} compared the $(1+1)$ and $(1,2)$ EAs on the dynamic \textsc{BitMatching} problem. Droste \cite{Droste02A,Droste03} studied the time complexity of the $(1+1)$ EA with both one-bit mutation and bitwise mutation (with mutation rate $1/n$) on the dynamic \textsc{BitMatching} problem. These results established the theoretical foundations of the field. However, theoretical investigation that aims at revealing the relationship between the adaptation of mutation rate and the performance of EAs on DOPs still lacks, though there has been some related empirical studies \cite{Rand05dynamic,Stanhope98}.

In this paper, we demonstrate theoretically the relationship between time-variable mutation schemes and the performances of both individual-based and population-based EAs on the BDOP class. In our study, we still adopt the classical time complexity criterion, the first hitting time \cite{HeYao01,HeYao02,HeYao03}, to measure the performances of EAs theoretically.
Based on the measure, an EA is said to be \emph{efficient} on a DOP, if and only if the corresponding first hitting time is polynomial in the problem size with a probability that is at least the reciprocal of a polynomial function of the problem size $n$. Otherwise, the EA is said to be \emph{inefficient} (on the DOP). According to the popular perspective that adaptation of mutation rate might be helpful, one might expect that there are some time-variable mutation scheme that can improve the performances of EAs so that they can cope with BDOPs efficiently. However, in this paper, we find that once the asymptotic order of the shifting rate $\sigma$ is larger than $\log n/n^2$ (i.e., $\sigma=\omega(\log n/n^2)$), the BDOPs will become essentially hard for $(1+1)$ EA with any time-variable mutation scheme in which the mutation rate at every generation is upper bounded by $1-1/\log n$; When the asymptotic order of the shifting rate $\sigma$ is larger than $\log n/n$ (i.e., $\sigma=\omega(\log n/n)$), the BDOPs will become essentially hard for the $(1+\lambda)$ EA\footnote{The $(1+\lambda)$ EA is a population-based EA which maintains a unique parent individual and generates $\lambda$ offspring individuals at each generation.} ($\lambda$ can be any polynomial function of $n$) with any time-variable mutation scheme in which the mutation rate at every generation is upper bounded by $1-1/\log n$. Case studies on an instance of the BDOP class called the \textsc{BitMatching}$_D$ problem further demonstrate the limitations of \emph{any} time-variable mutation scheme via concrete time complexity results on both $(1+1)$ and $(1+\lambda)$ EAs, and also show the positive impact of population on the performances of EAs.

The main contributions and their significance of this paper are summarized as follows: First, our investigation provides theoretical evidence for the view that, compared with a fixed mutation rate, adopting some time-variable mutation rate scheme is not always a significantly better choice. It is the first time that some general results for \emph{any} time-variable mutation rate scheme are given and proven rigorously. Second, this paper substantially increases the understanding of the role of mutation in the context of DOP, and the relationship between time-variable mutation rate schemes and time complexities of EAs are investigated theoretically in depth. Third, by comparing individual-based and population-based EAs on BDOPs, our investigations revealed theoretically for the first time that population may have positive impact on the performances of EAs for solving DOPs.


The rest of the paper is organized as follows: Section \ref{section:P_A} introduces the DOPs and algorithms analyzed in this paper. Section \ref{sec:11EA_hard_BDOPs} studies the impact of time-variable mutation rate schemes on performance of the $(1+1)$ EA over the BDOP class. Section \ref{sec:1lambdaEA_hard_BDOPs} further studies the impact of time-variable mutation rate schemes on performance of the $(1+\lambda)$ EA over the BDOP class. Section \ref{section:discussion} offers some discussions, interpretations and generalization for the theoretical results obtained in this paper. Section \ref{section:conclusion} concludes the whole paper.

\section{Problem and Algorithm}\label{section:P_A}
In this section, we introduce some preliminaries for this paper, including the notations of asymptotic orders, the BDOP class and EAs discussed in this paper.
\subsection{Preliminaries}
To facilitate our analysis, we first introduce some notations that are used in comparing the asymptotic growth order of functions. Let $g_1=g_1(n)$ and $g_2=g_2(n)$ be two positive functions of $n$, then \cite{jan05}:
\begin{itemize}
\item $g_1=O(g_2)$, iff $\exists n_0\in \mathbb{N}, c\in
\mathbb{R}^+$: $\forall n>n_0$, $g_1(n)\le cg_2(n)$ ($g_1$ is asymptotically bounded above by $g_2$ up to a constant factor, and the asymptotic order of $g_1$ is no larger than that of $g_2$);
\item $g_1=\Omega(g_2)$, iff $g_2=O(g_1)$ (The asymptotic
order of $g_1$ is no smaller than that of $g_2$);
\item $g_1=\Theta(g_2)$, iff $g_1=O(g_2)$ and $g_1=\Omega(g_2)$ both
hold (The asymptotic order of $g_1$ is the same to that of $g_2$);
\item $g_1=o(g_2)$, iff $\lim_{n\to \infty}g_1(n)/g_2(n)=0$ ($g_1/g_2$
approximates $0$ when $n\to\infty$, and the asymptotic order of $g_1$ is smaller than that of $g_2$);
\item $g_1=\omega(g_2)$, iff $g_2=o(g_1)$ (The asymptotic
order of $g_1$ is larger than that of $g_2$).
\end{itemize}
Moreover, to distinguish the polynomial functions from those super-polynomial functions of $n$, we utilize the following notations:
\begin{itemize}
\item $g_1\prec Poly(n) \mbox{ and } \frac{1}{g_1}\succ \frac{1}{Poly(n)}$
both hold iff $\exists c\in \mathbb{R}^+_0: g_1(n)=O(n^c)$;
\item $g_1\succ SuperPoly(n) \mbox{ and } \frac{1}{g_1}\prec \frac{1}{SuperPoly(n)}$ both hold iff $\forall c\in \mathbb{R}^+_0: g_1(n)=\omega(n^c)$.
\end{itemize}

In this paper, the above notations are further utilized in representing the asymptotic orders related to probabilities:

\begin{definition}[\textbf{Overwhelming Probability} and \textbf{Super-polynomially Small Probability}]A probability $P_1$ is regarded as ``an overwhelming probability'' if and only if there exists some super-polynomial function $g(n)$ of the problem size $n$ ($g(n)\succ SuperPoly(n)$) and a positive integer $n_0$ such that $ \forall n\ge n_0: P_1\ge 1-1/g(n)$ holds. A probability $P_2$ is regarded as ``a super-polynomially small probability'' (or ``a probability that is super-polynomially close to $0$'') if and only if there exists some super-polynomial function $g(n)$ of the problem size $n$ ($g(n)\succ SuperPoly(n)$) and a positive integer $n_0$ such that $ \forall n\ge n_0: P_2\le 1/g(n)$ holds.
\end{definition}

In the following parts of the section, we will first describe the general DOP model for our theoretical analysis. After that, we will present the concrete DOP analyzed in this paper. Finally, we introduce the $(1+1)$ EA and time-variable mutation rate scheme.

\subsection{A Theoretical DOP Model}\label{sec:DOPmodel}

In this subsection, we define a theoretical model for dynamical optimization problems on binary search space. Briefly, our DOP model characterizes a common feature of most DOPs investigated by the evolutionary computation community \cite{Jin05,Rohlfshagen09dynamic,Stanhope98,Yang08,yuxin08}, that is, the global optimum of a DOP is probabilistically changing over time. Like those DOPs intensively studied by the community, the DOP model allows uncertain events to occur at discrete time points only and accomplish without delay. In this way, the shortest duration of a stationary objective function can be guaranteed, which greatly facilitates optimization algorithms, and, in the meantime, reflects reasonable simplifications that are widely employed when solving sophisticated DOPs in practice. Taking the dynamic vehicle routing problem as an example, it is difficult to always take any real-time factor into account when searching for the optimal routing. Instead, such factors are often temporarily collected, and contribute to the modification of objective function only at some discrete time points.

Concretely, we define the DOP phase to be the time interval in which the objective function of a DOP remains stationary. For the sake of simplicity, we assume the duration of each DOP phase is $1$ such that the time point index $t$ ($t\in\mathbb{N}$, $t=0$ is the starting time point) can be utilized to distinguish one DOP phase from another. In the $t^{th}$ DOP phase, the change with respect to the objective function occurs and finishes at time point $t$. Within the DOP phase, the objective function, denoted by $f_t:\{0,1\}^n\to \mathbb{R}$, remains stationary and only has a unique global optimum in the search space. The overall goal of solving the DOP is to \textbf{maximize} the objective function in the presence of movements of the global optimum. Besides, we do not consider any constraint-based stationary objective function in our investigations, and models involving such settings will be left as our future work. By summarizing the above descriptions, we present the following definition:

\begin{definition}[DOP Model]\label{definition:DOP_model}A DOP is a maximization problem whose stationary objective function may change at any time point $t\in\mathbb{N}$. At the $t^{th}$ DOP phase, the fitness (value of objective function) with respect to a given solution $x\in\{0,1\}^n$ is given by $f_t(x)$, where $f_t:\{0,1\}^n\to \mathbb{R}$ is the stationary objective function at the $t^{th}$ DOP phase.
\end{definition}

As stated, a notable characteristic of most DOPs is that their global optima may change over time. In this paper, we characterize the movements of global optima as a kind of pure random walks in the binary solution space, which is a simple and natural way of modeling stochastic movements in theory:

\begin{definition}[Bitwise Shifting Global Optimum (BSGO)]\label{def:shifting_rate} The global optimum of a DOP is called a BSGO, if it is shifting following the rule $\forall t\in\mathbb{N}: x_{t+1}^*=\mathcal{B}_n(x_{t}^*)$, where $x_t^*$ is the global optimum at the $t^{th}$ DOP phase, $\mathcal{B}_n:\{0,1\}^n\to\{0,1\}^n$ flips every bit of the input binary string with a probability of $\sigma\in(0,1/2]$, and $\sigma$ is called the \textbf{shifting rate}. \end{definition}

The DOPs whose unique global optima are BSGOs form the BDOP class:
\begin{definition}[BDOP Class] For a DOP following the model defined in Definition \ref{definition:DOP_model}, if it only has a \textbf{unique} global optimum and the optimum is a BSGO, then the DOP is a BDOP, and we also say that the DOP belongs to the BDOP class. \end{definition}

In the rest of the paper, theoretical investigations will be carried out on the BDOP class, where the algorithms for solving such optimization problems will be introduced in the next subsection.

\subsection{Time-variable Mutation Rate
Schemes and Evolutionary Algorithms}\label{sec:EA_mutationrate} In this paper, both individual-based and population-based EAs will be employed in our theoretical analysis, and the aim is to demonstrate the impact of time-variable mutation rate schemes on different EAs when solving DOPs. Concretely, the individual-based EA studied in this paper is called the $(1+1)$ EA. At each generation, the EA maintains a unique parent individual, and the parent individual can only generate a unique offspring individual via mutation; The selection operator preserves the one with better fitness between the parent and offspring individuals (i.e., $1$ parent + $1$ offspring). A concrete description of the $(1+1)$ EA studied in this paper is given below:
\begin{alg}[\textbf{$(1+1)$
EA}]\label{definition:1plus1EA}Choose the initial individual $x_0^{(P)}$ randomly by the uniform distribution over the whole search space. Set the initial generation index $t=0$. The $t^{th}$ generation of the EA consists of the following steps:
\begin{itemize}
\item \textbf{Mutation:} Each bit of the \textbf{parent individual} $x^{(P)}_t$ is flipped with the probability of
$P_m(n,t)\in[0,1]$, where $n$ is the problem size (i.e., length of the binary string). After that, an \textbf{offspring individual} $x^{(O)}_t$ is obtained.
\item \textbf{Fitness Evaluation:} Evaluate the fitness of $x^{(P)}_t$ and $x^{(O)}_t$ based on the
stationary fitness function $f_t$ at the $t^{th}$ generation ($t^{th}$ DOP phase).
\item \textbf{Selection:} If $f_t(x^{(O)}_t)\ge f_{t}(x^{(P)}_t)$, then set $x^{(P)}_{t+1}=x^{(O)}_t$;
Otherwise, set $x^{(P)}_{t+1}=x^{(P)}_t$.
\end{itemize}
If the given stopping criterion is met after the $t^{th}$ generation, then the EA stops; Otherwise, set $t=t+1$ and a new generation begins.
\end{alg}
A significant difference between the above algorithm and the $(1+1)$ EA for static problems \cite{Droste02} is that, at every generation the former evaluates not only the fitness of the offspring but also that of the parent, while the latter only evaluates the fitness of the offspring. The reason of employing two fitness evaluations in one generation of our $(1+1)$ EA is that the fitness of the parent individual may change in response to the change of objective function. Another difference between the above EA and the traditional $(1+1)$ EA is that the former allows the mutation rate to vary over generation (i.e., the $(1+1)$ EA adopts some \emph{time-variable mutation rate scheme}), while the latter only adopts the fixed mutation rate $P_m=1/n$, where $n$ is the problem size. To make our description formal, the concrete definition of time-variable mutation rate scheme for our $(1+1)$ EA is given below:

\begin{definition}[Time-variable mutation rate scheme for $(1+1)$ EA]\label{def:scheme_11} The time-variable mutation rate scheme of the $(1+1)$ EA is a mapping $P_m:\mathbb{N}\times\mathbb{N}\to [0,1]$. Such a scheme sets the mutation rate at the $t^{th}$ generation be $P_m(n,t)$, where $n$ is the problem size.
\end{definition}

In addition to studying the time-variable mutation scheme used in the above $(1+1)$ EA, we also study the time-variable mutation schemes in the context of the following $(1+\lambda)$ EA ($\lambda$ is polynomial in $n$):
\begin{alg}[\textbf{$(1+\lambda)$
EA}]\label{definition:1plus_lambdaEA}Choose the initial individual $x_0^{(P)}$ randomly by the uniform distribution over the whole search space. Set the initial generation index $t=0$. The $t^{th}$ generation of the EA consists of the above steps:
\begin{itemize}
\item \textbf{Mutation:} The \textbf{parent individual} $x^{(P)}_t$ generates $\lambda$ ($\lambda\prec Poly(n)$) offspring individuals $x^{(1)}_t,\dots, x^{(\lambda)}_t$ independently. When generating the $\chi^{th}$ \textbf{offspring individual} $x^{(\chi)}_t$ ($\chi\in\{1,\dots,\lambda\}$), each bit of the \emph{parent individual} $x^{(P)}_t$ is flipped with the probability of $P_m(n,t,\chi)\in[0,1]$.
\item \textbf{Fitness Evaluation:} Evaluate the fitness of $x^{(P)}_t$, $x^{(1)}_t$, $\dots$, and $x^{(\lambda)}_t$ based on the
stationary fitness function $f_t$ at the $t^{th}$ generation ($t^{th}$ DOP phase).
\item \textbf{Selection:} If $\max\left\{f_{t}(x^{(1)}_t),\dots,f_{t}(x^{(\lambda)}_t)\right\}\ge f_t(x^{(P)}_t)$, then set $x^{(P)}_{t+1}=\arg \max_{\chi\in\{1,\dots, \lambda\}} f_{t}(x^{(\chi)}_t)$;
Otherwise, set $x^{(P)}_{t+1}=x^{(P)}_t$.
\end{itemize}
If the given stopping criterion is met after the $t^{th}$ generation, then the EA stops; Otherwise, set $t=t+1$ and a new generation begins.
\end{alg}
The $(1+\lambda)$ EA is a population-based EA adopting the offspring-population strategy. To be specific, unlike those population-based EAs which maintain multiple parents and offsprings at each generation, the $(1+\lambda)$ EA maintains a unique parent individual and a population of offspring individuals generated from the same parent. To guarantee that there is a unique parent at each generation, the selection operator imposes extremely high selection pressure, and preserves the one with the best fitness among the total $1+\lambda$ individuals (i.e., $1$ parent + $\lambda$ offsprings). Also, the $(1+\lambda)$ EA can be considered as a special case of the $(\lambda+\lambda)$ EA, where the selection operator always copies the selected individual for $\lambda$ times to construct the population of the next generation. In this paper, it is worth noting that the $(1+\lambda)$ EA introduced above allows different offsprings at the same generation to be generated via distinct mutation rates, which offers larger freedom for the adaptation of mutation rates than using the same mutation rate in generating all offspring individuals. Formally, the time-variable mutation rate scheme utilized by our $(1+\lambda)$ EA is defined as follows:
\begin{definition}[Time-variable mutation rate scheme for $(1+\lambda)$ EA]\label{def:scheme_1lambda} The time-variable mutation rate scheme of the $(1+\lambda)$ EA is a mapping $P_m:\mathbb{N}\times\mathbb{N}\times\{1,\dots,\lambda\}\to [0,1]$. Such a scheme sets the mutation rate for obtaining the $\chi^{th}$ offspring individual at the $t^{th}$ generation to be $P_m(n,t,\chi)$, where $n$ is the problem size.
\end{definition}
Apparently, when $\lambda=1$, the $(1+\lambda)$ EA is equivalent to the $(1+1)$ EA, and Definitions \ref{def:scheme_11} and \ref{def:scheme_1lambda} are equivalent to each other.

In our theoretical investigations, we assume that \emph{there are always enough parallel computational resources available to support a polynomial number of simultaneous fitness evaluations}, which is critical to make the $(1+\lambda)$ EA valid for solving DOPs. Moreover, we also assume that \emph{the $t^{th}$ ($t\in\mathbb{N}$) generation of every EA starts at the beginning of the $t^{th}$ DOP phase, and finish at the end of the $t^{th}$ DOP phase}, which is important for theoretical analysis since it can avoid the degenerate cases where the objective function changes when an EA is carrying out fitness evaluations.
\subsection{Measure of Time
Complexity}\label{section:measure} \label{subsection:algor_measure} So far we have introduced the problem and algorithm investigated in this paper. In this subsection, we present the measure of performances of EAs, which is indispensable to our theoretical studies. Traditionally, the performance of an EA on a static optimization problem can be measured by the first hitting time \cite{HeYao01,HeYao02,HeYao03,ZhouHeNie08SAT,YuZhou08}. This concept measures the number of generations needed by an EA to find the optimum of a static optimization problem, which can be generalized to facilitate theoretical analysis evolutionary dynamic optimization.
For $(1+1)$ and $(1+\lambda)$ EAs on DOPs, we formally define the first hitting time as follows:
\begin{definition}[First hitting time]\label{definition:fht}
On a DOP $\{f_t: t\in\mathbb{N}\}$, the first hitting time of a $(1+\lambda)$ EA ($\lambda\in\mathbb{N}^+$ is polynomial in $n$), denoted by $\tau$, is defined as follows:
\begin{eqnarray}\label{equation:first hitting time_re_eva}
&\tau:=\min\left\{t\ge 0;\left(x^{(P)}_t=x^*_t\right)\vee\left(x^{(1)}_t=x^*_t\right)\vee\dots\vee\left(x^{(\lambda)}_t=x^*_t\right)\right\},
\end{eqnarray}
where $x^{(P)}_t$ is the parent individual at the $t^{th}$ generation, and $x^{(1)}_t,\dots,x^{(\lambda)}_t$ are the $\lambda$ offspring individuals generated by $x^{(P)}_t$. Setting $\lambda=1$ in the above definition and replacing the notation $x^{(1)}_t$ with $x^{(O)}_t$ yield the first hitting time of the $(1+1)$ EA.
\end{definition}

Based on the problem and algorithms introduced in this section, in the rest of the paper we study time-variable mutation rate schemes in terms of first hitting times of EAs.

\section{$(1+1)$ EA with Time-Variable Mutation Schemes}\label{sec:11EA_hard_BDOPs}
During the past decade, a number of studies have been dedicated to prove or validate that some specific time-variable mutation rate schemes are helpful to improve performances of EAs, although it is unclear whether this is generally true. In this section, we present several theoretical results concerning the performance of the $(1+1)$ EA with different time-variable mutation rate schemes on BDOPs. In the first subsection, we offer a general result, demonstrating that the BDOPs with shifting rate $\sigma=\omega(\log n/n^2)$ cannot be solved efficiently by the $(1+1)$ EA with any time-variable mutation rate scheme satisfying $\forall t\in \mathbb{N}: P_m(n,t)\in[0,1-1/\log n]$. In the second subsection, we generalize the above result to a specific BDOP called the \textsc{BitMatching}$_D$ problem, and show that the $(1+1)$ EA with {\bf{\em any}} time-variable mutation rate scheme (i.e., $\forall t\in \mathbb{N}: P_m(n,t)\in[0,1]$) fails to optimize the \textsc{BitMatching}$_D$ problem efficiently when the shifting rate is $\sigma=\omega(\log n/n^2)$.

\subsection{A General Result}\label{sec:small_sigma}
The BDOPs studied in this subsection are with shifting rate $\sigma=\omega(\log n/n^2)$, which implies that the shifting rate of the BDOPs satisfies that $\lim_{n\to \infty}(\log n/n^2)/\sigma=0$. The average movement of global opttimum at each DOP phase (which is equivalent to a generation of EA, as mentioned), measured by the Hamming distance, is larger than $\Theta(\log n/n)$. This includes the case that at each DOP phase the global optimum changes by less than a bit on average. Intuitively, such a small movement speed of global optimum seems not to seriously affect the optimization process, and the $(1+1)$ EA may probably cope with such situations by switching to appropriate mutation rates. In this subsection it is discovered by Theorem \ref{theorem:generalization_low_sigma} that even such small movements will have significant influence on the first hitting time of the $(1+1)$ EA with different time-variable mutation schemes. The main result in this subsection is formally presented as follows:

\begin{theorem}\label{theorem:generalization_low_sigma} Given any BDOP with shifting rate $\sigma=\omega(\log n/n^2)$ and any time-variable mutation rate scheme $\left\{P_m(n,t)\in[0,1-1/\log n]:t\in\mathbb{N}\right\}$, the first hitting time of the $(1+1)$ EA is super-polynomial with an overwhelming probability.
 \end{theorem}

The above theorem holds when $\forall t\in\mathbb{N}:P_m(n,t)\in[0,1-1/\log n]$, which means that the largest mutation rate that the EA can switch to is $1-1/\log n$. Within the interval $[0,1-1/\log n]$, the $(1+1)$ EA can adjust its time-variable mutation rate \emph{\textbf{freely}} following an \emph{\textbf{oracle}}, i.e., at each generation the EA can even choose the mutation rate best suiting the current situation. The theorem can be proven given the following lemma first:

\begin{lemma}\label{lemma:_generalization_large_sigma} Given any BDOP with shifting rate $\sigma=\omega(\log n/n)$ and any time-variable mutation rate scheme $\left\{P_m(n,t)\in[0,1-1/\log n]:t\in\mathbb{N}\right\}$, the first hitting time of the $(1+1)$ EA is super-polynomial with an overwhelming probability. \end{lemma}

\textbf{\emph{Proof Idea of Lemma \ref{lemma:_generalization_large_sigma}.}} Generally speaking, when proving Lemma \ref{lemma:_generalization_large_sigma}, we should keep in mind that the $(1+1)$ EA can adjust its time-variable mutation rate following an oracle. To be specific, we must carry out a best-case analysis so as to bound all potential behaviors of catching the global optimum of a BDOP. Given a solution $x$, define the number of matching bits of the solution (to the global optimum) to be the problem size $n$ minus the Hamming distance between the solution and the current global optimum $x_t^*$ (i.e., $n-H(x,x_t^*)$). A formal definition of Hamming distance $H(\cdot,\cdot)$ mentioned above is given as below:
\begin{definition}[Hamming distance] The Hamming distance between two solutions $x=(s_1,\dots,s_n)$ and $y=(s'_1,\dots,s'_n)$ ($x,y\in\{0,1\}^n$) is given by $H(x,y):=\sum_{i=1}^{n}|s_i - s'_i|$. \end{definition}
Let $x_t^{(P)}$ and $x_t^{(O)}$ be the parent and offspring individuals of $t^{th}$ generation ($t\in\mathbb{N}$) of the $(1+1)$ EA, respectively. Let $x_t$ be the one with higher fitness between $x_t^{(P)}$ and $x_t^{(O)}$ (at the $t^{th}$ generation):
\begin{eqnarray*}
\forall t\in\mathbb{N}: x_{t}:=\left\{
\begin{array}{llll}
 x_{t}^{(P)}, &\mbox{if}\quad f_{t}(x_t^{(P)})> f_{t}(x_t^{(O)});\\
 x_{t}^{(O)}, &\mbox{if}\quad f_{t}(x_t^{(P)})\le  f_{t}(x_t^{(O)}).
 \end{array}
\right.
\end{eqnarray*}
Let $N_t^{(P)}:=n-H\left(x_t^{(P)},x_t^*\right)$, $N_t^{(O)}:=n-H\left(x_t^{(O)},x_t^*\right)$, and
\begin{eqnarray}\label{equation:N_t}
N_t:=n-H(x_t,x_t^*).
\end{eqnarray}
It follows from the above definitions that
\begin{eqnarray}\label{equation:N_t_range}
\max\left\{N_t^{(P)},N_t^{(O)}\right\}\ge N_t\ge \min\left\{N_t^{(P)},N_t^{(O)}\right\}.
\end{eqnarray}
The definition of $N_t$ ($t\in\mathbb{N}$) also implies the overall mapping that maps $N_t$ to $N_{t+1}$, is determined by not only the EA, but also the optimum-shifting in BDOP. The reason is that, the $(1+1)$ EA maps one solution to another only, while the optimum-shifting in BDOP can be considered as a mapping that describes the movements of the global optimum. The above two factors, together with $N_t$, determine the Hamming distance between a solution and the current global optimum at the $(t+1)^{th}$ generation.

\begin{figure} [htbg]
\centering
\includegraphics[width=0.6\textwidth]{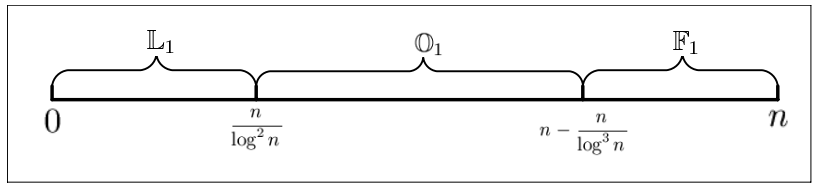}
\caption{Interval Decomposition for Lemma \ref{lemma:_generalization_large_sigma}.} \label{fig:large_sigma}
\end{figure}

To explain the proof idea, we define a number axis (real number line) with respect to the number of matching bits (to the current global optimum), ranging from $0$ to $n$. As illustrated in Fig. \ref{fig:large_sigma}, we further define several intervals on the axis:
 \begin{definition}The First Forbidden Interval and
First LongJump Interval are defined as follows:

\begin{enumerate}
\item[1.] \textbf{First Forbidden Interval: }The First Forbidden Interval is the interval $\mathbb{F}_1:=[n-n/\log^3 n,n]$, where $n$ is the problem size.

\item[2.] \textbf{First LongJump Interval:} The First LongJump Interval is the interval $\mathbb{L}_1:=[0,n/\log^2 n]$.
\end{enumerate}

\end{definition}
In addition, we let $\mathbb{O}_1:=(n/\log^2n,n-n/\log^3 n)$ such that $\mathbb{L}_1\cup \mathbb{O}_1\cup \mathbb{F}_1 = [0,n]$. The above decomposition is illustrated in Fig. \ref{fig:large_sigma}. The purpose of defining the three intervals is to quantitatively characterize three general states of the $(1+1)$ EA respectively. Intuitively speaking, for the First Forbidden Interval, $N_t\in\mathbb{F}_1$ demonstrates the situation that the better one between the parent and offspring at the $t^{th}$ generation is very close to the moving global optimum of a BDOP. Similarly, belonging to the First LongJump interval ($N_t\in\mathbb{L}_1$) indicates the situation that the solution found by the EA is extremely far away from the global optimum, but it can reach the First Forbidden Interval $\mathbb{F}_1$ by an extremely long jump resulted from a very large mutation rate (i.e., $1-1/\log n$). Finally, belonging to the interval $\mathbb{O}_1$ represents the situation that the EA is still far from finding the optimum, no matter what concrete mutation rate the EA adopts.

In our best-case analysis for proving Lemma \ref{lemma:_generalization_large_sigma}, as long as the EA has found solutions belonging to the intervals $\mathbb{F}_1$ and $\mathbb{L}_1$, we optimistically consider that the EA has reached the global optimum. By noting an important fact in Definition \ref{definition:1plus1EA} that the EA starts in $\mathbb{O}_1$ (i.e., the number of matching bits of the initial solution belongs to $\mathbb{O}_1$) with an overwhelming probability, to prove Lemma \ref{lemma:_generalization_large_sigma}, we only need to prove that the transition from $\mathbb{O}_1$ to $\mathbb{F}_1$ and $\mathbb{L}_1$ is very unlikely to happen.

Formally, we need to prove the following propositions when proving Lemma \ref{lemma:_generalization_large_sigma}:
\begin{enumerate}
\item[]\textbf{Proposition A1.1:} $\mathbb{P}(N_{0}\in \mathbb{F}_1\cup \mathbb{L}_1)\prec 1/SuperPoly(n)$.
\item[]\textbf{Proposition A1.2:}
$\forall t\in\mathbb{N}^+: \mathbb{P}(N_{t}^{(P)}\in\mathbb{F}_1\cup \mathbb{L}_1\mid N_{t-1}\in \mathbb{O}_1)\prec 1/SuperPoly(n)$.
\item[]\textbf{Proposition A1.3}:
$\forall t\in\mathbb{N}^+: \mathbb{P}(N_{t}^{(P)}\in\mathbb{O}_1, N_{t}^{(O)}\in\mathbb{F}_1\cup \mathbb{L}_1\mid N_{t-1}\in \mathbb{O}_1)\prec 1/SuperPoly(n)$.
\end{enumerate}
For the detailed proof following the above sketch, interested readers can refer to the appendix of the paper. $\hfill\square$

From Lemma \ref{lemma:_generalization_large_sigma}, when proving Theorem \ref{theorem:generalization_low_sigma} we only need to cope with smaller shifting rates satisfying the conditions $\sigma=O(\log n/n)$ and $\sigma=\omega(\log n/n^2)$. Given the condition $\sigma=O(\log n/n)$, we let $\sigma\le \delta\log n/n$ in the proof of Lemma \ref{lemma:_generalization_large_sigma}, where $\delta$ is an arbitrary positive constant. Having identified the above conditions, we define $\gamma=\gamma(n,\sigma)$ as follows:
\begin{eqnarray*}
\gamma=\gamma(n,\sigma):=\min\bigg\{\frac{n}{\log n},\sigma\cdot\frac{n^{2}}{\log n}\bigg\}.
\end{eqnarray*}
Further, let $G=G(n,\sigma)$ be defined by
\begin{eqnarray}\label{equation:def_G}
G=G(n,\sigma):=\gamma^{4/7}\log n.
\end{eqnarray}
The purpose of introducing the notations $\gamma$ and $G$ is to further define subintervals for the interval $[0,n]$ with respect to the number of matching bits of a solution to the current optimum, in addition to $\mathbb{F}_1$, $\mathbb{L}_1$ and $\mathbb{O}_1$. Concretely, we consider the following new intervals:
\begin{definition}\label{definition:decomposition_I}Some intervals are introduced below:
\begin{enumerate}

\item[1.] \textbf{Second Forbidden Interval:} The Second Forbidden Interval is the interval $\mathbb{F}_2:=[n-G,n]$, where $n$ is the problem size.

\item[2.] \textbf{Primary Adjacent Intermediate Interval:} The Primary Adjacent Intermediate Interval is the interval
$\mathbb{A}_1:=[n-2G,n-G)$.

\item[3.] \textbf{Secondary Adjacent Intermediate Interval:} The Secondary Adjacent Intermediate Interval is the interval
$\mathbb{A}_2:=[n-3G,n-2G)$.

\item[4.] \textbf{Second LongJump Interval:} The Second LongJump Interval is the interval $\mathbb{L}_2:=[0,4G]$, where $n$ is the problem size.

\item[5.] \textbf{Primary Remote Intermediate Interval:} The Primary Remote Intermediate Interval is the interval
$\mathbb{B}_1:=(4G,5G]$.

\item[6.] \textbf{Secondary Remote Intermediate Interval:} The Secondary Remote Intermediate Interval is the interval
$\mathbb{B}_2:=(5G,6G]$.
\end{enumerate}

\end{definition}
\begin{figure} [htbg]
\centering
\includegraphics[width=0.6\textwidth]{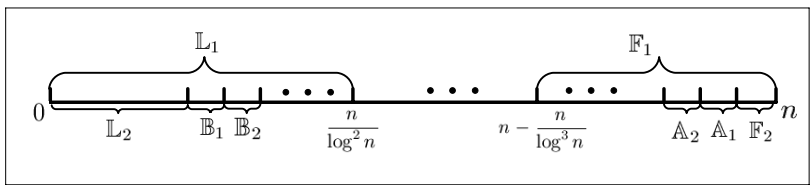}
\caption{Interval Decomposition for Theorem \ref{theorem:generalization_low_sigma}.} \label{fig:the2}
\end{figure}
The above intervals, along with $\mathbb{F}_1$ and $\mathbb{L}_1$, are illustrated in Fig. \ref{fig:the2}. Generally speaking, the above interval decomposition inherits and generalizes the intuitive idea utilized in the proof of Lemma \ref{lemma:_generalization_large_sigma}. Similar to $\mathbb{F}_1$ and $\mathbb{L}_1$ in the proof of Lemma \ref{lemma:_generalization_large_sigma}, the Second Forbidden Interval $\mathbb{F}_2$ is used to characterize the state that a solution found by the EA is very close to the current global optimum; The Second LongJump Interval $\mathbb{L}_2$ is used to characterize the state that a solution found by the EA is extremely far away from the current global optimum, which is very likely to further reach the optimum by employing an extremely large mutation rate (e.g., $1-1/\log n$). However, due to different values of shifting rates ($\omega(\log n/n)$ holds in Lemma \ref{lemma:_generalization_large_sigma} while $\omega(\log n/n^2)$ and $O(\log n/n)$ hold in Theorem \ref{theorem:generalization_low_sigma}), the concrete sizes of $\mathbb{F}_2$ and $\mathbb{L}_2$ are different from those of $\mathbb{F}_1$ and $\mathbb{L}_1$ respectively. In addition to the configurations of ``forbidden'' and ``long-jump'' intervals, here we employ some extra intervals which serve as intermediate intervals for reaching the ``forbidden'' and ``long-jump'' intervals, $\mathbb{F}_2$ and $\mathbb{L}_2$. Concretely, at the very beginning of optimization, the solution found by the EA belongs to the interval $\mathbb{O}_1$ defined in the last subsection. To find a solution in $\mathbb{F}_2$ by employing a small mutation rate, the EA must find some solution in the Secondary Adjacent Intermediate Interval $\mathbb{A}_2$ first. Afterwards, the EA needs to travel through the Primary Adjacent Intermediate Interval $\mathbb{A}_1$, i.e., try to find solutions that exceeds $\mathbb{A}_1$ and finally reach $\mathbb{F}_2$. On the other hand, to find a solution in $\mathbb{F}_2$ by employing an extremely large mutation rate, the EA has to find some solution in the ``long-jump'' interval $\mathbb{L}_2$ first. Nevertheless, to reach $\mathbb{L}_2$, the EA has to find some solution in the Secondary Remote Intermediate Interval $\mathbb{B}_2$ first, and afterwards travel through the Primary Remote Intermediate Interval $\mathbb{B}_1$ so as to reach $\mathbb{L}_2$. In our best-case analysis, once the EA has found solutions belonging to the intervals $\mathbb{F}_2$ and $\mathbb{L}_2$, we optimistically consider that the EA has reached the global optimum.

So far we have briefly introduced the interval decomposition utilized in the proof of Theorem \ref{theorem:generalization_low_sigma}. Next, we provide the detailed sketch for proving Theorem \ref{theorem:generalization_low_sigma}.
\begin{enumerate}
\item[]\textbf{Proposition B\ref{theorem:generalization_low_sigma}.1}:
$\mathbb{P}(N_0\in \mathbb{A}_2\cup\mathbb{A}_1\cup\mathbb{F}_2\cup\mathbb{B}_2\cup\mathbb{B}_1\cup\mathbb{L}_2)\prec 1/SuperPoly(n)$.

\item[]\textbf{Proposition B\ref{theorem:generalization_low_sigma}.2}:
 $\forall t$ that satisfies $t\in\mathbb{N}^+$ and
$P_m(n,t)<\gamma^{1/14}\log n/n$,
\begin{eqnarray*}
\mathbb{P}\bigg(|N_t-N_{t-1}|<\delta\gamma^{1/7}\log n\mid P_m(n,t)<\frac{\gamma^{1/14}\log n}{n},\sigma<\frac{\delta\log n}{n}\bigg)\succ 1-\frac{1}{SuperPoly(n)}
\end{eqnarray*}
holds.

\item[]\textbf{Proposition B\ref{theorem:generalization_low_sigma}.3}:
\begin{enumerate}
\item[](B\ref{theorem:generalization_low_sigma}.3a) To reach $\mathbb{F}_2$ within a polynomial number of generations, with an
overwhelming probability the EA must reach $\mathbb{A}_2$ or $\mathbb{L}_2$ first.

\item[](B\ref{theorem:generalization_low_sigma}.3b) To reach $\mathbb{L}_2$ within a polynomial number of generations, with an overwhelming
probability the EA must reach $\mathbb{B}_2$ first.

\end{enumerate}

\item[]\textbf{Proposition B\ref{theorem:generalization_low_sigma}.4}:
\begin{enumerate}
\item[](B\ref{theorem:generalization_low_sigma}.4a) $\forall t$ that satisfies $t\in\mathbb{N}^+$ and $N_{t-1}\in\mathbb{A}_2\cup\mathbb{A}_1\cup\mathbb{F}_2$,
\begin{eqnarray*}
\mathbb{P}\bigg(N_t-N_{t-1}<\gamma^{1/7}\mid N_{t-1}\in\mathbb{A}_2\cup\mathbb{A}_1\cup\mathbb{F}_2,\sigma<\frac{\delta\log n}{n}\bigg) \succ 1-\frac{1}{SuperPoly(n)}
\end{eqnarray*}
holds.

\item[](B\ref{theorem:generalization_low_sigma}.4b) $\forall t$ that satisfies $t\in\mathbb{N}^+$ and $N_{t-1}\in\mathbb{B}_2\cup\mathbb{B}_1\cup\mathbb{L}_2$,
\begin{eqnarray*}
\mathbb{P}\bigg(N_{t-1}-N_t<\gamma^{1/7}\mid N_{t-1}\in\mathbb{B}_2\cup\mathbb{B}_1\cup\mathbb{L}_2,\sigma<\frac{\delta\log n}{n}\bigg)\succ 1-\frac{1}{SuperPoly(n)}
\end{eqnarray*}
holds.
\end{enumerate}
\item[]\textbf{Proposition B\ref{theorem:generalization_low_sigma}.5}: To travel through $\mathbb{A}_1$,
the EA will spend a super-polynomial number of generations with an overwhelming probability.
\end{enumerate}
One may note that Propositions B\ref{theorem:generalization_low_sigma}.1, B\ref{theorem:generalization_low_sigma}.3 and B\ref{theorem:generalization_low_sigma}.5 are in response to our discussions in the previous paragraph. In addition, Propositions B\ref{theorem:generalization_low_sigma}.2 and B\ref{theorem:generalization_low_sigma}.4 are both formal propositions concerning the gain of the better solution (between the parent and offspring at the $t^{th}$ generation) found by the EA, compared with that of the previous generation. However, the two propositions tackle different conditions concerning the shifting rate of BDOP (i.e., $\sigma$), mutation rate ($P_m(n,t)$) and so on, which are crucial for the formal proof of Propositions B\ref{theorem:generalization_low_sigma}.3 and B\ref{theorem:generalization_low_sigma}.5. A detailed proof of Theorem \ref{theorem:generalization_low_sigma} is in the appendix of the paper.

Theorem \ref{theorem:generalization_low_sigma} presents a general result showing that any BDOP with shifting rate $\omega(\log n/n^2)$ is essentially hard to the $(1+1)$ EA with any time-variable mutation rate scheme $\left\{P_m(n,t):t\in\mathbb{N}\right\}$ satisfying $\forall t\in\mathbb{N}: P_m(n,t)\in[0,1-1/\log n]$. These essentially hard BDOPs, which characterize the movements of global optimum as random walks in the binary solution space, demonstrate the potential weakness of time-variable mutation rate scheme on DOPs with moving optimum. However, it is worth noting that the results obtained in this section focus on those time-variable mutation rate schemes that satisfy $\forall t\in\mathbb{N}: P_m(n,t)\in[0,1-1/\log n]$, i.e., the largest value that the mutation rate can take is $1-1/\log n$. It is unclear whether an even larger mutation rate that exceeds $1-1/\log n$ would help, though knowing when to apply such an extreme mutation rate relies on some ideal oracle (i.e., we know the consequence of such decisions in advance). To further validate the potential failure of \emph{all} time-variable mutation rate schemes on some specific BDOP, we will extend Theorem \ref{theorem:generalization_low_sigma} for the BDOP class to a concrete example, named the \textsc{BitMatching}$_D$ problem.

\subsection{A More Precise Result on
\textsc{BitMatching}$_D$}\label{sec:case_studies_bitmatching} In this subsection, we draw our attentions to a specific example of BDOPs class called \textsc{BitMatching}$_D$. Concretely, the \textsc{BitMatching}$_D$ problem is a BDOP using the following stationary function at the $t^{th}$ DOP phase:
\begin{eqnarray}\label{equation:bitmatching}
\textsc{BitMatching}_{D}(x,t):= n-H(x,x_t^*),
\end{eqnarray}
where $x_{t}^*$ is the global optimum of \textsc{BitMatching}$_D$ at the $t^{th}$ DOP phase.

To have a more comprehensive understanding of the performances of time-variable mutation rate schemes on \textsc{BitMatching}$_D$, here we consider the time complexity of finding approximate solutions with some certain quality, instead of considering only the time complexity of finding the moving global optimum. Here, the following specific characteristic for approximate solutions of DOPs, named the best-case DOP approximation ratio, is taken into account:
\begin{definition}[Best-case DOP approximation ratio]\label{definition:approximate_ratio}
Suppose we have a maximization DOP $I$ and an optimization algorithm $A$. Let $f_t(x_t^*)$ be the optimal value of the objective function of the DOP $I$ at the $t^{th}$ generation ($t\in\mathbb{N}$), $x_t$ be the best solution found by the algorithm $A$ at the $t^{th}$ generation, $f_t(x_t)$ be the value of the objective function with respect to the solution $x_t$ at the $t^{th}$ generation. For the algorithm $A$, the ratio $\sup_{\hat{t}\le t}\frac{f_{\hat{t}}(x_{\hat{t}})}{f_{\hat{t}}(x_{\hat{t}}^*)}$ is called the best-case DOP approximation ratio at the $t^{th}$ generation.
\end{definition}

Given the above definition, The definition of first hitting time in Eq. \ref{equation:first hitting time_re_eva} can be generalized to the so-called $(1-\epsilon)$-first hitting time concerning the time for finding the solutions that reach a certain best-case DOP approximation ratio, say, $1-\epsilon$:
\begin{definition}[$(1-\epsilon)$-first hitting
time for $(1+1)$ EA]\label{definition:epsilonFHT}On a DOP $\{f_t: t=1,2\dots\}$, the $(1-\epsilon)$-first hitting time of an EA, denoted by $\tau_\epsilon$, is defined as follows:
\begin{eqnarray}\label{equation:first hitting time_re_eva_solution_set}
&\tau_\epsilon:=\min\left\{t\ge 0; \left(x^{(P)}_t\in J_t\right)\vee\left(x^{(O)}_t\in J_t\right)\right\},
\end{eqnarray}
where $J_t:=\left\{x_t:\frac{f_t(x_t)}{f_t(x_t^*)}\ge 1-\epsilon\right\}$ and $\epsilon\in [0,1)$.
\end{definition}

The \textsc{BitMatching}$_D$ problem is a special case of the BDOP class. By generalizing the proof ideas mentioned in the last subsection, we are able to obtain a number of new theoretical results. The following lemma can be derived from Lemma \ref{lemma:_generalization_large_sigma}:
\begin{lemma}\label{lemma:large_sigma}  Given the \textsc{BitMatching}$_D$ problem with shifting rate $\sigma=\omega(\log n/n)$ and any time-variable mutation rate scheme $\left\{P_m(n,t)\in[0,1]:t\in\mathbb{N}\right\}$ of the $(1+1)$ EA, the $(1-1/\log^3 n)$-first hitting time of the $(1+1)$ EA is super-polynomial with an overwhelming probability. \end{lemma}

Based on Lemma \ref{lemma:large_sigma} and the proof idea of Theorem \ref{theorem:generalization_low_sigma}, we can obtain the following theorem:

\begin{theorem}\label{theorem:px_low_sigma} Given the \textsc{BitMatching}$_D$ problem with shifting rate $\sigma=\omega(\log n/n^2)$, and any time-variable mutation rate scheme $\left\{P_m(n,t)\in[0,1]:t\in\mathbb{N}\right\}$ of the $(1+1)$ EA, the $(1-\gamma^{4/7}\log n/n)$-first hitting time of $(1+1)$ EA is super-polynomial with an overwhelming probability, where $\gamma$ is defined by
\begin{eqnarray}\label{equation:def_gamma}
\gamma=\gamma(n,\sigma):=\min\bigg\{\frac{n}{\log n},\sigma\cdot\frac{n^{2}}{\log n}\bigg\}.
\end{eqnarray}
 \end{theorem}
The proofs of Lemma \ref{lemma:large_sigma} and Theorem \ref{theorem:px_low_sigma} are similar to those of Lemma \ref{lemma:_generalization_large_sigma} and Theorem \ref{theorem:generalization_low_sigma} respectively. Interested readers can refer to the appendix for details. Lemma \ref{lemma:large_sigma} and Theorem \ref{theorem:px_low_sigma} will lead to the following corollary about the most commonly used fixed mutation rate $1/n$ directly, which was proven by Dorste \cite{Droste03}:

\begin{corollary} The first hitting time
of the $(1+1)$ EA with the fixed mutation rate $1/n$ on \textsc{BitMatching}$_D$ problem with $\sigma=\omega(\log n/n^2)$ is super-polynomial with an overwhelming probability. \end{corollary} Combining the above corollary with Theorem \ref{theorem:px_low_sigma}, we obtain an interesting result:
\begin{corollary} Given the \textsc{BitMatching}$_D$ problem with $\sigma=\omega(\log n/n^2)$, both the $(1+1)$ EA with any time-variable mutation scheme and the $(1+1)$ EA with the most commonly used fixed mutation rate $1/n$ performs inefficiently. \end{corollary}
Clearly, the corollary indicates that no time-variable mutation rate schemes significantly outperforms the most commonly used fixed mutation rate $1/n$ when the $(1+1)$ EA is employed as the optimizer of the \textsc{BitMatching}$_D$ problem with shifting rate $\sigma=\omega(\log n/n^2)$. Moreover, Droste \cite{Droste03} proved that the $(1+1)$ EA with the fixed mutation rate $1/n$ can reach the global optimum of \textsc{BitMatching}$_D$ with shifting rate $\sigma=O(\log n/n^2)$ with a polynomial average first hitting time:

\begin{theorem}[Droste \emph{\cite{Droste03}}]\label{theorem:Droste_theorem} The mean first hitting time
of the $(1+1)$ EA with the fixed mutation rate $P_m=1/n$ on \textsc{BitMatching}$_D$ with $\sigma=O(\log n/n^2)$ is polynomial in the problem size $n$. \end{theorem}

Given that fixed mutation rates are only special cases of time-variable mutation schemes, the above result can also be interpreted as that there exists some time-variable mutation scheme with which the $(1+1)$ EA can solve \textsc{BitMatching}$_D$ with $\sigma=O(\log n/n^2)$ with a polynomial mean first hitting time. It follows from Theorem \ref{theorem:px_low_sigma} that the \textsc{BitMatching}$_D$ problem with \emph{$\sigma=\Theta(\log n/n^2)$ is the hardest \textsc{BitMatching}$_D$ on which a $(1+1)$ EA can guarantee efficient performance}. In the next section, we show that by adopting a population, a $(1+\lambda)$ EA with some time-variable mutation schemes can break the above limitation. However, when the shifting rate $\sigma=\Omega(\log n/n^2)$, a $(1+\lambda)$ EA with different time-variable mutation schemes will still encounter bottleneck when optimizing BDOPs.

\section{$(1+\lambda)$ EA with Time-Variable Mutation Schemes}\label{sec:1lambdaEA_hard_BDOPs}
So far we have analyzed the effectiveness of time-variable mutation schemes in the context of the $(1+1)$ EA. In this section, our analysis will be carried out in the context of a population-based EA called $(1+\lambda)$ EA. A case study on the \textsc{BitMatching}$_D$ problem will be given to show the overall impact of population and time-variable mutation schemes.

\subsection{A General Result}
The $(1+\lambda)$ EA studied in this paper follows the framework presented in Algorithm \ref{definition:1plus_lambdaEA}. The time variable mutation schemes for the $(1+\lambda)$ EA, defined in Definition \ref{def:scheme_1lambda}, allows the EA to utilize distinct mutation rates in generating different offsprings in the same generation. However, when solving BDOPs, such an EA may still be inefficient when the shifting rate of a BDOP exceeds $\Theta(\log n/n)$:
\begin{theorem}\label{theorem:large_sigma_1plus_lambda}Given any BDOP with shifting rate $\sigma=\omega(\log n/n)$ and any time-variable mutation rate scheme $\left\{P_m(n,t,\chi)\in [0,1-1/\log n]:t\in\mathbb{N},\chi\in\{1,\dots,\lambda\}\right\}$, the first hitting time of the $(1+\lambda)$ EA is super-polynomial with an overwhelming probability, where the offspring size $\lambda$ is a polynomial function of $n$. \end{theorem}
The proof of Theorem \ref{theorem:large_sigma_1plus_lambda} is a direct generalization of the proof idea of Lemma \ref{lemma:_generalization_large_sigma}. Interested readers can refer to the appendix for details.

Apparently, over the BDOP class, Theorem \ref{theorem:large_sigma_1plus_lambda} shows a theoretical limitation for time-variable mutation schemes associated with the $(1+\lambda)$ EA. Nevertheless, the theorem sheds some light on potential efficient performances of the $(1+\lambda)$ EA with time-variable mutation schemes on those BDOPs whose shifting rate is between $\Theta(\log n/n)$ and $\Theta(\log n/n^2)$ (note that Theorem \ref{theorem:generalization_low_sigma} tells us that the $(1+1)$ EA performs inefficiently on these BDOPs). In fact, compared with the $(1+1)$ EA, the $(1+\lambda)$ EA has indeed been strengthened by two factors. First, the offspring-population strategy, which is a specific way of utilizing population, offers larger selection pressure for the $(1+\lambda)$ EA. When optimizing DOPs, this feature is very helpful for tracking the movement of global optimum. Second, in one generation the $(1+\lambda)$ EA is capable of exploring different subsets of the search space via distinct step sizes. Owing to these two factors, it can be expected that the $(1+\lambda)$ EA can significantly outperform the $(1+1)$ EA on some BDOPs. In the next subsection, we present such a theoretical example.

\subsection{Case Studies on \textsc{BitMatching}$_D$} \label{sec:1pluslambdaBitMa}
As in Section \ref{sec:case_studies_bitmatching}, we still employ the \textsc{BitMatching}$_D$ problem as an example of the BDOP class. We show that the $(1+\lambda)$ EA with time-variable mutation schemes can improve the performance of the $(1+1)$ EA. Meanwhile, by the same theoretical result we are able to demonstrate that the general limitation of $(1+\lambda)$ EA over the BDOP class, predicted by Theorem \ref{theorem:large_sigma_1plus_lambda}, is almost tight. The main result of this subsection is as follows:

\begin{theorem}\label{theorem:1+lambda}
Given the \textsc{BitMatching}$_D$ problem with shifting rate $\sigma\le 1/(5n)$, if
\begin{enumerate}
\item[]\textbf{Cond. 1} the time-variable mutation rate scheme
$\left\{P_m(n,t,\chi)\in [0,1]:t\in\mathbb{N},\chi\in\{1,\dots,\lambda\}\right\}$ satisfies
\begin{eqnarray*}
&&\forall t\in \mathbb{N}: \sup_{t\in\mathbb{N},\chi\in\{1,\dots,\lambda\}}P_m(n,t,\chi)=O\left(\frac{\log n}{n}\right),\quad\quad \inf_{t\in\mathbb{N},\chi\in\{1,\dots,\lambda\}}P_m(n,t,\chi)\succ \frac{1}{Poly(n)};
\end{eqnarray*}
\item[]\textbf{Cond. 2} the polynomial offspring size $\lambda$ of the EA satisfies
\begin{eqnarray}\label{equation:offspring_size}
\lambda=\omega\left(\left(1-\sup_{t\in\mathbb{N},\chi\in\{1,\dots,\lambda\}}P_m(n,t,\chi)\right)^{-n}\left(\inf_{t\in\mathbb{N},\chi\in\{1,\dots,\lambda\}}P_m(n,t,\chi)\right)^{-1}\right).
\end{eqnarray}
\end{enumerate}
then the mean first hitting time of the $(1+\lambda)$ EA with the above time-variable mutation rate scheme is bounded from above by $8n/5$.
\end{theorem}
Theorem \ref{theorem:1+lambda} demonstrates that, when a time-variable mutation rate scheme satisfies certain conditions, then the $(1+\lambda)$ EA adopts such a mutation scheme will perform efficiently on the \textsc{BitMatching}$_D$ problem whose shifting rate is no larger than $1/(5n)$. The proof of the theorem utilizes the drift analysis technique proposed by He and Yao \cite{HeYao01}:
\begin{lemma}[Drift Analysis \cite{HeYao01}]\label{lemma:driftcondition}
Let $\xi_t$ ($t\in\mathbb{N}$) be the population at the $t^{th}$ generation of the EA, $D(\xi_t,t)$ be the distance metric measuring the distance between the population $\xi_t$ and the global optimum at the $t^{th}$ generation, and $\{D(\xi_t,t):t=0,1,\dots \}$ be a super-martingale describes an EA, if for any time $t=1,2,\dots$, if $D(\xi_t,t)>0$ and
\begin{eqnarray*}
\mathbb{E}\left[D\left(\xi_t,t\right)-D\left(\xi_{t+1},t+1\right)\mid \xi_t\right]\ge c_l>0,
\end{eqnarray*}
then the mean first hitting time satisfies
\begin{eqnarray*}
\mathbb{E}[\tau\mid \xi_0]\le \frac{D(\xi_0,0)}{c_l}\le \frac{\sup_{X\in\mathcal{P},t\in\mathbb{N}}D(X,t)}{c_l},
\end{eqnarray*}
where $\xi_0$ is the initial population of the EA, $\mathcal{P}$ is the set of all populations.
\end{lemma}

Before providing the formal proof for Theorem \ref{theorem:1+lambda}, we introduce some notations. Let $x_t^{(P)}$ be the parent individual of $t^{th}$ generation ($t\in\mathbb{N}$) of the $(1+\lambda)$ EA, and $x_t^{(1)},\dots,x_t^{(\lambda)}$ be the $\lambda$ offspring individuals generated by $x_t^{(P)}$. Let $x_t^{(O)}$ be the one with highest fitness among $x_t^{(1)},\dots,x_t^{(\lambda)}$:
\begin{eqnarray*}
\forall t\in\mathbb{N}: x_{t}^{(O)}:=\arg \max_{\chi\in\{1,\dots, \lambda\}} f_{t}(x^{(\chi)}_t).
\end{eqnarray*}
Let $x_t$ be the one with higher fitness between $x_t^{(P)}$ and $x_t^{(O)}$:
\begin{eqnarray*}
\forall t\in\mathbb{N}: x_{t}:=\left\{
\begin{array}{llll}
 x_{t}^{(P)}, &\mbox{if}\quad f_{t}(x_t^{(P)})> f_{t}(x_t^{(O)});\\
 x_{t}^{(O)}, &\mbox{if}\quad f_{t}(x_t^{(P)})\le  f_{t}(x_t^{(O)}).
 \end{array}
\right.
\end{eqnarray*}
Let $N_t^{(P)}:=n-H\left(x_t^{(P)},x_t^*\right)$, $N_t^{(O)}:=n-H\left(x_t^{(O)},x_t^*\right)$, and $N_t:=n-H(x_t,x_t^*)$. Specifically, for the \textsc{BitMatching}$_D$ problem, it is clear that $N_t$ is the larger one between $N_t^{(P)}$ and $N_t^{(O)}$, i.e., $N_t=\max\left\{N_t^{(P)},N_t^{(O)}\right\}$. Furthermore, define $\pi_{i,j}$ ($i,j\in\{0,1,\dots,n\}$) and the corresponding generalized notation as follows:
\begin{eqnarray*}
&&\pi_{i,j}:=\mathbb{P}\left(N_t^{(P)}=j\mid
N_{t-1}=i\right);\\
&&\pi_{i,\oplus j}:=\mathbb{P}\left(N_t^{(P)}\oplus j\mid N_{t-1}=i\right),
\end{eqnarray*}
where the ``$\oplus$'' at both sides can be replaced simultaneously by ``$>$'',``$<$'',``$\ge$'' or``$\le$''. $\pi_{i,j}$ is independent of the generation index $t$ since at each DOP phase the global optimum moves with the same shifting rate, as defined in Definition \ref{def:shifting_rate}. Define $\tilde{p}_{i,j}(t)$ ($i,j\in\{0,1,\dots,n\}$) and the corresponding generalized notation as follows:
\begin{eqnarray*}
&&\tilde{p}_{i,j}(t):=\mathbb{P}\left(N_t^{(O)}=j\mid
N_t^{(P)}=i\right);\\
&&\tilde{p}_{i,\oplus j}(t):=\mathbb{P}\left(N_t^{(O)}\oplus j\mid N_t^{(P)}=i\right),
\end{eqnarray*}
where the ``$\oplus$'' at both sides can be replaced simultaneously by ``$>$'',``$<$'',``$\ge$'' or``$\le$''. Based upon the above lemma and notations, we provide the proof of Theorem \ref{theorem:1+lambda}.

\textbf{\emph{Proof of Theorem \ref{theorem:1+lambda}.}} To apply Lemma \ref{lemma:driftcondition}, we need to define a suitable distance function, and estimate the corresponding one step mean drift at the $t^{th}$ ($t\in\mathbb{N}^+$) generation of the $(1+\lambda)$ EA. Given the union of the parent and its $\lambda$ offsprings as a whole population $X$, the distance function $D(X,t)$ ($t\in\mathbb{N}$), which measures the distance between $X$ and the moving global optimum at the $t^{th}$ generation, is formally defined by
\begin{eqnarray*}
D(X,t):=\min\left\{H(x,x_t^*);x\in X\right\},
\end{eqnarray*}
where $x^*_t$ is the global optimum at the $t^{th}$ generation ($t\in\mathbb{N}^+$). Then denote by $\Delta D_i(t)$ the one step mean drift at the $t^{th}$ generation, conditional on that the largest number of matching bits found at the $(t-1)^{th}$ generation is $N_{t-1}=i$:
\begin{eqnarray*}
\Delta D_i(t):=\mathbb{E}\left[D(X_{t-1},t-1)-D(X_t,t)|N_{t-1}=i\right]=\sum_{j=1}^{\infty}j\cdot\mathbb{P}\left(D(X_{t-1},t-1)-D(X_t,t)=j|N_{t-1}=i\right),
\end{eqnarray*}
which is the sum of the following four components:
\begin{eqnarray*}
&&\Delta_i\left[N_t^{(P)}\ge i,N_t^{(O)}\le N_t^{(P)}\right]\\
&&\quad\quad\quad:=\mathbb{E}\left[D(X_{t-1},t-1)-D(X_t,t);N_t^{(P)}\ge i,N_t^{(O)}\le N_t^{(P)}\big|N_{t-1}=i\right];\\
&&\Delta_i\left[N_t^{(P)}\ge i,N_t^{(O)}> N_t^{(P)}\right]\\
&&\quad\quad\quad:=\mathbb{E}\left[D(X_{t-1},t-1)-D(X_t,t); N_t^{(P)}\ge i,N_t^{(O)}> N_t^{(P)}\big|N_{t-1}=i\right];\\
&&\Delta_i\left[N_t^{(P)}< i, N_t^{(O)}\le i\right]\\
&&\quad\quad\quad:=\mathbb{E}\left[D(X_{t-1},t-1)-D(X_t,t); N_t^{(P)}< i, N_t^{(O)}\le i\big|N_{t-1}=i\right];\\
&&\Delta_i\left[N_t^{(P)}< i, N_t^{(O)}>i\right]\\
&&\quad\quad\quad:=\mathbb{E}\left[D(X_{t-1},t-1)-D(X_t,t);N_t^{(P)}< i, N_t^{(O)}>i\big|N_{t-1}=i\right].
\end{eqnarray*}
Formally, based the above notations the one step mean drift can be rewritten as
\begin{eqnarray*}
\Delta D_i(t)&=& \Delta_i\left[N_t^{(P)}\ge i,N_t^{(O)}\le N_t^{(P)}\right]+\Delta_i\left[N_t^{(P)}\ge i,N_t^{(O)}> N_t^{(P)}\right]\\&&+\Delta_i\left[N_t^{(P)}< i, N_t^{(O)}\le i\right]+\Delta_i\left[N_t^{(P)}< i, N_t^{(O)}>i\right].
\end{eqnarray*}
Next we estimate the four components one after another. By dividing the event ``$N_t^{(P)}\ge i,N_t^{(O)}\le N_t^{(P)}$'' into a number of sub-events ``$N_t^{(P)}=i,N_t^{(O)}\le N_t^{(P)}$'', ``$N_t^{(P)}=i+1,N_t^{(O)}\le N_t^{(P)}$'', $\dots$, ``$N_t^{(P)}=n,N_t^{(O)}\le N_t^{(P)}$'' whose probabilities are all positive, we know the following fact about the first component of the one step mean drift: {\small
\begin{eqnarray*}
&&\Delta_i\left[N_t^{(P)}\ge i,N_t^{(O)}\le N_t^{(P)}\right]=\sum_{j=i}^{n}(j-i)\pi_{i,j}\cdot\mathbb{P}\left(N_t^{(P)}\ge N_t^{(O)}\mid N_t^{(P)}=j\right)>0,
\end{eqnarray*}
}

Similarly, we can divide the event $N_t^{(P)}\ge i,N_t^{(O)}>N_t^{(P)}$ into several sub-events, and then estimate the following component of the one step mean drift: {\small
\begin{eqnarray*}
&&\Delta_i\left[N_t^{(P)}\ge i,N_t^{(O)}>N_t^{(P)}\right]=\sum_{j=i}^{n}\sum_{k=1}^{n-j}(j-i+k)\pi_{i,j}\cdot\mathbb{P}\left(N_t^{(P)}+k=N_t^{(O)}\mid
N_t^{(P)}=j\right)\\
&&\quad>\sum_{j=i}^{n}(j-i+1)\pi_{i,j}\cdot\mathbb{P}\left(N_t^{(O)}>N_t^{(P)}\mid
N_t^{(P)}=j\right)\\
&&\quad>\inf_{0\le k<n, t\in \mathbb{N}}\mathbb{P}\left(N_t^{(O)}>N_t^{(P)}\mid
N_t^{(P)}=k\right)\cdot\sum_{j=i}^{n}(j-i+1)\pi_{i,j}\\
&&\quad=\inf_{0\le k<n, t'\in \mathbb{N}}\tilde{p}_{k,>k}(t')\cdot\Bigg\{\left(\sum_{j=i}^{n}\pi_{i,j}\right)+\left(\sum_{j=i}^{n}(j-i){n-i\choose
j-i}\sigma^{j-i}(1-\sigma)^{n-j+i}\right)\Bigg\}\\
&&\quad>\inf_{0\le k<n, t'\in \mathbb{N}}\tilde{p}_{k,>k}(t')\cdot\Bigg\{\pi_{i,\ge i}+\left((1-\sigma)^{i}\sum_{j=0}^{n-i}j{n-i\choose
j}\sigma^{j}(1-\sigma)^{n-i-j}\right)\Bigg\}\\
&&\quad>\inf_{0\le k<n, t'\in \mathbb{N}}\tilde{p}_{k,>k}(t')\cdot\big\{\pi_{i,\ge i}+(1-\sigma)^{i}(n-i)\sigma\big\}\\
&&\quad>\inf_{0\le k<n, t'\in \mathbb{N}}\tilde{p}_{k,k+1}(t')\cdot\big\{\pi_{i,\ge i}+(1-\sigma)^{n}\sigma\big\},
\end{eqnarray*}}
where we utilize the fact that $\pi_{i,j}>{n-i\choose j-i}\sigma^{j-i}(1-\sigma)^{n-(j-i)}$ holds for all $j\ge i$. By similar calculations, we further obtain the following component of the one step mean drift: {\small\begin{eqnarray*} &&\Delta_i\left[N_t^{(P)}< i, N_t^{(O)}\le i\right]=\sum_{j=0}^{i-1}\sum_{k=1}^{i-j}(j-i+k)\pi_{i,j}\cdot\mathbb{P}\left(N_t^{(P)}+k=N_t^{(O)}\mid
N_t^{(P)}=j\right)\\
&&\quad\quad\quad\quad\quad\quad\quad\quad\quad\quad\quad\quad +\sum_{j=0}^{i-1}(j-i)\pi_{i,j}\cdot\mathbb{P}\left(N_t^{(O)}\le N_t^{(P)}\mid
N_t^{(P)}=j\right)\\
&&\quad>\sum_{j=0}^{i-1}\pi_{i,j}\cdot\sum_{k=1}^{i-j}\mathbb{P}\left(N_t^{(P)}+k=N_t^{(O)}\mid N_t^{(P)}=j\right)-\sum_{j=0}^{i-1}(i-j)\pi_{i,j}\cdot\sum_{k=1}^{i-j}\mathbb{P}\left(N_t^{(P)}+k=N_t^{(O)}\mid
N_t^{(P)}=j\right)\\
&&\quad\quad\quad\quad\quad\quad\quad\quad\quad\quad\quad\quad-\sum_{j=0}^{i-1}(i-j)\pi_{i,j}\cdot\mathbb{P}\left(N_t^{(O)}\le
N_t^{(P)}\mid N_t^{(P)}=j\right)\\
&&\quad\ge\sum_{j=0}^{i-1}\pi_{i,j}\cdot\sum_{k=1}^{i-j}\mathbb{P}\left(N_t^{(P)}+k=N_t^{(O)}\mid
N_t^{(P)}=j\right)-\sum_{j=0}^{i-1}(i-j)\pi_{i,j}\\
&&\quad>\inf_{0\le k<n, t'\in \mathbb{N}}\tilde{p}_{k,k+1}(t')\cdot\sum_{j=0}^{i-1}\pi_{i,j}-\sum_{j=0}^{i-1}(i-j){i \choose i-j}\sigma^{i-j}>\inf_{0\le k<n, t'\in
\mathbb{N}}\tilde{p}_{k,k+1}(t')\cdot\sum_{j=0}^{i-1}\pi_{i,j}-\sum_{k=1}^{i}k(i\sigma)^{k}\\
&&\quad=\inf_{0\le k<n, t'\in\mathbb{N}}\tilde{p}_{k,k+1}(t')\cdot\sum_{j=0}^{i-1}\pi_{i,j}-(i\sigma)\cdot\frac{i(i\sigma)^{i+1}-
(i+1)(i\sigma)^{i}+1}{(1-i\sigma)^2}\\
&&\quad>\inf_{0\le k<n, t'\in \mathbb{N}}\tilde{p}_{k,k+1}(t')\cdot\pi_{i,<i}-(n\sigma)\cdot\frac{n(n\sigma)^{n+1}- (n+1)(n\sigma)^{n}+1}{(1-n\sigma)^2}
\end{eqnarray*}}
Finally, the fourth component of the one step mean drift $\Delta D_i(t)$, as the first component, is positive: {\small
\begin{eqnarray*}
&&\Delta_i\left[N_t^{(P)}< i,N_t^{(O)}>i\right]>\sum_{j=i}^{n}\pi_{i,j}\cdot\mathbb{P}\left(N_t^{(O)}>i\mid N_t^{(P)}=j\right)>0.
\end{eqnarray*}
} The lower bounds of the four components yield the lower bound of $\Delta D_i(t)$: {\small
\begin{eqnarray*}
&&\Delta D_i(t)=\Delta_i\left[N_t^{(P)}\ge i,N_t^{(O)}\le N_t^{(P)}\right]+\Delta_i\left[N_t^{(P)}\ge i,N_t^{(O)}> N_t^{(P)}\right]+\Delta_i\left[N_t^{(P)}< i, N_t^{(O)}\le
i\right]+\Delta_i\left[N_t^{(P)}< i, N_t^{(O)}>i\right]\\
&&\quad>\inf_{0\le k<n, t'\in \mathbb{N}}\tilde{p}_{k,k+1}(t')\cdot\big\{\pi_{i,\ge i}+(1-\sigma)^{n}\sigma\big\}+\inf_{0\le k<n, t'\in
\mathbb{N}}\tilde{p}_{k,k+1}(t')\cdot\pi_{i,<i}-\sum_{k=1}^{i}k(i\sigma)^{k}\\
&&\quad>\inf_{0\le k<n, t'\in \mathbb{N}}\tilde{p}_{k,k+1}(t')\cdot\pi_{i,\ge i}+\inf_{0\le k<n, t'\in
\mathbb{N}}\tilde{p}_{k,k+1}(t')\cdot\pi_{i,<i}-\sum_{k=1}^{i}k(i\sigma)^{k}\\
&&\quad=\inf_{0\le k<n, t'\in \mathbb{N}}\tilde{p}_{k,k+1}(t')-\cdot\frac{i(i\sigma)^{i+2}- (i+1)(i\sigma)^{i+1}+(i\sigma)}{(1-i\sigma)^2},
\end{eqnarray*}
} where $\sum_{k=1}^{i}k(i\sigma)^{k}$ is monotonically increasing with positive $i$ and $\sigma$. Given the fact $i<n$ and the condition $\sigma\le 1/(5n)$, we further have: {\small
\begin{eqnarray*}
\Delta D_i(t)&>&\inf_{0\le k<n, t'\in \mathbb{N}}\tilde{p}_{k,k+1}(t')-\frac{\frac{n}{5^{n+2}}- \frac{n+1}{5^{n+1}}+\frac{1}{5}}{\frac{16}{25}}>\inf_{0\le k<n, t'\in
\mathbb{N}}\tilde{p}_{k,k+1}(t')-\frac{5}{16}\\
&>&1-\left(1-\left(1-\sup_{t\in\mathbb{N},\chi\in\{1,\dots,\lambda\}}P_m(n,t,\chi)\right)^{n-1}\inf_{t\in\mathbb{N},\chi\in\{1,\dots,\lambda\}}P_m(n,t,\chi)\right)^\lambda-\frac{5}{16}\\
&>&\frac{11}{16}-\left(1-\left(1-\sup_{t\in\mathbb{N},\chi\in\{1,\dots,\lambda\}}P_m(n,t,\chi)\right)^{n}\inf_{t\in\mathbb{N},\chi\in\{1,\dots,\lambda\}}P_m(n,t,\chi)\right)^\lambda.
\end{eqnarray*}
} Under the conditions $\sup_{t\in\mathbb{N},\chi\in\{1,\dots,\lambda\}}P_m(n,t,\chi)=O(\log n/n)$ and $\inf_{t\in\mathbb{N},\chi\in\{1,\dots,\lambda\}}P_m(n,t,\chi)\succ 1/Poly(n)$,
\begin{eqnarray*}
\left(1-\sup_{t\in\mathbb{N},\chi\in\{1,\dots,\lambda\}}P_m(n,t,\chi)\right)^{n}\inf_{t\in\mathbb{N},\chi\in\{1,\dots,\lambda\}}P_m(n,t,\chi)\succ \frac{1}{Poly(n)},
\end{eqnarray*}
and the above item is strictly less than $1$. Hence, there exists a \emph{polynomial} offspring size
\begin{eqnarray*}
\lambda=\omega\left(\left(1-\sup_{t\in\mathbb{N},\chi\in\{1,\dots,\lambda\}}P_m(n,t,\chi)\right)^{-n}\left(\inf_{t\in\mathbb{N},\chi\in\{1,\dots,\lambda\}}P_m(n,t,\chi)\right)^{-1}\right),
\end{eqnarray*}
such that the one step mean drift at the $t^{th}$ generation ($\forall t<\tau, t\in\mathbb{N}^+$) has a common lower bound
\begin{eqnarray*}
\forall t<\tau, t\in\mathbb{N}^+:\Delta_i(t)>\frac{5}{8}.
\end{eqnarray*}
According to Lemma \ref{lemma:driftcondition}, given the offspring size $\lambda$ specified by Eq. \ref{equation:offspring_size}, the mean first hitting time of the EA under two conditions of Theorem \ref{theorem:1+lambda} satisfies {\small\begin{eqnarray*} \mathbb{E}\left[\tau | \text{Cond. 1}, \text{Cond. 2}\right]<\frac{8n}{5}.
\end{eqnarray*}}
Under this circumstance, \emph{the number of function evaluations} of finding the global optimum for the first time, denoted by $T$, satisfies {\small\begin{eqnarray*} \mathbb{E}\left[T | \text{Cond. 1}, \text{Cond. 2}\right]<\frac{8\lambda n}{5}.
\end{eqnarray*}}
$\hfill\square$

The above theorem offers a sufficient condition for the $(1+\lambda)$ EA to achieve efficient performance on the \textsc{BitMatching}$_D$ problem with shifting rate $\sigma\le 1/(5n)$. As a direct corollary of Theorem \ref{theorem:1+lambda}, we present the following time complexity result of the $(1+n^2\log n)$ EA whose time-variable mutation scheme $\{P_m(n,t,\chi)\in [0,1]:t\in\mathbb{N},\chi\in\{1,\dots,\lambda\}\}$ satisfying $\sup_{t\in\mathbb{N},\chi\in\{1,\dots,\lambda\}}P_m(n,t,\chi)=\log n/n$, $\inf_{t\in\mathbb{N},\chi\in\{1,\dots,\lambda\}}P_m(n,t,\chi)= 1/n$ (note that the fixed mutation rate $1/n$ is a special case of such time-variable schemes):

\begin{corollary}
Given the \textsc{BitMatching}$_D$ problem with shifting rate $\sigma\le 1/(5n)$, if the time-variable mutation rate scheme $\left\{P_m(n,t,\chi)\in [0,1]:t\in\mathbb{N},\chi\in\{1,\dots,\lambda\}\right\}$ of the $(1+n^2\log n)$ EA satisfies
\begin{eqnarray*}
\forall t\in \mathbb{N}: \sup_{t\in\mathbb{N},\chi\in\{1,\dots,\lambda\}}P_m(n,t,\chi)=\log n/n,\quad\quad \inf_{t\in\mathbb{N},\chi\in\{1,\dots,\lambda\}}P_m(n,t,\chi)=\frac{1}{n},
\end{eqnarray*}
then the mean number of function evaluations of finding the global optimum for the first time is bounded from above by $8n^3\log n/5$.
\end{corollary}

Theorem \ref{theorem:1+lambda} tells us, when the shifting rate of the global optimum is smaller than $1/(5n)$ (each movement only changes no more than $1/5$ bit of the global optimum on average), the $(1+\lambda)$ EA is able to compensate the negative influence brought by the movement of the global optimum via generating a number of offspring individuals with different mutation rates (in each generation), and the EA can achieve efficient performance on the \textsc{BitMatching}$_D$ problem. However, by the following theorem we will know that, when the shifting rate $\sigma$ further grows to $\omega(\log n/n)$ (each movement changes at least $\log n$ bits of the global optimum on average), even the multiple-offspring strategy and time-variable mutation rate schemes cannot help to achieve efficient performance on the \textsc{BitMatching}$_D$ problem.

\begin{theorem}\label{theorem:large_sigma_1plus_lambda_bitmatching} Given the \textsc{BitMatching}$_D$ problem with shifting rate $\sigma=\omega(\log n/n)$, and \textbf{any} time-variable mutation rate scheme $\left\{P_m(n,t,\chi)\in [0,1]:t\in\mathbb{N},\chi\in\{1,\dots,\lambda\}\right\}$, the first hitting time of the $(1+\lambda)$ EA is super-polynomial with an overwhelming probability. \end{theorem}
The proof of Theorem \ref{theorem:large_sigma_1plus_lambda_bitmatching} follows the proof idea of Lemma \ref{lemma:_generalization_large_sigma} and Theorem \ref{theorem:large_sigma_1plus_lambda}. Interested readers can refer to the appendix for details.

\section{Discussions}\label{section:discussion}
In this section, we discuss some issues related to the theoretical results presented in previous sections.

\subsection{Generalizations of Theoretical Results}
Here we discuss potential ways of generalizing our theoretical results from different perspectives.
\subsubsection{What if the shifting rate of a DOP is also time-variable?}
Technically, the theoretical results presented so far can further be generalized to a broader class of DOPs. In particular, we can modify the definition of Bitwise Shifting Global Optimum (BSGO) (Definition \ref{def:shifting_rate}) by allowing the global optimum to move with different shifting rates at different DOP phases:

\begin{definition}[Bitwise Shifting Global Optimum with Time-variable Shifting Rate (BSGO-TSR)]\label{def:shifting_rate_updated} The global optimum of a DOP is called a BSGO-TSR, if it is shifting following the rule $\forall t\in\mathbb{N}: x_{t+1}^*=\mathcal{B}_{n,t}(x_{t}^*)$, where $\mathcal{B}_{n,t}:\{0,1\}^n\to\{0,1\}^n$ flips every bit of the input binary string with a probability of $\sigma(t)\in(0,1/2]$, and $\sigma(t)$ is called the \textbf{time-variable shifting rate}. \end{definition}

The theoretical results obtained in this paper can be generalized by replacing the time-invariable shifting rate of the BDOP with the above time-variable shifting rate, and the corresponding proofs will not be significantly different from existing ones. As an example, we present a generalized version of Theorem \ref{theorem:px_low_sigma} as an example:
\begin{proposition} Given the \textsc{BitMatching}$_D$ problem with shifting rate $\{\sigma(t)=\omega(\log n/n^2):t\in\mathbb{N}^+\}$, and any time-variable mutation rate scheme $\left\{P_m(n,t)\in[0,1]:t\in\mathbb{N}\right\}$ of EA, the $(1-\gamma^{4/7}\log n/n)$-first hitting time of $(1+1)$ EA is super-polynomial with an overwhelming probability, where $\gamma$ is defined by
\begin{eqnarray}\label{equation:def_gamma}
\gamma=\gamma(n,\sigma):=\min\bigg\{\frac{n}{\log n},\sigma\cdot\frac{n^{2}}{\log n}\bigg\}.
\end{eqnarray}
 \end{proposition}
The above proposition replaces the fixed shifting rate $\sigma=\omega(\log n/n^2)$ in Theorem \ref{theorem:px_low_sigma} with the time-variable shifting rate $\{\sigma(t)=\omega(\log n/n^2):t\in\mathbb{N}^+\}$. Similarly, we can generalize other theorems in the paper. Such generalizations are correct since our theoretical analysis does not utilize concrete values of the shifting rate $\sigma$, but only relies on the upper bound, lower bound or asymptotic order of $\sigma$. Hence, the original proofs of our theorems can easily be relaxed and utilized as proofs of such generalizations. For the sake of brevity, we do not provide the detailed analysis in this paper.

\subsubsection{Characterizing all forms of adaptations by condition-variable mutation rate schemes}
The time-variable mutation rate schemes studied in this paper are defined in Definitions \ref{def:scheme_11} and \ref{def:scheme_1lambda}. In our theoretical analysis showing theoretical limitations of time-variable mutation rate schemes, we avoid to utilize the concrete values of mutation rates. Instead, we optimistically considered that, by the help of an \emph{oracle}, an EA can always choose the most promising mutation rates in every generation. Such a notion can be alternatively characterized by explicitly involving sophisticated information (e.g, fitness of current individuals) as conditions for specifying mutation rates, which yields the definition of condition-variable mutation rate schemes for $(1+\lambda)$ EA\footnote{As stated, when $\lambda=1$, the $(1+\lambda)$ EA is equivalent to the $(1+1)$ EA}:
\begin{definition}[Condition-variable mutation rate scheme for $(1+\lambda)$ EA] The condition-variable mutation rate scheme of the $(1+\lambda)$ EA is a mapping $P_m:\mathbb{N}\times\mathbb{N}\times\{1,\dots,\lambda\}\times\mathbb{CS}\to [0,1]$, where $\mathbb{CS}$ is the condition space consisting of all potential conditions that may contribute to the decision of mutation rates. Such a scheme sets the mutation rate for obtaining the $\chi^{th}$ offspring individual at the $t^{th}$ generation be $P_m(n,t,\chi,\textbf{C}_t)$ in the presence of $\textbf{C}_t$.
\end{definition}
The above definition is general enough to characterize all forms of adaptations. Owing to the oracle notion utilized in our paper, all theoretical results proven for time-variable mutation schemes also hold accordingly when condition-variable schemes replace time-variable schemes. For example, the generalized versions of Theorems \ref{theorem:generalization_low_sigma} and \ref{theorem:large_sigma_1plus_lambda} with respect to condition-variable schemes are

\begin{proposition} Given any BDOP with shifting rate $\sigma=\omega(\log n/n^2)$ and any condition-variable mutation rate scheme $\left\{P_m(n,t,\textbf{C}_t)\in[0,1-1/\log n]:t\in\mathbb{N},\textbf{C}_t\in \mathbb{CS}\right\}$, the first hitting time of the $(1+1)$ EA is super-polynomial with an overwhelming probability.
 \end{proposition}

\begin{proposition}Given any BDOP with shifting rate $\sigma=\omega(\log n/n)$ and any condition-variable mutation rate scheme $\left\{P_m(n,t,\chi,\textbf{C}_t)\in [0,1-1/\log n]:t\in\mathbb{N},\chi\in\{1,\dots,\lambda\},\textbf{C}_t\in \mathbb{CS}\right\}$, the first hitting time of the $(1+\lambda)$ EA is super-polynomial with an overwhelming probability, where the offspring size $\lambda$ is a polynomial function of $n$. \end{proposition}
For the correctness of the propositions, a straightforward and intuitive explanation is that no matter which condition-variable mutation rate scheme an EA adopts, in the optimization process it has to follow a time-dependent configuration of concrete mutation rates that can be viewed as \emph{a time-variable mutation rate scheme specified by an oracle}. Since we have proven the high failure probability of each time-variable mutation rate scheme, we cannot expect that such a concrete setting can be effective if it is online-specified by some condition-variable mutation scheme. From a technical perspective, by looking at the details of the proofs of the theorems, it is easy to find that they have considered in detail all possible random transitions between different subsets decomposed from the solution space, and the decisions of any condition-variable mutation rate scheme under different conditions in every generation have been included in the analysis. To sum up, the theoretical results obtained in this paper are general enough to show theoretical limitations of adaptations of mutation rates in evolutionary algorithms.

\subsection{Conjectures about $(\mu+\lambda)$ EA}
In the evolutionary computation community, the $(\mu+\lambda)$ EAs, which maintain $\mu$ parents and generate $\lambda$ offsprings in each generation, have received extensive investigations over the past decades. Apparently, the $(1+1)$ and $(1+\lambda)$ EAs studies in this paper are special cases of $(\mu+\lambda)$ EAs. After showing the theoretical limitations of time-variable mutation rate schemes for both EAs, a natural question is, whether such theoretical results can be generalized to other $(\mu+\lambda)$ EAs' cases?

Assume that each time-variable mutation rate scheme of a $(\mu+\lambda)$ EA allows the algorithm to adopt $\lambda$ (not necessarily different) mutation rates when generating $\lambda$ offsprings at each generation. For any of such EAs, we conjecture that the time-variable mutation rate schemes fail to help them to perform efficiently when the shifting rate of a BDOP exceeds some threshold. However, for different settings of $\mu$ and $\lambda$, the concrete thresholds might be different (as shown by our results). Intuitively, we conjecture that when the ratio $\lambda/\mu$ becomes larger, the threshold of shifting rate will become higher. Nevertheless, this does not mean that we can excessively enhance the threshold by increasing the ratio $\lambda/\mu$, and the maximal threshold of shifting rate for any $(\mu+\lambda)$ EA might converge to $\Theta(\log n/n)$. When the shifting rate grows to $\omega(\log n/n)$, from one DOP phase to the next phase the global optimum of a BDOP will change more than $\log n$ of its bits on average, which is too drastic for an EA to track. The rigorous proofs for the above conjectures will be left as our future work.

\subsection{Impact of Population on Evolutionary Dynamic Optimization}
We study both the $(1+1)$ and $(1+\lambda)$ EAs in this paper such that the impact of population can be demonstrated. The former is an individual-based EA, and the latter can be regarded as a population-based EA adopting the multiple-offspring strategy (a concrete way of utilizing population). Our theoretical results clearly demonstrate the positive impact of population on the performance of EA in terms of BDOPs with distinct shifting rates. To be specific, in the absence of the multiple-offspring strategy, the largest shifting rate of the BDOP class that a $(1+1)$ EA can deal with efficiently is $\Theta(\log n/n^2)$ (Theorems \ref{theorem:generalization_low_sigma} and \ref{theorem:Droste_theorem}). After adopting the multiple-offspring strategy, the $(1+\lambda)$ EA can solve efficiently the \textsc{BitMatching}$_D$ problem with a shifting rate growing to $\sigma\le 1/(5n)$. In the evolutionary computation field, this is the first time that the positive impact is validated in the context of evolutionary dynamic optimization.

\subsection{Adaptation of Mutation Rate is not a Panacea}

In this paper, the effectiveness of time-variable mutation rate schemes is investigated on two testbeds, that is, the $(1+1)$ and $(1+\lambda)$ EAs. On a BDOP whose global optimum is consistently shifting, one might expect that there is some time-variable mutation scheme which can assist the EA to track the optimum by ``cleverly'' and dynamically choosing appropriate mutation rate, such that the mutation rates of an EA can ``fit'' the stochastic movement of the global optimum. However, for both the $(1+1)$ and $(1+\lambda)$ EAs we show that there are classes of BDOPs on which various time-variable mutation rate schemes fail to help EAs to perform efficiently. Moreover, our theoretical analysis has further been generalized to a concrete instance of the BDOP class called \textsc{BitMatching}$_D$. When optimizing the \textsc{BitMatching}$_D$ problem whose shifting rate exceeds the theoretical threshold $\Theta(\log n/n^2)$, no time-variable mutation rate scheme can assist the $(1+1)$ EA to optimize efficiently the problem. When optimizing the \textsc{BitMatching}$_D$ problem whose shifting rate exceeds the theoretical bound $\Theta(\log n/n)$, no time-variable mutation rate scheme can assist the $(1+\lambda)$ EA ($\lambda$ is polynomial in $n$) to optimize efficiently the problem. Given the fact that the static \textsc{BitMatching} problem can be solved by the $(1+1)$ EA with $O(n\ln n)$ generations \cite{Droste02}, and by the $(1+\lambda)$ EA with $O(n\ln n)$ function evaluations given an appropriate $\lambda$ \cite{jan05}, for both EAs the hardness of \textsc{BitMatching}$_D$ mainly comes from the movement of the global optimum. For real-world DOPs with not-too-simple stationary objective functions and a moving global optimum, it is highly likely that even a well-designed time-variable mutation rate scheme is insufficient to improve the performance of an EA, or even the ``promising'' time-variable mutation rate scheme does not exist. Meanwhile, noting that designing a delicate time-variable scheme could be rather time-consuming, it might be better to follow the well-known Occam's razor and use some fixed mutation rate, unless one can ensure that the optima of DOPs are not moving too fast.

\section{Conclusion and Future Work} \label{section:conclusion}
In this paper, we theoretically study the relationship between time-variable mutation rate scheme and time complexity of EAs on a class of DOPs. The analytical results are given in terms of the first hitting time of finding the moving global optimum. By decomposing the search space and estimating transitions among the resultant subspaces (intervals), our analysis shows that, when optimizing a class of DOPs, theoretical limitations do exist for both $(1+1)$ and $(1+\lambda)$ EAs with any time-variable mutation rate. Such theoretical results may lead to new understanding of the role of mutation in solving DOPs: although some specific time-variable mutation schemes have proven or validated to be helpful on some static optimization problems, it might be not be beneficial to seek for some sophisticated time-variable mutation rate scheme to improve the performances of EAs on many DOPs with moving global optima.

It is worth noting that we have not taken the interactions among the adaptations of parameters in different operators (e.g., mutation and crossover) or strategies (e.g., population) of EAs into account. It is possible that the combinations of different strategies can improve the performances of EAs on DOPs. Nevertheless, it seems likely that the EA will still meet some new theoretical limitation when optimizing BDOPs. In the future, we will try to carry out such theoretical studies following the methodology utilized in this paper, so as to gain more insight into the adaptations of EAs' operators and strengthen the theoretical foundations of EAs.

\appendix

\section{Analytical Tools} Before providing the proof of Lemma \ref{lemma:_generalization_large_sigma} and Theorem \ref{theorem:generalization_low_sigma}, we need to introduce a number of lemmas. First, three mathematical tools from
previous literatures are presented directly without proofs.
 \begin{lemma}[\textbf{Chernoff bounds} \cite{Motwani95book}]
\label{chernoff} Let $a_1,a_2,\dots,a_k \in \{0,1\}$ be $k$ independent random variables with the same distribution:
\begin{eqnarray*}
 \forall i\neq j: \mathbb{P}(a_i=1)=\mathbb{P}(a_j=1),
\end{eqnarray*}
 where $i,j\in\{1,\dots,k\}$. Let $a=\sum_{i=1}^k a_i$.
\begin{itemize}
 \item $\forall
 0<\delta<1$:
\begin{eqnarray}\label{equation:chernoff1}
 \mathbb{P}\Big(a<(1-\delta)\mathbb{E}[a]\Big)<e^{-\mathbb{E}[a]\delta^2/2}.
\end{eqnarray}
 \item $\forall \delta\le 2e-1$:
 \begin{eqnarray}\label{equation:chernoff2}
 \mathbb{P}\Big(a>(1+\delta)\mathbb{E}[a]\Big)<e^{-\mathbb{E}[a]\delta^2/4}.
\end{eqnarray}
 \item $\forall \delta>0$:
 \begin{eqnarray}\label{equation:chernoff3}
 \mathbb{P}\Big(a>(1+\delta)\mathbb{E}[a]\Big)<\Bigg(\frac{e^\delta}{(1+\delta)^{1+\delta}}\Bigg)^{\mathbb{E}[a]}.
\end{eqnarray}
\end{itemize}
\end{lemma}
Chernoff bounds are widely used in theoretical analysis of EAs \cite{Droste02,Chen08b,Chen11}, and play important role in proving the theoretical result presented in this paper. Moreover, we present the following lemma proven by Droste \cite{Droste03}.

 \begin{lemma}[Droste \cite{Droste03}]\label{lemma:droste} Let
$w_1,\dots,w_i, v_1,\dots,v_j \in \{0,1\}$ be $i+j$ independent random variables with the same distribution, then
\begin{eqnarray*}
\mathbb{P}\Bigg(\sum_{k=1}^i w_k>\sum_{k=1}^j v_k\Bigg)\le \frac{i}{j}.
\end{eqnarray*}\\
\end{lemma}
\begin{lemma}[\cite{GKP94ConcreteMath}]\label{lemma:bi_coeffi}
Given integers $c$ and $d$,
\begin{eqnarray*}
\sum_{k}{r\choose c+k}{s\choose d-k}={r+s\choose c+d} \quad\quad\text{holds.}
\end{eqnarray*}
\end{lemma}

\section{Transition Lemmas with Proofs}

\subsection{Transition Lemmas}
Based on the three basic lemmas, the definitions and notations introduced in previous sections, we will present several ``transition lemmas'', concerning the transition probabilities between different subintervals of $[0,n]$, in responses to the DOP change and the mutation operator of EA. The purpose of employing these lemmas is to pack the propositions concerning the transitions between different intervals defined in Section \ref{sec:11EA_hard_BDOPs}, so that they can be directly utilized in the proofs of Lemmas \ref{lemma:_generalization_large_sigma} and \ref{lemma:large_sigma}, Theorem \ref{theorem:generalization_low_sigma}, and \ref{theorem:px_low_sigma}. As a result, the above proofs can be significantly simplified. Given the notations $N_t^{(P)}$, $N_t^{(O)}$ and $N_t$ defined in Section \ref{sec:1pluslambdaBitMa}, the transition lemmas can be presented as follows:

\begin{lemma}\label{lemma:_adaptive_prob_sigma} \label{lemma:_adaptive_prob_sigmapm} Given any BDOP with $\sigma=\omega(\log n/n)$ and $\sigma\in (0,1/2]$, for the $t^{th}$ generation ($t\in\mathbb{N}$) of the $(1+\lambda)$ EA with the mutation rate configuration $\{P_m(n,t,\chi)\in [0,1]:\chi\in\{1,\dots,\lambda\}\}$,
\begin{enumerate}
\item[1.] if $N_{t-1}=i=o(n)$ and $N_{t-1}=i>n/\log^2n$ ($t\ge 1$), then
the probability of $N_{t}^{(P)}=j\in [n-\frac{n}{\log^3 n},n]$ is super-polynomially small.

\item[2.] if $\exists$ constants $\epsilon_1,\epsilon_2\in(0,1)$ such that
$\epsilon_2\le\epsilon_1$ and $\epsilon_2 n\le N_{t-1}=i\le\epsilon_1 n$, then the probability of $N_{t}^{(P)}=j\in [n-\frac{n}{\log^3 n},n]$ is super-polynomially small.

\item[3.] if $N_{t-1}=i=n-o(n)$ ($t\ge 1$), then
the probability of $N_{t}^{(P)}=j>N_{t-1}-i\sigma/4$ is super-polynomially small, where $i\sigma/4=\omega(\log n)$.
\item[4.] the above three propositions also hold if $N_t^{(P)}$ is replaced by $N_t^{(\chi)}$, and $i\sigma$ is replaced by $i(P_m(n,t,\chi)+\sigma-2P_m(n,t,\chi)\sigma)$ (in the $3^{rd}$ proposition).
\end{enumerate}
\end{lemma}

%
%

\begin{lemma}\label{lemma:_adaptive_prob_pm}\label{lemma:_adaptive_prob_sigmapm_any_eta}
Given any BDOP with $\sigma\in (0,1/2]$, for the $t^{th}$ generation ($t\in\mathbb{N}$) of the $(1+\lambda)$ EA with the mutation rate configuration $\{P_m(n,t,\chi)\in [0,1]:\chi\in\{1,\dots,\lambda\}\}$,
%
\begin{enumerate}
\item[1.] if $N_{t}^{(P)}=i=o(n)$, $N_{t}^{(P)}=i>n/\log^2n$ and
 $P_m(n,t,\chi)=\omega(\log n/n)$, then
the probability of $N_{t}^{(\chi)}=j\in [n-\frac{n}{\log^3 n},n]$ is super-polynomially small.

\item[2.] if $\exists$ constants $\epsilon_1,\epsilon_2\in(0,1)$ such that
$\epsilon_2\le\epsilon_1$, $\epsilon_2 n\le N_{t}^{(P)}=i\le\epsilon_1 n$ and
 $P_m(n,t,\chi)=\omega(\log n/n)$, then
the probability of $N_{t}^{(\chi)}=j\in [n-\frac{n}{\log^3 n},n]$ is super-polynomially small.

\item[3.] if $N_{t}^{(P)}=i=n-o(n)$ and
 $P_m(n,t,\chi)=\omega(\log n/n)$, then
the probability of $N_{t}^{(\chi)}=j>N_{t}^{(P)}-iP_m(n,t,\chi)/4$ in one generation is super-polynomially small, where $iP_m(n,t,\chi)/4=\omega(\log n)$.
\item[4.] the above three propositions also hold if ``$t\in\mathbb{N}$'' is replaced by ``$t\in\mathbb{N}^+$'', $N_t^{(P)}$ is replaced by $N_{t-1}$, and $iP_m(n,t,\chi)$ is replaced by $i(P_m(n,t,\chi)+\sigma-2P_m(n,t,\chi)\sigma)$ (in the $3^{rd}$ proposition).
\end{enumerate}
\end{lemma}

%
%

\begin{lemma}\label{lemma:_adaptive_prob_pm_sym}
Given the \textsc{BitMatching}$_D$ problem with $\sigma\in (0,1/2]$, for the $t^{th}$ generation ($t\in\mathbb{N}$) of the $(1+\lambda)$ EA with the mutation rate configuration $\{P_m(n,t,\chi)\in [0,1]:\chi\in\{1,\dots,\lambda\}\}$,
\begin{enumerate}
\item[1.] if $P_m(n,t,\chi)=\omega(\log n/n)$, and $\exists$ constants $\epsilon_1$ and $\epsilon_2$ such that
$0<\epsilon_2\le\epsilon_1<1$, $\epsilon_2 n\le N_{t}^{(P)}=i\le\epsilon_1 n$, then the probability of $N_{t}^{(\chi)}=j\in [0,\frac{n}{\log^2 n}]$ is super-polynomially small.

\item[2.] if $N_{t}^{(P)}=i=o(n)$ and
 $P_m(n,t,\chi)=\omega(\log n/n)$, then
the probability of $N_{t}^{(\chi)}=j<N_{t}^{(P)}+(n-i)P_m(n,t,\chi)/4$ is super-polynomially small, where $(n-i)P_m(n,t,\chi)/4=\omega(\log n)$.

\end{enumerate}
\end{lemma}

\begin{lemma}\label{lemma:_adaptive_prob_sigma_sym} \label{lemma:_adaptive_prob_sigmapm_sym}
Given any BDOP with $\sigma=\omega(\log n/n)$ and $\sigma\in (0,1/2]$, for the $t^{th}$ generation ($t\in\mathbb{N}$) of the $(1+\lambda)$ EA with the mutation rate configuration $\{P_m(n,t,\chi)\in [0,1-1/\log n]:\chi\in\{1,\dots,\lambda\}\}$ (\emph{Specifically, if the BDOP is \textsc{BitMatching}$_D$, then the condition $\{P_m(n,t,\chi)\in [0,1-1/\log n]:\chi\in\{1,\dots,\lambda\}\}$ can be further relaxed to $\{P_m(n,t,\chi)\in [0,1]:\chi\in\{1,\dots,\lambda\}\}$}),
%
\begin{enumerate}
\item[1.] if $N_{t-1}=i=n-o(n)$ and $N_{t-1}=i<n-n/\log^3n$, then
the probability of $N_{t}^{(P)}=j\in [0,\frac{n}{\log^2 n}]$ is super-polynomially small.

\item[2.] if $\exists$ constants $\epsilon_1,\epsilon_2\in(0,1)$ such that
$\epsilon_2\le\epsilon_1$, and $\epsilon_2 n\le N_{t-1}=i\le\epsilon_1 n$, then the probability of $N_{t}^{(P)}=j\in [0,\frac{n}{\log^2 n}]$ is super-polynomially small.

\item[3.] if $N_{t-1}=i=o(n)$, then
the probability of $N_{t}^{(P)}=j<N_{t-1}+(n-i)\sigma/4$ is super-polynomially small, where $(n-i)\sigma/4=\omega(\log n)$.

\item[4.] the above three propositions also hold if $N_t^{(P)}$ is replaced by $N_t^{(\chi)}$, and $(n-i)\sigma$ is replaced by $(n-i)(P_m(n,t,\chi)+\sigma-2P_m(n,t,\chi)\sigma)$ (in the $3^{rd}$ proposition).
\end{enumerate}
\end{lemma}

%
%

When $\lambda=1$, the above lemmas hold for the $(1+1)$ EA. The proofs of the above lemmas are very similar to each other. For the sake of brevity, next we only provide the detailed proofs of Lemmas \ref{lemma:_adaptive_prob_sigma} and \ref{lemma:_adaptive_prob_sigmapm_sym}.1.

\subsection{Proof of Lemma \ref{lemma:_adaptive_prob_sigma}}
To prove the transition lemmas, we need the following lemma:
\begin{lemma}\label{lemma:two_bitwise} Let
$r(n,t,\chi)=P_m(n,t,\chi)(1-\sigma)+\sigma(1-P_m(n,t,\chi))$ be the composite bitwise mapping rate for for the $\chi^{th}$ offspring generated at the $t^{th}$ generation ($t\in\mathbb{N}^+$). It satisfies that
\begin{eqnarray}\label{equation:two_bitwise}
\forall t\in\mathbb{N}^+:r(n,t,\chi)\in \Big(\min\bigg\{\frac{1}{2},\max\Big\{\sigma,P_m(n,t,\chi)\Big\}\bigg\},\max\Big\{\frac{1}{2},P_m(n,t,\chi)\Big\}\Big].
\end{eqnarray}
\end{lemma}

 Noting that
$\sigma\in(0,1/2]$, we have:
\begin{enumerate}
\item If $P_m(n,t,\chi)\in(0,\frac{1}{2}]$,
\begin{eqnarray*}
&&r(n,t,\chi)=\sigma+(1-2\sigma)P_m(n,t,\chi)\le\sigma+\frac{1}{2}(1-2\sigma)=\frac{1}{2},\\
&&r(n,t,\chi)=P_m(n,t,\chi)+(1-2P_m(n,t,\chi))\sigma>P_m(n,t,\chi),\\
&&r(n,t,\chi)=\sigma+(1-2\sigma)P_m(n,t,\chi)>\sigma;
\end{eqnarray*}

\item If $P_m(n,t,\chi)\in (\frac{1}{2},1)$,
\begin{eqnarray*}
&&r(n,t,\chi)=P_m(n,t,\chi)+(1-2P_m(n,t,\chi))\sigma<P_m(n,t,\chi)\\
&&r(n,t,\chi)=\sigma+(1-2\sigma)P_m(n,t,\chi)>\sigma+\frac{1}{2}(1-2\sigma)=\frac{1}{2}.
\end{eqnarray*}
\end{enumerate}
By summarizing the above inequalities, we have
\begin{eqnarray*}
r(n,t,\chi)\in \Big(\min\bigg\{\frac{1}{2},\max\Big\{\sigma,P_m(n,t,\chi)\Big\}\bigg\},\max\Big\{\frac{1}{2},P_m(n,t,\chi)\Big\}\Big].
\end{eqnarray*}
$\hfill\hfill\square$
\subsubsection{Lemma \ref{lemma:_adaptive_prob_sigma}.1 --- Lemma \ref{lemma:_adaptive_prob_sigma}.3}
We prove the three propositions of the lemma respectively.

\emph{Proof of Lemma \ref{lemma:_adaptive_prob_sigma}.1} Noting that $\sigma\le 1/2$ holds, we estimate the probability that in one generation the EA finds number of matching bits $N_{t}^{(P)}= j\in [n-n/\log^3 n,n]$:
\begin{eqnarray}
\nonumber&&\mathbb{P}\Bigg(N_{t}^{(P)}=j\mid N_{t-1}=i=o(n),N_{t-1}=i>\frac{n}{\log^2n},j\in \Big[n-\frac{n}{\log^3 n},n\Big], \sigma=\omega\Big(\frac{\log
n}{n}\Big)\Bigg)\\
\nonumber&=&\sum_{k=0}^{\min\{i,n-j\}}{n-i\choose j-i+k}{i\choose
k}\sigma^{j-i+2k}(1-\sigma)^{n-(j-i+2k)}\\
\nonumber&<&\sigma^{j-i}\sum_{k=0}^{\min\{i,n-j\}}{n-i\choose
n-j-k}{i\choose k}={n\choose n-j}\sigma^{j-i}\le n^{n-j}\Big(\frac{1}{2}\Big)^{j-i} \quad\quad\text{(by Lemma \ref{lemma:bi_coeffi})}\\
\nonumber&=& n^{o\big(\frac{n}{\log n}\big)}\Big(\frac{1}{2}\Big)^{n-o(n)}=2^{o(n)}\Big(\frac{1}{2}\Big)^{n-o(n)}\prec\frac{1}{SuperPoly(n)},
\end{eqnarray}
which is a super-polynomially small probability.

\emph{Proof of Lemma \ref{lemma:_adaptive_prob_sigma}.2}  The following probability can be estimated similarly as in the first case of the proof of Lemma \ref{lemma:_adaptive_prob_sigma}.1. For the sake of brevity, we provide the result directly:
\begin{eqnarray}
\nonumber&&\mathbb{P}\Bigg(N_{t}= j\mid \epsilon_2 n\le N_{t-1}=i\le\epsilon_1 n, j\in \Big[n-\frac{n}{\log^3 n},n\Big], \sigma=\omega\Big(\frac{\log n}{n}\Big)\Bigg)\prec\frac{1}{SuperPoly(n)},
\end{eqnarray}
which is a super-polynomially small probability.

\emph{Proof of Lemma \ref{lemma:_adaptive_prob_sigma}.3} We will prove this result by applying Chernoff bounds to the matching bits and non-matching bits respectively. After a DOP change, if the number of \emph{flipped} matching bits is no smaller than that of the \emph{flipped} non-matching bits, then the number of matching cannot increase after the DOP change.

According to the condition of Lemma \ref{lemma:_adaptive_prob_sigma}.3, $N_{t-1}=i=n-o(n)$ holds, thus the number of non-matching bits is $n-i=o(n)$. Let $D_{t}^+$ and $D_{t}^-$ be the numbers of flipped non-matching and matching bits after the DOP change at the beginning of the $t^{th}$ generation, respectively. According to Chernoff bounds, we have
\begin{eqnarray*}
&&\mathbb{P}\bigg(D_{t}^+>\frac{i\sigma}{4}\mid N_{t-1}=i=n-o(n), \sigma=\omega\Big(\frac{\log n}{n}\Big)\bigg)<\bigg(\frac{e}{c(n)}\bigg)^{\Theta(n\sigma)}=\bigg(\frac{e}{\omega(\log
n)}\bigg)^{\omega(\log n)}\\
&&\quad\quad\prec\frac{1}{SuperPoly(n)},
\end{eqnarray*}
where $c(n)$ is a polynomial function of the problem size $n$ that satisfies $c(n)=o(n\sigma)$ and $c(n)=\omega(1)$. On the other hand, for the number of flipped matching bits, we can also use Chernoff bounds:
\begin{eqnarray*}
\mathbb{P}\bigg(D_{t}^-<\frac{i\sigma}{2}\mid N_{t-1}=i=n-o(n), \sigma=\omega\Big(\frac{\log n}{n}\Big)\bigg)<e^{-\frac{(n-o(n))\sigma}{8}}=e^{-\omega(\log n)}\prec\frac{1}{SuperPoly(n)}.
\end{eqnarray*}
Thus, given the condition $N_{t-1}=i=i(n)=n-o(n)$ (where $i(n)$ is a function of the problem size $n$), the following probability is super-polynomially small: {\small
\begin{eqnarray*}
&&\mathbb{P}\bigg(D_{t}^++\frac{i\sigma}{4}>D_{t}^-\mid N_{t-1}=n-o(n),\sigma=\omega\Big(\frac{\log n}{n}\Big)
\bigg)\\
&<&\sum_{k=0}^{n-1}\mathbb{P}\bigg(D_{t}^-=k\mid N_{t-1}=n-o(n), \sigma=\omega\Big(\frac{\log n}{n}\Big)\bigg)\mathbb{P}\bigg(D_{t}^++\frac{i\sigma}{4}>k\mid N_{t-1}=n-o(n), \sigma=\omega\Big(\frac{\log
n}{n}\Big)\bigg)\\
&=&\sum_{k=0}^{\frac{i(n)\sigma}{2}-1}\mathbb{P}\bigg(D_{t}^-=k\mid N_{t-1}=n-o(n), \sigma=\omega\Big(\frac{\log n}{n}\Big)\bigg)\mathbb{P}\bigg(D_{t}^++\frac{i\sigma}{4}>k\mid N_{t-1}=n-o(n), \sigma=\omega\Big(\frac{\log
n}{n}\Big)\bigg)\\
&&+\sum_{k=\frac{i(n)\sigma}{2}}^{n-1}\mathbb{P}\bigg(D_{t}^-=k\mid N_{t-1}=n-o(n), \sigma=\omega\Big(\frac{\log n}{n}\Big)\bigg)\mathbb{P}\bigg(D_{t}^++\frac{i\sigma}{4}>k\mid N_{t-1}=n-o(n), \sigma=\omega\Big(\frac{\log
n}{n}\Big)\bigg)\\
&<&\frac{1}{SuperPoly(n)}\sum_{k=0}^{\frac{i(n)\sigma}{2}-1}\mathbb{P}\bigg(D_{t}^++\frac{i\sigma}{4}>k\mid N_{t-1}=n-o(n), \sigma=\omega\Big(\frac{\log
n}{n}\Big)\bigg)\\
&&\quad+\frac{1}{SuperPoly(n)}\sum_{k=\frac{i(n)\sigma}{2}}^{n-1}\mathbb{P}\bigg(D_{t}^-=k\mid N_{t-1}=n-o(n), \sigma=\omega\Big(\frac{\log
n}{n}\Big)\bigg)\\
&\prec&\frac{1}{SuperPoly(n)}
\end{eqnarray*}}
As a consequence,
\begin{eqnarray*}
&&\mathbb{P}\bigg(D_{t}^+-D_{t}^-> -\frac{i\sigma}{4}\mid N_{t-1}=n-o(n),\sigma=\omega\Big(\frac{\log n}{n}\Big) \bigg)\prec \frac{1}{SuperPoly(n)}
\end{eqnarray*}
holds. Combining the above inequality with the fact
\begin{eqnarray*}
N_{t}^{(P)}=N_{t-1}+D_{t}^+-D_{t}^-,
\end{eqnarray*}
we have proven Lemma \ref{lemma:_adaptive_prob_sigma}.3. $\hfill\square$

\subsubsection{Lemma \ref{lemma:_adaptive_prob_sigma}.4} In context of the details redefined in Lemma \ref{lemma:_adaptive_prob_sigma}.4, we prove the three propositions of the lemma respectively. In this proof, the number of matching bits of the $\chi^{th}$ offspring individual generated at the $t^{th}$ generation (i.e., $N_t^{(\chi)}$) will be considered.

\emph{Proof of Lemma \ref{lemma:_adaptive_prob_sigmapm}.1} We need to consider two different cases: $P_m(n,t,\chi)\le 1-h$ and $P_m(n,t,\chi)>1-h$, where $h\in(0,1)$ is a constant.

$-$ \emph{First Case:} $P_m(n,t,\chi)\le 1-h$ holds. Given the conditions that $\sigma=\omega(\log n/n)$ and $P_m(n,t,\chi)\le 1-h$, by applying Lemma \ref{lemma:two_bitwise} we obtain:
\begin{eqnarray*}
&&r(n,t,\chi)=\omega(\log n/n),\\
&&r(n,t,\chi)\le \max\Big\{\frac{1}{2},1-h\Big\}.
\end{eqnarray*}
By above inequalities, we estimate the probability that in one generation the EA finds number of matching bits $N_{t}^{(\chi)}= j\in [n-n/\log^3 n,n]$: {\small
\begin{eqnarray}
\nonumber&&\mathbb{P}\Bigg(N_{t}^{(\chi)}=j\mid N_{t-1}=i=o(n),N_{t-1}=i>\frac{n}{\log^2 n},j\in
\Big[n-\frac{n}{\log^3 n},n\Big], P_m(n,t,\chi)\le 1-h,\sigma=\omega\Big(\frac{\log n}{n}\Big)\Bigg)\\
\nonumber&=&\sum_{k=0}^{\min\{i,n-j\}}{n-i\choose j-i+k}{i\choose
k}r(n,t,\chi)^{j-i+2k}(1-r(n,t,\chi))^{n-(j-i+2k)}\\
\nonumber&<&r(n,t,\chi)^{j-i}\sum_{k=0}^{\min\{i,n-j\}}{n-i\choose
n-j-k}{i\choose k}={n\choose n-j}r(n,t,\chi)^{j-i}< n^{n-j}\max\Big\{\frac{1}{2},1-h\Big\}^{j-i}\quad\quad\text{(by Lemma \ref{lemma:bi_coeffi})}\\
\nonumber&=& n^{o\big(\frac{n}{\log n}\big)}\max\Big\{\frac{1}{2},1-h\Big\}^{n-o(n)}=2^{o(n)} \max\Big\{\frac{1}{2},1-h\Big\}^{n-o(n)}\prec\frac{1}{SuperPoly(n)},
\end{eqnarray}
} which is a super-polynomially small probability.

$-$ \emph{Second Case:} $P_m(n,t,\chi)>1-h$ holds. Given the condition that $\sigma=\omega(\log n/n)$ and $P_m(n,t,\chi)> 1-h$, by applying Lemma \ref{lemma:two_bitwise} we obtain:
\begin{eqnarray*}
r(n,t,\chi)>\min\bigg\{\frac{1}{2},\max\Big\{\sigma,P_m\Big\}\bigg\}\ge\min\bigg\{\frac{1}{2}, \max\Big\{\sigma,1-h\Big\}\bigg\}\ge\min\Big\{\frac{1}{2},1-h\Big\}.
\end{eqnarray*}
On the other hand, we must note the fact called symmetrical bitwise mapping: given the condition that the number of matching bits $N_{t-1}$ equals $i$ and the composite bitwise mapping rate $r(n,t,\chi)$, the consequence of the bitwise mapping is equivalent to that of the case in which $N_{t-1}$ equals $n-i$ and the composite bitwise mapping rate equals $1-r(n,t,\chi)$.

Formally, we have
\begin{eqnarray*}
1-r(n,t,\chi)<1-\min\Big\{\frac{1}{2},1-h\Big\}=\max\Big\{\frac{1}{2},h\Big\}.
\end{eqnarray*}
Noting the fact described above, we know $\forall i$ that satisfies $i=o(n)$ and $i>n/\log^2 n$, the following equation holds in response to the symmetrical bitwise mapping:
\begin{eqnarray*}
&&\mathbb{P}\Bigg(N_t^{(\chi)}=j\mid N_{t-1}=i,i=o(n),i>\frac{n}{\log^2 n},j\in \Big[n-\frac{n}{\log^3
n},n\Big], r(n,t,\chi)\Bigg)\\
&=&\mathbb{P}\Bigg(N_t^{(\chi)}=j\mid N_{t-1}^*=n-i,i=o(n),i>\frac{n}{\log^2 n},j\in \Big[n-\frac{n}{\log^3 n},n\Big], r^*(n,t,\chi)=1-r(n,t,\chi)\Bigg),
\end{eqnarray*}
where we use $r^*(n,t,\chi)$ to represent the \emph{notional} composite bitwise mapping rate with the value of $1-r(n,t,\chi)$, and $N_{t-1}^*$ to represent the \emph{notional} number of matching bits (found by the EA) at the end of the $(t-1)^{th}$ generation.

According to the value of $r^*(n,t,\chi)$, there are further two subcases for us to consider. In the first situation, $r^*(n,t,\chi)=O(\log n/n)$ holds. By Chernoff bounds, we know that with an overwhelming probability there are at most $\log^2 n$ flipped bits among the total $n$ bits:
\begin{eqnarray*}
\mathbb{P}\bigg(|N_t^{(\chi)}-N_{t-1}^*|< \log^2 n\mid r^*(n,t,\chi)=O\Big(\frac{\log n}{n}\Big)\bigg)>1-\bigg(\frac{e}{\Omega(\log n)}\bigg)^{\Omega(\log^2 n)}\succ 1-\frac{1}{SuperPoly(n)}.
\end{eqnarray*}
Consequently, with an overwhelming probability the number of matching bits will decrease or increase by at most $\log^2 n$ after the overall bitwise mapping. It follows from $N_{t-1}^*=n-i$, $i=o(n)$ and $i>n/\log^2 n$ that the above upper bound implies that $N_t^{(\chi)}<n-n/\log^2 n+\log^2n< n-n/\log^3 n$ and $N_t^{(\chi)}>n/\log^2 n$ hold with an overwhelming probability. In other words,
\begin{eqnarray*}
&&\mathbb{P}\Bigg(N_t^{(\chi)}=j\mid N_{t-1}^*=n-i,i=o(n),i>\frac{n}{\log^2 n},j\in \Big[n-\frac{n}{\log^3
n},n\Big], r^*(n,t,\chi)=O\Big(\frac{\log n}{n}\Big)\Bigg)\\
&&\quad\quad\prec \frac{1}{SuperPoly(n)}.
\end{eqnarray*}

In the second subcase, $r^*(n,t,\chi)=\omega(\log n/n)$ holds. In the proof of Lemma \ref{lemma:_adaptive_prob_sigmapm}.3, we will consider the case $r(n,t,\chi)=\omega(\log n/n)$. Since there is no essential difference between the proofs related to $r^*(n,t,\chi)$ and $r(n,t,\chi)$, we will not provide the proof here for the sake of brevity. For details, one can refer to the proof of Lemma \ref{lemma:_adaptive_prob_sigmapm}.3 below.

Combining the above two subcases together, we obtain that:
\begin{eqnarray*}
&&\mathbb{P}\Bigg(N_t^{(\chi)}=j\mid N_{t-1}=i,i=o(n),i>\frac{n}{\log^2 n},j\in \Big[n-\frac{n}{\log^3
n},n\Big], r(n,t,\chi)>\max\Big\{\frac{1}{2},1-h\Big\}\Bigg)\\
&=&\mathbb{P}\Bigg(N_t^{(\chi)}=j\mid N_{t-1}^*=n-i,i=o(n),i>\frac{n}{\log^2 n},j\in \Big[n-\frac{n}{\log^3
n},n\Big], r^*(n,t,\chi)<\max\Big\{\frac{1}{2},h\Big\}\Bigg)\\
&\prec& \frac{1}{SuperPoly(n)}.
\end{eqnarray*}
Thus we have finished the proof of the second case. Combining the first and second cases together, we have proven Lemma \ref{lemma:_adaptive_prob_sigmapm}.1.

\emph{Proof of Lemma \ref{lemma:_adaptive_prob_sigmapm}.2} For convenience, we omit it the index ``$(n,t)$'' in the proof, since the rest part of the proof are restricted in the $t^{th}$ generation only. The following probability can be estimated similarly as in the first case of the proof of Lemma \ref{lemma:_adaptive_prob_sigmapm}.1. For the sake of brevity, we provide the result directly:
\begin{eqnarray}
\nonumber&&\mathbb{P}\Bigg(N_{t+1}= j\mid \epsilon_2 n\le N_{t}=i\le\epsilon_1 n, j\in \Big[n-\frac{n}{\log^3 n},n\Big],
r=\omega\Big(\frac{\log n}{n}\Big)\Bigg)\\
\nonumber&=&\sum_{k=0}^{\min\{i,n-j\}}{n-i\choose j-i+k}{i\choose
k}r^{j-i+2k}(1-r)^{n-(j-i+2k)}\\
\nonumber&<&r^{j-i}(1-r)^{n-(j-i+2n/\log^3n)}\sum_{k=0}^{\min\{i,n-j\}}{n-i\choose
n-j-k}{i\choose k}\\
\nonumber&=&{n\choose n-j}r^{j-i}(1-r)^{n-(j-i+2n/\log^3n)}< n^{n-j}r^{j-i}(1-r)^{n-(j-i+2n/\log^3n)}\quad\quad\text{(by Lemma \ref{lemma:bi_coeffi})}\\
\nonumber&=& n^{o\big(\frac{n}{\log n}\big)}\omega\Big(\frac{\log n}{n}\Big)^{\Theta(n)}\bigg(1-\omega\Big(\frac{\log n}{n}\Big)\bigg)^{\Theta(n)}\prec\frac{1}{SuperPoly(n)},
\end{eqnarray}
which is a super-polynomially small probability. Thus we have proven Lemma \ref{lemma:_adaptive_prob_sigmapm}.2.

\emph{Proof of Lemma \ref{lemma:_adaptive_prob_sigmapm}.3} We will prove this result by applying Chernoff bounds to the matching bits and non-matching bits respectively. In one generation, if the number of \emph{flipped} matching bits is no smaller than that of the \emph{flipped} non-matching bits, then the number of matching cannot increase.

 According to the condition of Lemma \ref{lemma:_adaptive_prob_sigmapm}.3,
the number of matching bits at the $(t-1)^{th}$ generation satisfies $N_{t-1}=i=n-o(n)$, thus the number of non-matching bits is $n-i=o(n)$. Let $L_{t}^+$ and $L_{t}^-$ be the numbers of flipped non-matching and matching bits after the DOP change and the mutation at the $t^{th}$ generation, respectively. By Chernoff bounds, we have
\begin{eqnarray*}
&&\mathbb{P}\bigg(L_{t}^+>\frac{ir}{4}\mid N_{t-1}=i=n-o(n), r=\omega\Big(\frac{\log n}{n}\Big)\bigg)<\bigg(\frac{e}{c(n)}\bigg)^{\Theta(nr)}=\bigg(\frac{e}{\omega(\log
n)}\bigg)^{\omega(\log n)}\\
&&\quad\quad\prec\frac{1}{SuperPoly(n)},
\end{eqnarray*}
where $c(n)$ is a polynomial function of the problem size $n$ that satisfies $c(n)=o(npr)$ and $c(n)=\omega(1)$. On the other hand, for the number of flipped matching bits, we also apply Chernoff bounds:
\begin{eqnarray*}
\mathbb{P}\bigg(L_{t}^-<\frac{ir}{2}\mid N_{t-1}=i=n-o(n), r=\omega\Big(\frac{\log n}{n}\Big)\bigg)<e^{-\frac{(n-o(n))r}{8}}=e^{-\omega(\log n)}\prec\frac{1}{SuperPoly(n)}.
\end{eqnarray*}
Thus we know that, given the condition $N_{t}=i=i(n)=n-o(n)$ (where $i(n)$ is a function of the problem size $n$), the following probability is super-polynomially close to $0$: {\small
\begin{eqnarray*}
&&\mathbb{P}\bigg(L_{t}^++\frac{ir}{4}>L_{t}^-\mid N_{t-1}=n-o(n),r=\omega\Big(\frac{\log n}{n}\Big)
\bigg)\\
&<&\sum_{k=0}^{n-1}\mathbb{P}\bigg(L_{t}^-=k\mid N_{t-1}=n-o(n), r=\omega\Big(\frac{\log n}{n}\Big)\bigg)\mathbb{P}\bigg(L_{t}^++\frac{ir}{4}>k\mid N_{t-1}=n-o(n), r=\omega\Big(\frac{\log
n}{n}\Big)\bigg)\\
&=&\sum_{k=0}^{\frac{i(n)r}{2}-1}\mathbb{P}\bigg(L_{t}^-=k\mid N_{t-1}=n-o(n), r=\omega\Big(\frac{\log n}{n}\Big)\bigg)\mathbb{P}\bigg(L_{t+1}^++\frac{ir}{4}>k\mid N_{t-1}=n-o(n), r=\omega\Big(\frac{\log
n}{n}\Big)\bigg)\\
&&+\sum_{k=\frac{i(n)r}{2}}^{n-1}\mathbb{P}\bigg(L_{t}^-=k\mid N_{t-1}=n-o(n), r=\omega\Big(\frac{\log n}{n}\Big)\bigg)\mathbb{P}\bigg(L_{t}^++\frac{ir}{4}>k\mid N_{t-1}=n-o(n), r=\omega\Big(\frac{\log
n}{n}\Big)\bigg)\\
&<&\frac{1}{SuperPoly(n)}\sum_{k=0}^{\frac{i(n)r}{2}-1}\mathbb{P}\bigg(L_{t}^++\frac{ir}{4}>k\mid N_{t-1}=n-o(n), r=\omega\Big(\frac{\log
n}{n}\Big)\bigg)\\
&&\quad+\frac{1}{SuperPoly(n)}\sum_{k=\frac{i(n)r}{2}}^{n-1}\mathbb{P}\bigg(L_{t}^-=k\mid N_{t-1}=n-o(n), r=\omega\Big(\frac{\log
n}{n}\Big)\bigg)\\
&\prec&\frac{1}{SuperPoly(n)}
\end{eqnarray*}}
Noting that
\begin{eqnarray*}
N_t^{(\chi)}=N_{t-1}+L_{t}^+-L_{t}^-,
\end{eqnarray*}
we have proven Lemma \ref{lemma:_adaptive_prob_sigmapm}.3. $\hfill\square$

\subsection{Proof of Lemma \ref{lemma:_adaptive_prob_sigmapm_sym}.1}
Lemma \ref{lemma:_adaptive_prob_sigmapm_sym}.1 has two versions. The first version presents a general result for any BDOP whose shifting rate satisfies $\sigma=\omega(\log n/n)$ and $\sigma\in (0,1/2]$, where the time-variable mutation rate should be smaller than $1-1/\log n$. The second version is for the \textsc{BitMatching}$_D$ problem whose shifting rate satisfies $\sigma=\omega(\log n/n)$ and $\sigma\in (0,1/2]$, where the time-variable mutation rate can take any value between $0$ and $1$.
\subsubsection{General result for BDOP Class} Let
us first study $N_{t}^{(P)}$ under the conditions $N_{t-1}=n-o(n)$ and $N_{t-1}<n-\frac{n}{\log^3n}$. According to Chernoff bounds, we know that with an overwhelming probability there are at most $3n/4$ flipped bits among the total $n$ bits after the DOP change at the beginning of the $t^{th}$ generation:
\begin{eqnarray*}
\mathbb{P}\bigg(|N_{t}^{(P)}-N_{t-1}|< \frac{3}{4}n\mid \sigma\in \Big(0,\frac{1}{2}\Big]\bigg)>1-e^{-n/24}\succ 1-\frac{1}{SuperPoly(n)}.
\end{eqnarray*}
Noting that the range between $n/\log ^2n$ and $n-o(n)$ is much larger than $3n/4$ (i.e., $n-o(n)-n/\log^2n>3n/4$), we know that the probability of $N_t^{(P)}\in\mathbb{L}_1$, conditional on $N_{t-1}=n-o(n)$ and $N_{t-1}<n-n/\log^3n$, is super-polynomially close to $0$. Formally,
\begin{eqnarray}\label{equation:parent_fail_to_longjump}
&&\mathbb{P}\bigg(N_t^{(P)}\le \frac{n}{\log^2n}\mid N_{t-1}=n-o(n),N_{t-1}<n-\frac{n}{\log^3n}\bigg)\prec\frac{1}{SuperPoly(n)}.
\end{eqnarray}
Thus the original version of Lemma \ref{lemma:_adaptive_prob_sigmapm_sym}.1 for the general BDOP class is proven.

Let us further consider the proof when $N_t^{(P)}$ is replaced by $N_t^{(\chi)}$ in Lemma \ref{lemma:_adaptive_prob_sigmapm_sym}.1. It follows from the two conditions $\sigma=\omega(\log n/n)$ and $P_m(n,t,\chi)\le 1-1/\log n$ that $r(n,t,\chi)\le 1-1/\log n$. According to Lemma \ref{lemma:two_bitwise}, we estimate the probability that $N_t^{(\chi)}\in\mathbb{L}_1$, conditional on $N_{t-1}\in\mathbb{A}_2\cup\mathbb{A}_1$ and $r(n,t,\chi)\le 1-1/\log n$. Let $L_{t}^-$ be the number of flipped matching bits after the DOP change and the mutation at the $t^{th}$ generation, we have
\begin{eqnarray*}
&&\mathbb{P}\bigg(N_t^{(\chi)}\in\mathbb{L}_1\mid N_{t-1}=n-o(n),N_{t-1}<n-\frac{n}{\log^3n}, r(n,t,\chi)\le 1-\frac{1}{\log n}
\bigg)\\
&=&\mathbb{P}\bigg( N_t^{(\chi)}\le \frac{n}{\log^2 n}\mid N_{t-1}=n-o(n),N_{t-1}<n-\frac{n}{\log^3n}, r(n,t,\chi)\le 1-\frac{1}{\log n}
\bigg)\\
&\le&\mathbb{P}\bigg(n-\Big(L_{t}^-+(n-N_{t-1})\Big)\le\frac{n}{\log^2 n}\mid N_{t-1}=n-o(n),N_{t-1}<n-\frac{n}{\log^3n}, r(n,t,\chi)\le 1-\frac{1}{\log n}
\bigg)\\
&=&\mathbb{P}\bigg(L_{t}^-\ge N_{t-1}-\frac{n}{\log^2n}\mid N_{t-1}=n-o(n),N_{t-1}<n-\frac{n}{\log^3n}, r(n,t,\chi)\le 1-\frac{1}{\log n} \bigg).
\end{eqnarray*}

On one hand, if $r(n,t,\chi)>\frac{1}{4e}$, then by Chernoff bound we further have
\begin{eqnarray}
\nonumber&&\mathbb{P}\bigg(N_t^{(\chi)}\in\mathbb{L}_1\mid N_{t-1}=n-o(n),N_{t-1}<n-\frac{n}{\log^3n}, \frac{1}{4e}<r(n,t,\chi)\le
1-\frac{1}{\log n} \bigg)\\
\nonumber&\le&\mathbb{P}\bigg(L_{t}^-\ge N_{t-1}-\frac{n}{\log^2n}=(1+\rho_1)\hat{\mathbb{E}}[L_{t}^-]\mid N_{t-1}=n-o(n),N_{t-1}<n-\frac{n}{\log^3n}, \frac{1}{4e}<r(n,t,\chi)\le 1-\frac{1}{\log n}
\bigg)\\
\nonumber&=&\mathbb{P}\bigg(L_{t}^-\ge(1+\rho_1)\hat{\mathbb{E}}[L_{t}^-]\mid N_{t-1}=n-o(n),N_{t-1}<n-\frac{n}{\log^3n}, \frac{1}{4e}<r(n,t,\chi)\le 1-\frac{1}{\log n}
\bigg)\\
\label{equation:r_large_lemma}&<&e^{-\hat{\mathbb{E}}[L_{t}^-]\rho_1^2/4}\prec\frac{1}{SuperPoly(n)},
\end{eqnarray}
where $\rho_1=(N_{t-1}-n/\log^2n)/(N_{t-1}-N_{t-1}/\log n)-1=\Theta(1)$ (by $N_{t-1}=n-o(n)$), and $\hat{\mathbb{E}}[L_{t}^-]=\mathbb{E}[L_{t}^-\mid N_{t-1}=n-o(n),N_{t-1}<n-n/\log^3n, 1/4e<r(n,t,\chi)\le 1-1/\log n]=N_{t-1}r(n,t,\chi)=\Theta(n)$ (by $N_{t-1}=n-o(n)$ and $r(n,t,\chi)>1/4e$).

On the other hand, let us consider the case $r(n,t,\chi)\le\frac{1}{4e}$. By Chernoff bound, we have
\begin{eqnarray}
\nonumber&&\mathbb{P}\bigg(N_t^{(\chi)}\in\mathbb{L}_1\mid N_{t-1}=n-o(n),N_{t-1}<n-\frac{n}{\log^3n}, r(n,t,\chi)\le \frac{1}{4e}
\bigg)\\
\nonumber&=&\mathbb{P}\bigg(L_{t}^-\ge N_{t-1}-\frac{n}{\log^2n}=(1+\rho_2)\bar{\mathbb{E}}[L_{t}^-]\mid N_{t-1}=n-o(n),N_{t-1}<n-\frac{n}{\log^3n}, r(n,t,\chi)\le \frac{1}{4e}
\bigg)\\
\label{equation:r_small_lemma}&<&\Big(\frac{e}{1+\rho_2}\Big)^{(1+\rho_2)\bar{\mathbb{E}}[L_{t}^-]} =\Big(\frac{e}{1+\rho_2}\Big)^{N_{t-1}-\frac{n}{\log^2n}}\prec\frac{1}{SuperPoly(n)},
\end{eqnarray}
where $\bar{\mathbb{E}}[L_{t}^-]=\mathbb{E}[L_{t}^-\mid N_{t-1}=n-o(n),N_{t-1}<n-n/\log^3n, r(n,t,\chi)\le 1/4e]\le n/4e$ and $\rho_2=(N_{t-1}-n/\log^2n)/\bar{\mathbb{E}}[L_{t}^-]-1>(n/2)/(n/4e)-1\ge 2e-1$. Thus we have proven Lemmas \ref{lemma:_adaptive_prob_sigmapm_sym}.1.

\subsubsection{Specific result for \textsc{BitMatching}$_D$}
By Chernoff bounds, we know that with an overwhelming probability there are at most $3n/4$ flipped bits among the total $n$ bits after the DOP change at the beginning of the $t^{th}$ generation:
\begin{eqnarray}\label{equation:_lemma_fact}
\mathbb{P}\bigg(|N_{t}^{(P)}-N_{t-1}|< \frac{3}{4}n\mid \sigma\in \Big(0,\frac{1}{2}\Big]\bigg)>1-e^{-n/24}\succ 1-\frac{1}{SuperPoly(n)}.
\end{eqnarray}
Consequently, with an overwhelming probability, the number of matching bits will decrease or increase by at most $3n/4$ after the DOP change. Recall that $N_{t-1}=i=n-o(n)$ is one of the conditions of Lemma \ref{lemma:_adaptive_prob_sigma_sym}.1, we know that $N_{t}^{(P)}>n/\log^2 n$ holds with an overwhelming probability. Thus we have proven the original version of Lemma \ref{lemma:_adaptive_prob_sigma_sym}.1.

Meanwhile, let us consider the alternative version of Lemma \ref{lemma:_adaptive_prob_sigma_sym}.1, where $N_{t}^{(P)}$ is replaced by $N_t^{(\chi)}$. Since the EA always preserves the one with better fitness between the parent and offspring individuals, we know that $N_t\ge N_{t}^{(P)}$ always holds given the \textsc{BitMatching}$_D$ problem. Combining the above fact with Eq. \ref{equation:_lemma_fact}, we obtain the alternative version of
 Lemma \ref{lemma:_adaptive_prob_sigmapm_sym}.1. $\hfill\square$

\section{Proofs of Lemmas \ref{lemma:_generalization_large_sigma} and \ref{lemma:large_sigma}, and Theorems \ref{theorem:large_sigma_1plus_lambda} and \ref{theorem:large_sigma_1plus_lambda_bitmatching}}

\subsection{Lemmas \ref{lemma:_generalization_large_sigma} and \ref{lemma:large_sigma}}
The only difference between the proofs of Lemmas \ref{lemma:_generalization_large_sigma} and \ref{lemma:large_sigma} is that the former utilizes the original version of Lemma \ref{lemma:_adaptive_prob_sigmapm_sym} for the general BDOP class, while the latter utilizes the specific version of Lemma \ref{lemma:_adaptive_prob_sigmapm_sym} for the \textsc{BitMatching}$_D$ problem. Hence, we only provide a unified proof for the sake of brevity. As mentioned in Section \ref{sec:small_sigma}, the proof contains the analysis related to Propositions A\ref{lemma:_generalization_large_sigma}.1, A\ref{lemma:_generalization_large_sigma}.2, and A\ref{lemma:_generalization_large_sigma}.3.

Here we study the above propositions one after another.

\textbf{Analysis of Proposition A\ref{lemma:_generalization_large_sigma}.1}. As the first step, we prove that the initial number of matching bits satisfies that $N_{0}^{(P)}\in \mathbb{O}_1$ holds with an overwhelming probability. Since the initial individual is generated randomly by the uniform distribution, we estimate the following two probabilities by Chernoff bounds (Lemma \ref{chernoff}):
\begin{eqnarray*}
&\mathbb{P}\Big(N_{0}^{(P)}<\frac{1}{4}n\Big)&<e^{-n/16}\prec
\frac{1}{SuperPoly(n)},\\
&\mathbb{P}\Big(N_{0}^{(P)}>\frac{3}{4}n\Big)&<\Big(\frac{8e}{27}\Big)^{n/4}\prec \frac{1}{SuperPoly(n)},
\end{eqnarray*}
In other words, $N_{0}^{(P)}\in [n/4,3n/4]$ (where $[n/4,3n/4]\subset \mathbb{O}_1$) holds with an overwhelming probability.

Given the condition $N_{0}^{(P)}\in [n/4,3n/4]$, we now prove that the probability of $N_{0}^{(O)}\in \mathbb{F}_1\cup \mathbb{L}_1$ is super-polynomially close to $0$. According to the mutation rate at the $0^{th}$ generation $P_m(n,0)$, there are two cases:

$-$ First case ($P_m(n,0)=\omega(\log n/n)$): According to Lemmas \ref{lemma:_adaptive_prob_pm}.2 and \ref{lemma:_adaptive_prob_pm_sym}.1, we know that:
\begin{eqnarray*}
\mathbb{P}\bigg(N_{0}^{(O)}\in \mathbb{F}_1\cup \mathbb{L}_1\mid N_{0}^{(P)}\in \Big[\frac{n}{4},\frac{3n}{4}\Big], P_m(n,0)=\omega\Big(\frac{\log n}{n}\Big)\bigg)\prec \frac{1}{SuperPoly(n)}.
\end{eqnarray*}

$-$ Second case ($P_m(n,0)=O(\log n/n)$): By Chernoff bounds, we know that with an overwhelming probability there are at most $\log^2 n$ flipped bits among the total $n$ bits after mutation (which implies that the number of matching bits can decrease or increase by at most $\log^2 n$ after mutation):
\begin{eqnarray*}
&&\mathbb{P}\bigg(|N_{0}^{(O)}-N_{0}^{(P)}|< \log^2 n\mid N_{0}^{(P)}\in \Big[\frac{n}{4},\frac{3n}{4}\Big],
P_m(n,0)=O\Big(\frac{\log n}{n}\Big)\bigg)\\
&>&1-\bigg(\frac{e}{\Omega(\log n)}\bigg)^{\Omega(\log^2 n)}\succ 1-\frac{1}{SuperPoly(n)}.
\end{eqnarray*}
In other words, $N_{0}^{(O)}\in [\log^2 n+ n/4,\log^2 n+3n/4]$ holds with an overwhelming probability (given the condition that $N_{0}^{(P)}\in [n/4,3n/4], P_m(n,0)=O(\log n/n)$). It then follows from $N_{0}^{(O)}\in [\log^2 n+ n/4,\log^2 n+3n/4]\nsubseteq \mathbb{F}_1\cup \mathbb{L}_1$ that $N_{0}^{(O)}\notin \mathbb{F}_1\cup \mathbb{L}_1$ holds with an overwhelming probability (given the conditions that $N_{0}^{(P)}\in [n/4,3n/4]$ and $P_m(n,0)=O(\log n/n)$).

Combining the above facts for $N_{0}^{(P)}$ and $N_{0}^{(O)}$, we obtain
\begin{eqnarray}\label{equation:prob_0}
\mathbb{P}(N_{0}\in \mathbb{F}_1\cup \mathbb{L}_1)\prec \frac{1}{SuperPoly(n)}.
\end{eqnarray}

\textbf{Analysis of Proposition A\ref{lemma:_generalization_large_sigma}.2}. Given the condition $N_{t-1}\in \mathbb{O}_1$ ($t\in\mathbb{N}^+$), we now prove that the probability of the event $\mathbb{O}_1\xrightarrow[]{C_t}\mathbb{F}_1\cup \mathbb{L}_1$ is super-polynomially close to $0$. The condition $N_{t-1}\in \mathbb{O}_1$ leads to one of the following cases, and Lemmas \ref{lemma:_adaptive_prob_sigma} and \ref{lemma:_adaptive_prob_sigma_sym} provide the corresponding probability results:
\begin{enumerate}
\item[] \textbf{Case i [\emph{$N_{t-1}=o(n)$ and $N_{t-1}>n/\log^2 n$ both hold}]}: By Lemmas \ref{lemma:_adaptive_prob_sigma}.1
and \ref{lemma:_adaptive_prob_sigma_sym}.3, the probability of the event $N_t^{(P)}\in\mathbb{F}_1\cup \mathbb{L}_1$, conditional on $N_{t-1}=o(n)$ and $N_{t-1}>n/\log^2 n$, is super-polynomially close to $0$;

\item[] \textbf{Case ii [\emph{$\exists$ constants $\epsilon_1$ and $\epsilon_2$ such that
$0<\epsilon_2\le\epsilon_1<1$ and $\epsilon_2 n\le N_{t-1}=i\le\epsilon_1 n$}]}: By Lemmas \ref{lemma:_adaptive_prob_sigma}.2 and \ref{lemma:_adaptive_prob_sigma_sym}.2, the probability of the event $N_t^{(P)}\in\mathbb{F}_1\cup \mathbb{L}_1$, conditional on the event that $\exists$ constants $\epsilon_1$ and $\epsilon_2$ such that $0<\epsilon_2\le\epsilon_1<1$ and $\epsilon_2 n\le N_{t-1}=i\le\epsilon_1 n$, is super-polynomially close to $0$;

\item[] \textbf{Case iii [\emph{$N_{t-1}=n-o(n)$ and $N_{t-1}<n-n/\log^3 n$ both hold}]}: By Lemma
\ref{lemma:_adaptive_prob_sigma}.3 and \ref{lemma:_adaptive_prob_sigma_sym}.1, the probability of the event $N_t^{(P)}\in\mathbb{F}_1\cup \mathbb{L}_1$, conditional on $N_{t-1}=n-o(n)$ and $N_{t-1}<n-n/\log^3 n$, is super-polynomially close to $0$.
\end{enumerate}
By summarizing the above results, we have:
\begin{eqnarray}\label{equation:one_bitwise_prob}
\mathbb{P}\Big(N_t^{(P)}\in\mathbb{F}_1\cup \mathbb{L}_1\mid N_{t-1}\in \mathbb{O}_1\Big)\prec \frac{1}{SuperPoly(n)}.
\end{eqnarray}

\textbf{Analysis of Proposition A\ref{lemma:_generalization_large_sigma}.3}. Next we aim at proving that the joint probability of the events $N_t^{(P)}\in\mathbb{O}_1$ and $N_t^{(O)}\in\mathbb{F}_1\cup\mathbb{L}_1$, conditional on $N_{t-1}\in \mathbb{O}_1$, is super-polynomially close to $0$. The above proposition is a direct corollary of the following inequality:
\begin{eqnarray*}
\mathbb{P}\Big(N_t^{(P)}\in\mathbb{F}_1\cup\mathbb{L}_1,N_t^{(O)}\in\mathbb{F}_1\cup\mathbb{L}_1\mid N_{t-1}\in \mathbb{O}_1\Big)+\mathbb{P}\Big(N_t^{(P)}\in\mathbb{O}_1,N_t^{(O)}\in\mathbb{F}_1\cup\mathbb{L}_1\mid N_{t-1}\in \mathbb{O}_1\Big)\prec \frac{1}{SuperPoly(n)}.
\end{eqnarray*}
Meanwhile, the above inequality is equivalent to the following inequality:
\begin{eqnarray}\label{equation:two_bitwise_porb}
\mathbb{P}\Big(N_t^{(O)}\in\mathbb{F}_1\cup\mathbb{L}_1\mid N_{t-1}\in \mathbb{O}_1\Big)\prec \frac{1}{SuperPoly(n)}.
\end{eqnarray}
Hence, to prove that Proposition A\ref{lemma:_generalization_large_sigma}.3, we only need to prove Eq. \ref{equation:two_bitwise_porb}.

As we know, the condition $N_{t-1}\in \mathbb{O}_1$ always implies three potential cases, and Lemma \ref{lemma:_adaptive_prob_sigmapm} provides the corresponding probability results:
\begin{enumerate}
\item[] \textbf{Case i [\emph{$N_{t-1}=o(n)$ and $N_{t-1}>n/\log^2 n$ both hold}]}: By
Lemmas \ref{lemma:_adaptive_prob_sigmapm}.1 and \ref{lemma:_adaptive_prob_sigmapm}.4 \footnote{For the $(1+1)$ EA, there is only one offspring individual at each generation, thus $N_t^{(O)}=N_t^{(1)}$.}, the probability of $N_t^{(O)}\in\mathbb{F}_1$, conditional on $N_{t-1}=o(n)$ and $N_{t-1}>n/\log^2 n$, is super-polynomially close to $0$; By Lemmas \ref{lemma:_adaptive_prob_sigmapm_sym}.3 and \ref{lemma:_adaptive_prob_sigmapm_sym}.4, the probability of $N_t^{(O)}\in\mathbb{L}_1$, conditional on $N_{t-1}=o(n)$, is super-polynomially close to $0$.

\item[] \textbf{Case ii [\emph{$\exists$ constants $\epsilon_1$ and $\epsilon_2$ such that
$0<\epsilon_2\le\epsilon_1<1$ and $\epsilon_2 n\le N_{t-1}=i\le\epsilon_1 n$}]}: By Lemmas \ref{lemma:_adaptive_prob_sigmapm}.2 and \ref{lemma:_adaptive_prob_sigmapm}.4, the probability of $N_t^{(O)}\in\mathbb{F}_1$, conditional on the event that $\exists$ constants $\epsilon_1$ and $\epsilon_2$ such that $0<\epsilon_2\le\epsilon_1<1$ and $\epsilon_2 n\le N_{t-1}=i\le\epsilon_1 n$, is super-polynomially close to $0$; By Lemmas \ref{lemma:_adaptive_prob_sigmapm_sym}.2 and \ref{lemma:_adaptive_prob_sigmapm_sym}.4, the probability of the event $N_t^{(O)}\in\mathbb{L}_1$, conditional on the event that $\exists$ constants $\epsilon_1$ and $\epsilon_2$ such that $0<\epsilon_2\le\epsilon_1<1$ and $\epsilon_2 n\le N_{t-1}=i\le\epsilon_1 n$, is super-polynomially close to $0$.

\item[] \textbf{Case iii [\emph{$N_{t-1}=n-o(n)$ and $N_{t-1}<n-n/\log^3 n$ hold}]}: By Lemmas
\ref{lemma:_adaptive_prob_sigma}.3 and \ref{lemma:_adaptive_prob_sigma}.4, the probability of the event $N_t^{(O)}\in\mathbb{F}_1$, conditional on $N_{t-1}=n-o(n)$ and $N_{t-1}<n-n/\log^3 n$, is super-polynomially close to $0$; By Lemmas \ref{lemma:_adaptive_prob_sigmapm_sym}.1 and \ref{lemma:_adaptive_prob_sigmapm_sym}.4, the probability of the event $N_t^{(O)}\in\mathbb{L}_1$, conditional on $N_{t-1}=n-o(n)$ and $N_{t-1}<n-n/\log^3n$, is super-polynomially close to $0$.
\end{enumerate}

By summarizing the above results, we can obtain Eq. \ref{equation:two_bitwise_porb}. Hence we have proven that:
\begin{eqnarray}\label{equation:two_bitwise_porb_sketch}
\mathbb{P}\Big(N_t^{(P)}\in\mathbb{O}_1,N_t^{(O)}\in\mathbb{F}_1\cup\mathbb{L}_1\mid N_{t-1}\in \mathbb{O}_1\Big)\prec \frac{1}{SuperPoly(n)}.
\end{eqnarray}

\textbf{Conclusion.} Combining the probabilities described in Eqs. \ref{equation:one_bitwise_prob}, and \ref{equation:two_bitwise_porb_sketch} together, we have
\begin{eqnarray*}
&&\mathbb{P}\Big(N_{t}\in \mathbb{F}_1\cup\mathbb{L}_1\mid N_{t-1}\in
\mathbb{O}_1\Big)\\
&\le&\mathbb{P}\Big(N_{t}^{(P)}\in\mathbb{F}_1\cup \mathbb{L}_1\mid N_{t-1}\in \mathbb{O}_1\Big) + \mathbb{P}(N_{t}^{(P)}\in\mathbb{O}_1, N_{t}^{(O)}\in\mathbb{F}_1\cup \mathbb{L}_1\mid N_{t-1}\in \mathbb{O}_1)\prec \frac{1}{SuperPoly(n)}
\end{eqnarray*}
holds $\forall t\in\mathbb{N}^+$. Let $\tau_{\mathbb{F}_1}$ be defined as the first hitting time to the interval $\mathbb{F}_1$, formally,
\begin{eqnarray}
\tau_{\mathbb{F}_1}= \min\Big\{t\ge0; (N^{(P)}_t\in \mathbb{F}_1)\vee(N^{(O)}_t\in \mathbb{F}_1)\Big\}.
\end{eqnarray}
Meanwhile, $\forall t\in\mathbb{N}^+$ we have:
\begin{eqnarray*}
&&\mathbb{P}(\tau_{\mathbb{F}_1}=t)<\mathbb{P}(N_{t}\in \mathbb{F}_1\cup\mathbb{L}_1, N_{t-1}\in
\mathbb{O}_1,\dots,N_0\in \mathbb{O}_1)\\
&=&\mathbb{P}(N_{t}\in\mathbb{F}_1\cup\mathbb{L}_1 \mid N_{t-1}\in
\mathbb{O}_1,\dots,N_0\in \mathbb{O}_1)\cdot\mathbb{P}(N_{t-1}\in \mathbb{O}_1,\dots,N_0\in \mathbb{O}_1)\\
&=&\mathbb{P}(N_{t}\in\mathbb{F}_1\cup\mathbb{L}_1 \mid N_{t-1}\in
\mathbb{O}_1)\cdot\mathbb{P}(N_{t-1}\in \mathbb{O}_1,\dots,N_0\in \mathbb{O}_1)\\
&\le& \mathbb{P}(N_{t}\in\mathbb{F}_1\cup\mathbb{L}_1\mid N_{t-1}\in \mathbb{O}_1)\prec \frac{1}{SuperPoly(n)}.
\end{eqnarray*}
Moreover, Eq. \ref{equation:prob_0} implies $\mathbb{P}(\tau_{\mathbb{F}_1}=0)\prec 1/SuperPoly(n)$. Consequently, $\forall g(n)\prec Poly(n)$, the first hitting time to the target interval $\mathbb{F}_1$, which is denoted by $\tau_{\mathbb{F}_1}$, satisfies that
\begin{eqnarray*}
&&\mathbb{P}\big(\tau_{\mathbb{F}_1}\le g(n)\big)=\sum_{t=0}^{g(n)}\mathbb{P}(\tau_{\mathbb{F}_1}=t) \le \big(g(n)+1\big)\max_{t\le g(n)}\big\{\mathbb{P}(\tau_{\mathbb{F}_1}=t)\big\}\prec \frac{Poly(n)}{SuperPoly(n)}\sim \frac{1}{SuperPoly(n)}.
\end{eqnarray*}
In other words, the probability of $\tau_{\mathbb{F}_1}\prec Poly(n)$ is super-polynomially close to $0$. Consequently, we have proven Lemmas \ref{lemma:_generalization_large_sigma} and \ref{lemma:large_sigma}.
 $\hfill\square$
\subsection{Theorems \ref{theorem:large_sigma_1plus_lambda} and \ref{theorem:large_sigma_1plus_lambda_bitmatching}}
The proof idea of Theorem \ref{theorem:large_sigma_1plus_lambda} is quite similar to that of Lemma \ref{lemma:_generalization_large_sigma}. After updating the definition of $N_t^{(O)}$ by $\max\{N_t^{(1)},\dots,N_t^{(\lambda)}\}$ in response to the multiple-offspring strategy employed by the $(1+\lambda)$ EA, the main difference between the two proofs is that the transition events with respect to every generation of the $(1+\lambda)$ EA must involve $\lambda$ mutations generating $\lambda$ different offspring individuals, while those of the $(1+1)$ EA only consider a single mutation leading to a unique offspring individual. Meanwhile, the definition of $\tau_{\mathbb{F}_1}$ must be updated accordingly
\begin{eqnarray}
\tau_{\mathbb{F}_1}= \min\left\{t\ge0; \left(N^{(P)}_t\in \mathbb{F}_1\right)\vee\left(N^{(1)}_t\in \mathbb{F}_1\right)\vee\dots\vee\left(N^{(\lambda)}_t\in \mathbb{F}_1\right)\right\},
\end{eqnarray}
where $N^{(1)}_t,\dots,N^{(\lambda)}_t$ are number of matching bits with respect to the $\lambda$ offspring individuals at the $t^{th}$ generation respectively.

The proof of Theorem \ref{theorem:large_sigma_1plus_lambda} can be obtained by replacing the probability propositions related to a unique mutation with probability propositions related to $\lambda$ mutations. The probability propositions, demonstrated by Propositions A\ref{lemma:_generalization_large_sigma}.2, A\ref{lemma:_generalization_large_sigma}.2, and A\ref{lemma:_generalization_large_sigma}.3, are all about probabilities that are super-polynomially close to $0$, and similar arguments will also hold for the $(1+\lambda)$ EA, since the offspring size $\lambda$ is polynomial of $n$ and will not increase a super-polynomially small probability (i.e., $1/SuperPoly(n)$) to a polynomially large probability (i.e., $1/Poly(n)$).

For the sake of brevity, we do not provide in detail the proof of Theorem \ref{theorem:large_sigma_1plus_lambda} here. Instead, to illustrate the above proof idea, we only prove a probability proposition for $(1+\lambda)$ EA as an instance. The probability proposition, which is an important step towards proving Theorem \ref{theorem:large_sigma_1plus_lambda}, is the same to Proposition A\ref{lemma:_generalization_large_sigma}.3 except that $N_t^{(O)}$ is redefined as $\max\{N_t^{(1)},\dots,N_t^{(\lambda)}\}$ in response to the multiple-offspring strategy:
$$\forall t\in\mathbb{N}^+: \mathbb{P}(N_{t}^{(P)}\in\mathbb{O}_1, N_{t}^{(O)}\in\mathbb{F}_1\cup \mathbb{L}_1\mid N_{t-1}\in \mathbb{O}_1)\prec 1/SuperPoly(n).$$

To prove the above proposition, we only need to prove the following inequality:
\begin{eqnarray*}
\mathbb{P}\Big(N_t^{(P)}\in\mathbb{F}_1\cup\mathbb{L}_1,N_t^{(O)}\in\mathbb{F}_1\cup\mathbb{L}_1\mid N_{t-1}\in \mathbb{O}_1\Big)+\mathbb{P}\Big(N_t^{(P)}\in\mathbb{O}_1,N_t^{(O)}\in\mathbb{F}_1\cup\mathbb{L}_1\mid N_{t-1}\in \mathbb{O}_1\Big)\prec \frac{1}{SuperPoly(n)},
\end{eqnarray*}
which is equivalent to the following inequality:
\begin{eqnarray*}
\mathbb{P}\Big(N_t^{(O)}\in\mathbb{F}_1\cup\mathbb{L}_1\mid N_{t-1}\in \mathbb{O}_1\Big)\prec \frac{1}{SuperPoly(n)}.
\end{eqnarray*}

As we have done in ``\emph{Analysis of Proposition A\ref{lemma:_generalization_large_sigma}.3}'', the condition $N_{t-1}\in \mathbb{O}_1$ can be divided into three potential cases, and Lemma \ref{lemma:_adaptive_prob_sigmapm} provides the corresponding probability results:
\begin{enumerate}
\item[] \textbf{Case i [\emph{$N_{t-1}=o(n)$ and $N_{t-1}>n/\log^2 n$ both hold}]}: By
Lemmas \ref{lemma:_adaptive_prob_sigmapm}.1 and \ref{lemma:_adaptive_prob_sigmapm}.4, the probability of $N_t^{(\chi)}\in\mathbb{F}_1$ ($\forall \chi\in\{1,\dots,\lambda\}$), conditional on $N_{t-1}=o(n)$ and $N_{t-1}>n/\log^2 n$, is super-polynomially close to $0$. Adding up such probabilities with respect to $\lambda$ ($\lambda$ is polynomial in $n$) different offspring individuals (i.e., different $\chi$) yields an upper bound for the probability of $N_t^{(O)}\in\mathbb{F}_1$ conditional on $N_{t-1}=o(n)$ and $N_{t-1}>n/\log^2 n$, which is still super-polynomially close to $0$.

By Lemmas \ref{lemma:_adaptive_prob_sigmapm_sym}.3 and \ref{lemma:_adaptive_prob_sigmapm_sym}.4, the probability of $N_t^{(\chi)}\in\mathbb{L}_1$, conditional on $N_{t-1}=o(n)$, is super-polynomially close to $0$. Adding up such probabilities with respect to $\lambda$ ($\lambda$ is polynomial in $n$) different offspring individuals yields an upper bound for the probability of $N_t^{(O)}\in\mathbb{L}_1$ conditional on $N_{t-1}=o(n)$ and $N_{t-1}>n/\log^2 n$, which is still super-polynomially close to $0$.

\item[] \textbf{Case ii [\emph{$\exists$ constants $\epsilon_1$ and $\epsilon_2$ such that
$0<\epsilon_2\le\epsilon_1<1$ and $\epsilon_2 n\le N_{t-1}=i\le\epsilon_1 n$}]}: By Lemmas \ref{lemma:_adaptive_prob_sigmapm}.2 and \ref{lemma:_adaptive_prob_sigmapm}.4, the probability of $N_t^{(\chi)}\in\mathbb{F}_1$ ($\forall \chi\in\{1,\dots,\lambda\}$), conditional on the event that $\exists$ constants $\epsilon_1$ and $\epsilon_2$ such that $0<\epsilon_2\le\epsilon_1<1$ and $\epsilon_2 n\le N_{t-1}=i\le\epsilon_1 n$, is super-polynomially close to $0$. Adding up such probabilities with respect to $\lambda$ ($\lambda$ is polynomial in $n$) different offspring individuals yields an upper bound for the probability of $N_t^{(O)}\in\mathbb{F}_1$ conditional on the same event, which is still super-polynomially close to $0$.

By Lemmas \ref{lemma:_adaptive_prob_sigmapm_sym}.2 and \ref{lemma:_adaptive_prob_sigmapm_sym}.4, the probability of the event $N_t^{(\chi)}\in\mathbb{L}_1$, conditional on the event that $\exists$ constants $\epsilon_1$ and $\epsilon_2$ such that $0<\epsilon_2\le\epsilon_1<1$ and $\epsilon_2 n\le N_{t-1}=i\le\epsilon_1 n$, is super-polynomially close to $0$. Adding up such probabilities with respect to $\lambda$ ($\lambda$ is polynomial in $n$) different offspring individuals yields an upper bound for the probability of $N_t^{(O)}\in\mathbb{L}_1$ conditional on the same event, which is still super-polynomially close to $0$.

\item[] \textbf{Case iii [\emph{$N_{t-1}=n-o(n)$ and $N_{t-1}<n-n/\log^3 n$ hold}]}: By Lemmas
\ref{lemma:_adaptive_prob_sigma}.3 and \ref{lemma:_adaptive_prob_sigma}.4, the probability of the event $N_t^{(\chi)}\in\mathbb{F}_1$ ($\forall \chi\in\{1,\dots,\lambda\}$), conditional on $N_{t-1}=n-o(n)$ and $N_{t-1}<n-n/\log^3 n$, is super-polynomially close to $0$. Adding up such probabilities with respect to $\lambda$ ($\lambda$ is polynomial in $n$) different offspring individuals yields an upper bound for the probability of $N_t^{(O)}\in\mathbb{F}_1$ conditional on the same event, which is still super-polynomially close to $0$.

By Lemmas \ref{lemma:_adaptive_prob_sigmapm_sym}.1 and \ref{lemma:_adaptive_prob_sigmapm_sym}.4, the probability of the event $N_t^{(\chi)}\in\mathbb{L}_1$, conditional on $N_{t-1}=n-o(n)$ and $N_{t-1}<n-n/\log^3n$, is super-polynomially close to $0$. Adding up such probabilities with respect to $\lambda$ ($\lambda$ is polynomial in $n$) different offspring individuals yields the upper bound for the probability of $N_t^{(O)}\in\mathbb{L}_1$ conditional on the same event, which is still super-polynomially close to $0$.
\end{enumerate}

By summarizing the above results, we have proven
$$\forall t\in\mathbb{N}^+: \mathbb{P}(N_{t}^{(P)}\in\mathbb{O}_1, N_{t}^{(O)}\in\mathbb{F}_1\cup \mathbb{L}_1\mid N_{t-1}\in \mathbb{O}_1)\prec 1/SuperPoly(n). $$
$\hfill\square$

The proof of Theorem \ref{theorem:large_sigma_1plus_lambda_bitmatching} is almost the same to that of Theorem \ref{theorem:large_sigma_1plus_lambda}. The only difference between them is that the former utilizes the original version of Lemma \ref{lemma:_adaptive_prob_sigmapm_sym} for the general BDOP class, while the latter utilizes the specific version of Lemma \ref{lemma:_adaptive_prob_sigmapm_sym} for the \textsc{BitMatching}$_D$ problem. For the sake of brevity, we do not provide the details here.

\section{Proof of Theorems \ref{theorem:generalization_low_sigma}
and \ref{theorem:px_low_sigma}} The main difference between the proofs of Theorems \ref{theorem:generalization_low_sigma} and \ref{theorem:px_low_sigma} is that the former utilizes the result of Lemma \ref{lemma:_generalization_large_sigma} so as to restrict the analysis of the general BDOP class to the case $\sigma\le \delta\log n/n$, while the latter utilizes Lemma \ref{lemma:large_sigma} to restrict the analysis of \textsc{BitMatching}$_D$ problem to the case $\sigma\le \delta\log n/n$, where $\delta$ is an arbitrary positive constant. Here we only provide a unified proof for the sake of brevity. As mentioned in Section \ref{sec:small_sigma}, the proof contains the analysis related to Propositions B\ref{theorem:generalization_low_sigma}.1, B\ref{theorem:generalization_low_sigma}.2, B\ref{theorem:generalization_low_sigma}.3, B\ref{theorem:generalization_low_sigma}.4, and B\ref{theorem:generalization_low_sigma}.5. Here we study the above propositions one after another.

Theorems \ref{theorem:generalization_low_sigma} and \ref{theorem:px_low_sigma} are about the $(1+1)$ EA which generates a unique offspring individual at each generation. For the sake of simplicity, in the proof, $P_m(n,t,1)$, which represents the concrete mutation rate employed by the mutation of the parent individual at the $t^{th}$ generation, is written as $P_m(n,t)$ for short, where we omit the offspring index. Similarly, the offspring index will also be omitted when applying the transition lemmas.

 \textbf{Analysis of Proposition B\ref{theorem:generalization_low_sigma}.1}. Since
$(\mathbb{A}_2\cup\mathbb{A}_1\cup\mathbb{F}_2\cup\mathbb{B}_2\cup\mathbb{B}_1\cup\mathbb{L}_2)\subset (\mathbb{F}_1\cup\mathbb{L}_1)$, the proof of this sketch is the same to ``\emph{Analysis of Proposition A1.1}'' part in the proof of Lemmas \ref{lemma:_generalization_large_sigma} and \ref{lemma:large_sigma}.

\textbf{Analysis of Proposition B\ref{theorem:generalization_low_sigma}.2}. Let $U_1=U_1(n,\sigma)=\delta\gamma^{1/7}\log n$. According to Chernoff bounds, we know that with an overwhelming probability there are at most $U_1$ flipped bits among the total $n$ bits after the DOP change (which implies that the number of matching bits can decrease or increase by at most $U_1$ after DOP change):
\begin{eqnarray}
\nonumber&&\mathbb{P}\bigg(|N_t^{(P)}-N_{t-1}|\ge U_1\mid P_m(n,t)<
\frac{\gamma^{1/14}\log n}{n},\sigma<\frac{\delta\log n}{n}\bigg)\\
\label{equation:step_size_small_mutation_parent} &<& \bigg(\frac{e}{\gamma^{1/7}}\bigg)^{U_1}=\bigg(\frac{e}{\omega(1)}\bigg)^{\omega(\log n)}\prec \frac{1}{SuperPoly(n)},
\end{eqnarray}
where $t\in\mathbb{N}^+$ is the generation index. On the other hand, concerning the number of matching bits of the offspring at the $t^{th}$ generation, we obtain another inequality by Chernoff bounds:
\begin{eqnarray*}
&&\mathbb{P}\bigg(|N_t^{(O)}-N_{t-1}|\ge U_1\mid P_m(n,t)
<\frac{\gamma^{1/14}\log n}{n},\sigma<\frac{\delta\log n}{n}\bigg)\\
&<&\bigg(\frac{e}{\Theta(\gamma^{1/14})}\bigg)^{U_1}=\bigg(\frac{e}{\omega(1)}\bigg)^{\omega(\log n)}\prec\frac{1}{SuperPoly(n)},
\end{eqnarray*}
where we utilize the fact that the composite bitwise mapping rate (including both DOP change and mutation) within the $t^{th}$ generation, denoted by $r(n,t)$, satisfies that $r(n,t)=(1-\sigma)P_m(n,t)+(1-P_m(n,t))\sigma=P_m(n,t)+\sigma-2P_m(n,t)\sigma<2\gamma^{1/14}\log n/n$ (since $P_m(n,t)<\gamma^{1/14}\log n/n$ and $\sigma<\delta\log n/n<\gamma^{1/14}\log n/n$ holds).

Noting that $N_t\in\{N_t^{(P)},N_t^{(O)}\}$, we obtain the following result by combining the above two inequalities together:
\begin{eqnarray*}
&&\mathbb{P}\bigg(|N_t-N_{t-1}|\ge U_1\mid P_m(n,t)
<\frac{\gamma^{1/14}\log n}{n},\sigma<\frac{\delta\log n}{n}\bigg)\\
&<&\mathbb{P}\bigg(|N_t^{(P)}-N_{t-1}|\ge U_1\mid P_m(n,t)<
\frac{\gamma^{1/14}\log n}{n},\sigma<\frac{\delta\log n}{n}\bigg)\\
&&\quad\quad+\mathbb{P}\bigg(|N_t^{(O)}-N_{t-1}|\ge U_1\mid P_m(n,t)
<\frac{\gamma^{1/14}\log n}{n},\sigma<\frac{\delta\log n}{n}\bigg)\\
&\prec&\frac{1}{SuperPoly(n)}.
\end{eqnarray*}
Consequently, we have
\begin{eqnarray*}
\mathbb{P}\bigg(|N_t-N_{t-1}|<U_1\mid P_m(n,t) <\frac{\gamma^{1/14}\log n}{n},\sigma<\frac{\delta\log n}{n}\bigg)\succ 1-\frac{1}{SuperPoly(n)},
\end{eqnarray*}
thus we have proven Proposition B\ref{theorem:generalization_low_sigma}.2.

 \textbf{Analysis of Proposition B\ref{theorem:generalization_low_sigma}.3}.
  Let us consider Proposition B\ref{theorem:generalization_low_sigma}.3a first. Proposition B\ref{theorem:generalization_low_sigma}.1 tells us that $N_0\notin
\mathbb{A}_2\cup\mathbb{A}_1\cup\mathbb{F}_2\cup\mathbb{B}_2\cup\mathbb{B}_1\cup\mathbb{L}_2$ holds with an overwhelming probability. To arrive at $\mathbb{A}_2\cup\mathbb{A}_1\cup\mathbb{F}_2$ at some generation (e.g., the $t^{th}$ generation), the EA has two choices when deciding the mutation rate of the $t^{th}$ generation:
\begin{enumerate}
\item Small Mutation Rate (SMR): $P_m(n,t)< \gamma^{1/14}\log n/n$;
\item Large Mutation Rate (LMR): $P_m(n,t)\ge\gamma^{1/14}\log n/n$.
\end{enumerate}

\begin{figure} [htbg]
\centering
\includegraphics[width=0.6\textwidth]{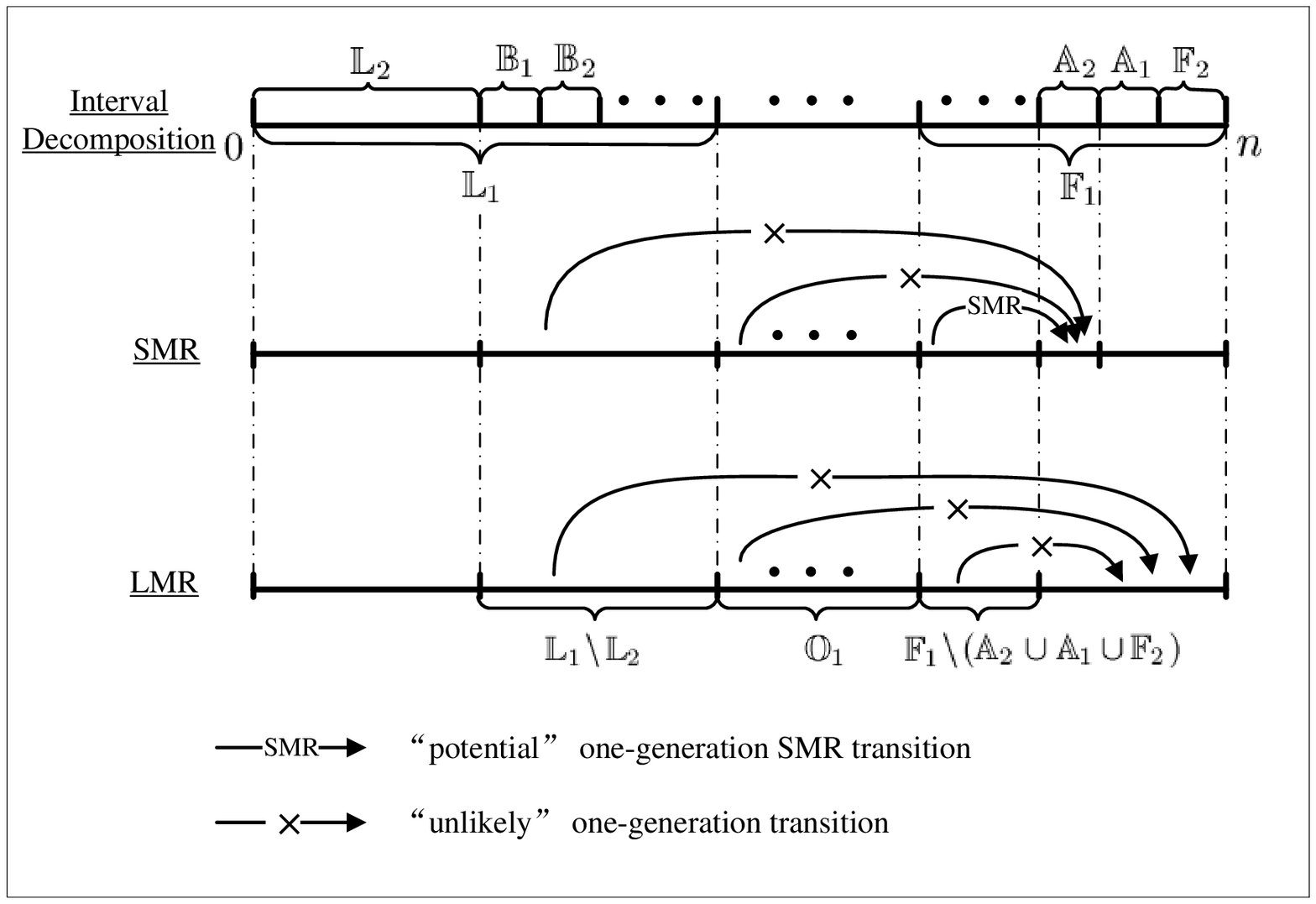}
\caption{Illustration of proof for Proposition B\ref{theorem:generalization_low_sigma}.3.} \label{fig:the2transF2}
\end{figure}

Let us investigate the case of reaching $\mathbb{A}_2\cup\mathbb{A}_1\cup\mathbb{F}_2$ by adopting SMR from $\mathbb{F}_1\setminus(\mathbb{A}_2\cup\mathbb{A}_1\cup\mathbb{F}_2)$, $\mathbb{O}_1$ and $\mathbb{L}_1\setminus\mathbb{L}_2$. Since in Proposition B\ref{theorem:generalization_low_sigma}.2 we have proven that SMR can only provide relatively smaller increment $U_1$ for the number of matching bits with an overwhelming probability, we know that the only region that is possible\footnote{Here ``possible'' means that the event is with at least a polynomially large probability (There exists a positive polynomial function of the problem size $n$, such that the probability is no smaller than the reciprocal of the polynomial function).} to reach $\mathbb{A}_2\cup\mathbb{A}_1\cup\mathbb{F}_2$ by adopting SMR, is a subset of $\mathbb{F}_1\setminus(\mathbb{A}_2\cup\mathbb{A}_1\cup\mathbb{F}_2)$.

According to Proposition B\ref{theorem:generalization_low_sigma}.2, once SMR is used, the number of matching bits can increase by at most $U_1=\delta\gamma^{1/7}\log n$ with an overwhelming probability. Given the condition that $N_{t-1}\in \mathbb{F}_1\setminus(\mathbb{A}_2\cup\mathbb{A}_1\cup\mathbb{F}_2)$, the fact $G=\omega(U_1)$ (with respect to the problem size $n$) implies that even the maximal one-generation increment $U_1$ is not valid to jump over the interval $\mathbb{A}_2$ with an overwhelming probability. Thus, in this case, the EA must reach some intermediate point belonging to $\mathbb{A}_2$ first (otherwise the first hitting time to $\mathbb{F}_2$ has already been proven to be super-polynomial, which is the final conclusion of the theorem). In other words, we have proven that one way to reach $\mathbb{F}_2$ is to reach $\mathbb{A}_2$ first.

Now we consider the case in which the EA reaches $\mathbb{A}_2\cup\mathbb{A}_1\cup\mathbb{F}_2$ by adopting LMR. Here two subcases must be considered further. In the first subcases, the offspring resulted from the LMR is not preserved by the selection operator of the EA, instead, the parent is preserved by the selection operator. This subcase can be viewed as adopting SMR with the value of $0$ (which will not mutate the parent at all), and thus can be included in the analysis of the last two paragraphs.

In the second subcases, the offspring resulted from the LMR is preserved by the selection operator of the EA. As we have shown in the third sketch in Fig. \ref{fig:the2transF2}, we want to prove that $\mathbb{L}_2$ is the only region that is possible to reach $\mathbb{A}_2\cup\mathbb{A}_1\cup\mathbb{F}_2$ by one-generation transition with LMR (and thus $\mathbb{L}_2$ is the only region that can reach $\mathbb{F}_2$ in one generation adopting LMR). As it is shown in the third sketch in Fig. \ref{fig:the2transF2}, to prove the above proposition, there are three intervals for us to exclude (prove to be ``unlikely''): $\mathbb{L}_1\setminus\mathbb{L}_2$,$\mathbb{O}_1$ and $\mathbb{F}_1\setminus(\mathbb{A}_2\cup\mathbb{A}_1\cup\mathbb{F}_2)$. According to Lemma \ref{lemma:_adaptive_prob_sigmapm_any_eta}, if $N_{t-1}\in\mathbb{O}_1\cup(\mathbb{F}_1\setminus(\mathbb{A}_2\cup\mathbb{A}_1\cup\mathbb{F}_2))=(n/\log^2 n,n-3G)$ and $P_m(n,t)\ge\gamma^{1/14}\log n/n$, the probability of $N_{t}^{(O)}\in\mathbb{A}_2\cup\mathbb{A}_1\cup\mathbb{F}_2$ is super-polynomially close to $0$.

In other words, to prove Proposition B\ref{theorem:generalization_low_sigma}.3a, we only need to concern the case in which $N_{t-1}\in \mathbb{L}_1\setminus\mathbb{L}_2=(4G,n/\log^2n]$ and $P_m(n,t)\ge\gamma^{1/14}\log n/n$ hold. Given the above two conditions, below we prove that the probability of $N_{t}^{(O)}\in\mathbb{A}_2\cup\mathbb{A}_1\cup\mathbb{F}_2=[n-3G,n]$ is super-polynomially close to $0$. Given an arbitrary constant $h\in(0,1)$,

$-$ if $\gamma^{1/14}\log n/n\le P_m(n,t)\le 1-h$ holds. Given the conditions that $\sigma<\delta\log n/n$ and $\gamma^{1/14}\log n/n\le P_m(n,t)\le 1-h$, by applying Lemma \ref{lemma:two_bitwise} we obtain:
\begin{eqnarray*}
&&r(n,t)=\omega(\log n/n),\\
&&r(n,t)\le \max\Big\{\frac{1}{2},1-h\Big\}.
\end{eqnarray*}
By above inequalities, we estimate the probability that in one generation the EA finds number of matching bits $N_{t}^{(O)}= j \in\mathbb{A}_2\cup\mathbb{A}_1\cup\mathbb{F}_2$: {\small
\begin{eqnarray}
\nonumber&&\mathbb{P}\Bigg(N_{t}^{(O)}=j\mid N_{t-1}=i\in\mathbb{L}_1\setminus\mathbb{L}_2,j\in\mathbb{A}_2\cup\mathbb{A}_1\cup\mathbb{F}_2,
\frac{\gamma^{1/14}\log n}{n}\le P_m(n,t)\le 1-h\Bigg)\\
\nonumber&=&\sum_{k=0}^{\min\{i,n-j\}}{n-i\choose j-i+k}{i\choose
k}r(n,t)^{j-i+2k}(1-r(n,t))^{n-(j-i+2k)}\\
\nonumber&<&r(n,t)^{j-i}\sum_{k=0}^{\min\{i,n-j\}}{n-i\choose
n-j-k}{i\choose k}={n\choose n-j}r(n,t)^{j-i}< n^{n-j}\max\Big\{\frac{1}{2},1-h\Big\}^{j-i}\quad\quad\text{(by Lemma \ref{lemma:bi_coeffi} in Appendix)}\\
\nonumber&=& n^{o\big(\frac{n}{\log n}\big)}\max\Big\{\frac{1}{2},1-h\Big\}^{n-o(n)}=2^{o(n)} \max\Big\{\frac{1}{2},1-h\Big\}^{n-o(n)}\prec\frac{1}{SuperPoly(n)},
\end{eqnarray}
} which is a super-polynomially small probability.

\begin{figure} [htbg]
\centering
\includegraphics[width=0.6\textwidth]{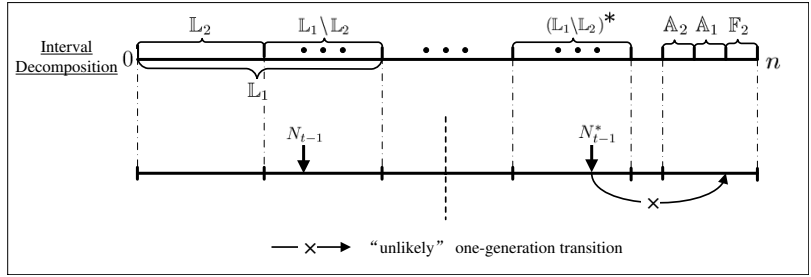}
\caption{Sketch of the case $P_m(n,t)>1-h$, where $(\mathbb{L}_1\setminus\mathbb{L}_2)^*=(n-n/\log^2n,n-4G)$.} \label{fig:the2sym}
\end{figure}

$-$ if $P_m(n,t)>1-h$ holds (i.e., $\mathbb{P}(P_m(n,t)>1-h)=1$). On one hand, given the condition that $\sigma=\omega(\log n/n)$ and $P_m(n,t)> 1-h$, by applying Lemma \ref{lemma:two_bitwise} we obtain:
\begin{eqnarray*}
r(n,t)>\min\bigg\{\frac{1}{2},\max\Big\{\sigma,P_m\Big\}\bigg\}>\min\bigg\{\frac{1}{2}, \max\Big\{\sigma,1-h\Big\}\bigg\}\ge\min\Big\{\frac{1}{2},1-h\Big\}.
\end{eqnarray*}
On the other hand, we must note the fact called symmetrical bitwise mapping (Fig. \ref{fig:the2sym}): given the condition that the number of matching bits $N_{t-1}$ equals $i$ and the composite bitwise mapping rate $r(n,t)$, the consequence of the bitwise mapping is equivalent to that of the case in which $N_{t-1}$ equals $n-i$ and the composite bitwise mapping rate equals $1-r(n,t)$. Formally, we have
\begin{eqnarray*}
1-r(n,t)< 1-\min\Big\{\frac{1}{2},1-h\Big\}=\max\Big\{\frac{1}{2},h\Big\}.
\end{eqnarray*}
Noting the fact described above, the following equation holds in response to the so-called symmetrical bitwise mapping:
\begin{eqnarray*}
&&\mathbb{P}\Bigg(N_{t}^{(O)}=j\mid
N_{t-1}=i,i\in\mathbb{L}_1\setminus\mathbb{L}_2,j\in\mathbb{A}_2\cup\mathbb{A}_1\cup\mathbb{F}_2, r(n,t)\Bigg)\\
&=&\mathbb{P}\Bigg(N_{t}^{(O)}=j\mid N_{t-1}^*=n-i,i\in\mathbb{L}_1\setminus\mathbb{L}_2,j\in\mathbb{A}_2\cup\mathbb{A}_1\cup\mathbb{F}_2, r^*(n,t)<1-r(n,t)\Bigg),
\end{eqnarray*}
where we use $r^*(n,t)$ to represent the \emph{notional} composite bitwise mapping rate with the value of $1-r(n,t)$, $N_{t-1}^*$ represent the \emph{notional} number of matching bits (found by the EA) at the end of the $(t-1)^{th}$ generation.

As shown in Fig. \ref{fig:the2sym}, we only need to prove that the probability of reaching $\mathbb{A}_2\cup\mathbb{A}_1\cup\mathbb{F}_2$ is super-polynomially close to $0$, given the conditions that the notional number of matching bits $N_{t-1}^*$ belongs to $(\mathbb{L}_1\setminus\mathbb{L}_2)^*=(n-n/\log^2n,n-4G)$ and the notional composite bitwise mapping rate $r^*(n,t)< \max\{1/2,h\}$. According to the value of $r^*(n,t)$, there are two situations.

In the first situation, $r^*(n,t)=O(\log n/n)$ holds. According to Chernoff bounds, we know that with an overwhelming probability there are at most $\gamma^{1/14}\log n$ flipped bits among the total $n$ bits:
\begin{eqnarray*}
\mathbb{P}\bigg(|N_{t}^{(O)}-N_{t-1}^*|< \gamma^{1/14}\log n\mid r^*(n,t)=O\Big(\frac{\log n}{n}\Big)\bigg)>1-\bigg(\frac{e}{\Omega(\gamma^{1/14})}\bigg)^{\gamma^{1/14}\log n}\succ 1-\frac{1}{SuperPoly(n)}.
\end{eqnarray*}
Consequently, with an overwhelming probability,  the number of matching bits will decrease or increase by at most $\gamma^{1/14}\log n$ after the overall bitwise mapping (including DOP change and mutation). Given the conditions that $N_{t-1}^*=n-i$ and $i\in\mathbb{L}_1\setminus\mathbb{L}_2$, the above upper bound implies that $N_{t}^{(O)}<n-4G+\gamma^{1/14}\log n< n-3G$ and $N_{t}^{(O)}\notin \mathbb{L}_2$ hold with an overwhelming probability. In other words,
\begin{eqnarray*}
\mathbb{P}\Bigg(N_{t}^{(O)}=j\mid N_{t-1}^*=n-i,i\in\mathbb{L}_1\setminus\mathbb{L}_2,j\in\mathbb{A}_2\cup\mathbb{A}_1\cup\mathbb{F}_2, r^*(n,t)=O\Big(\frac{\log n}{n}\Big)\Bigg)\prec \frac{1}{SuperPoly(n)}.
\end{eqnarray*}

In the second situation, $r^*(n,t)=\omega(\log n/n)$ holds. This case can be viewed as imposing a mutation with LMR $r^*(n,t)=\omega(\log n/n)$ to an individual with its number of matching bits belonging to $[n-n/\log^2n,n-4G)$. According to Lemma \ref{lemma:_adaptive_prob_sigmapm_any_eta}.3, we have:
\begin{eqnarray*}
\mathbb{P}\Bigg(N_{t}^{(O)}=j\mid N_{t-1}^*=n-i,i\in\mathbb{L}_1\setminus\mathbb{L}_2,j\in\mathbb{A}_2\cup\mathbb{A}_1\cup\mathbb{F}_2, r^*(n,t)=\omega\Big(\frac{\log n}{n}\Big)\Bigg)\prec \frac{1}{SuperPoly(n)}.
\end{eqnarray*}

Since there is no essential difference between the proofs related to $r^*(n,t)$ and $r(n,t)$, we will not provide the proof here for the sake of brevity. For details, one can refer to the proof of Lemma \ref{lemma:_adaptive_prob_sigmapm}.3 in the appendix.

Combining the above two situations of $r^*(n,t)$ together, we obtain that:
\begin{eqnarray}
\nonumber&&\mathbb{P}\Bigg(N_{t}^{(O)}=j\mid N_{t-1}=i,i\in\mathbb{L}_1\setminus\mathbb{L}_2,j\in\mathbb{A}_2\cup\mathbb{A}_1\cup\mathbb{F}_2, P_m(n,t)>1-h,\sigma<\frac{\delta\log
n}{n}\Bigg)\\
\label{equation:transfer_condition_1}&=&\mathbb{P}\Bigg(N_{t}^{(O)}=j\mid N_{t-1}=i,i\in\mathbb{L}_1\setminus\mathbb{L}_2,j\in\mathbb{A}_2\cup\mathbb{A}_1\cup\mathbb{F}_2,
r(n,t)>\max\Big\{\frac{1}{2},1-h\Big\}\Bigg)\\
\nonumber&=&\mathbb{P}\Bigg(N_{t}^{(O)}=j\mid N_{t-1}^*=n-i,i\in\mathbb{L}_1\setminus\mathbb{L}_2,j\in\mathbb{A}_2\cup\mathbb{A}_1\cup\mathbb{F}_2,
r^*(n,t)<\max\Big\{\frac{1}{2},h\Big\}\Bigg)\\
\nonumber&\prec& \frac{1}{SuperPoly(n)},
\end{eqnarray}
where we utilize $\mathbb{P}(P_m(n,t)>1-h,\sigma<\delta\log n/n)=1$ ($P_m(n,t)>1-h$ and $\sigma<\delta\log n/n$ are preconditions of the analysis here) to obtain Eq. \ref{equation:transfer_condition_1}.

 By applying total probability theorem
\cite{Papoulis84}, we further combine the cases $\gamma^{1/14}\log n/n\le P_m(n,t)\le 1-h$ and $P_m(n,t)>1-h$ together. As a result, we obtain
\begin{eqnarray*}
\mathbb{P}\Bigg(N_{t}^{(O)}=j\mid N_{t-1}\in\mathbb{L}_1\setminus\mathbb{L}_2,j\in\mathbb{A}_2\cup\mathbb{A}_1\cup\mathbb{F}_2, P_m(n,t)\ge\frac{\gamma^{1/14}\log n}{n},\sigma<\frac{\delta\log n}{n}\Bigg)\prec \frac{1}{SuperPoly(n)}.
\end{eqnarray*}
Till now, the cases of SMR and LMR have been analyzed rigorously, and we know that the probability of reaching $\mathbb{F}_2$ from $(4G,n-3G)$ by any one-generation transition is super-polynomially small. Recall that we have also proven that the EA must reach $\mathbb{A}_2$ before reaching $\mathbb{A}_1\cup\mathbb{F}_2$ if SMR is used in the one-generation transition that reaches $\mathbb{A}_2\cup\mathbb{A}_1\cup\mathbb{F}_2$, we have proven Proposition B\ref{theorem:generalization_low_sigma}.3a.

The proof of Proposition B\ref{theorem:generalization_low_sigma}.3b is similar to that of Proposition B\ref{theorem:generalization_low_sigma}.3a. The major difference between the ideas of the two proofs is that, in Proposition B\ref{theorem:generalization_low_sigma}.3a the probability of reaching $\mathbb{A}_2\cup\mathbb{A}_1\cup\mathbb{F}_2$ from $\mathbb{L}_2$ (given the condition\footnote{Since Lemma \ref{lemma:large_sigma} has already discuss the case of $\sigma=\omega(\log n/n)$, we only need to consider the case of $\sigma<\delta\log n/n$ here.} that $\sigma<\delta\log n/n$) by one-generation transition is \emph{not} super-polynomially close to $0$, while in Proposition B\ref{theorem:generalization_low_sigma}.3b \emph{the probability of reaching $\mathbb{L}_2$ from $\mathbb{A}_2\cup\mathbb{A}_1$ (given the condition that $\sigma<\delta\log n/n$) by one-generation transition (including the shift of optimum of BDOP and the mutation for the solution) is super-polynomially close to $0$}. The brief proofs of the latter proposition, conditional on the general BDOP class (for proving Theorem \ref{theorem:generalization_low_sigma}) and the \textsc{BitMatching}$_D$ problem (for proving Theorem \ref{theorem:px_low_sigma}), are presented respectively.

(\emph{For general BDOP class}) Noting that the shifting rate $\sigma\le\delta\log n/n$, we estimate the following probability by Chernoff bound:
\begin{eqnarray}
\nonumber&&\mathbb{P}\bigg(|N_t^{(P)}-N_{t-1}|\ge U_1\mid \sigma<\frac{\delta\log n}{n}\bigg)<\bigg(\frac{e}{\gamma^{1/7}}\bigg)^{U_1}=\bigg(\frac{e}{\omega(1)}\bigg)^{\omega(\log n)}\prec \frac{1}{SuperPoly(n)}.
\end{eqnarray}
Noting that the range between $\mathbb{L}_2$ and $\mathbb{A}_2\cup\mathbb{A}_1$ is much larger than $U_1$ (i.e., $n-7G>U_1$), we know that, given the condition that $\sigma<\delta\log n/n$, the probability of reaching $\mathbb{L}_2$ from $\mathbb{A}_2\cup\mathbb{A}_1$ \emph{by the parent of the $t^{th}$ generation} is super-polynomially close to $0$. Formally,
\begin{eqnarray}\label{equation:parent_fail_to_longjump}
&&\mathbb{P}\bigg(N_t^{(P)}\in\mathbb{L}_2\mid N_{t-1}\in\mathbb{A}_2\cup\mathbb{A}_1, \sigma<\frac{\delta\log n}{n} \bigg)\prec\frac{1}{SuperPoly(n)}.
\end{eqnarray}

Meanwhile, we estimate the probability that $N_t^{(O)}\in\mathbb{L}_2$, conditional on $N_{t-1}\in\mathbb{A}_2\cup\mathbb{A}_1$ and $r(n,t)\le 1-1/\log n$, where $r(n,t)\le 1-1/\log n$ is derived from $\sigma\le\delta\log n/n$ and $P_m(n,t)\le 1-1/\log n$ (a condition of Theorem \ref{theorem:generalization_low_sigma}) by Lemma \ref{lemma:two_bitwise}. Let $L_{t}^-$ be the number of flipped matching bits after the DOP change and the mutation at the $t^{th}$ generation, we have
\begin{eqnarray*}
&&\mathbb{P}\bigg(N_t^{(O)}\in\mathbb{L}_2\mid N_{t-1}\in\mathbb{A}_2\cup\mathbb{A}_1, r(n,t)\le 1-\frac{1}{\log n}
\bigg)\\
&=&\mathbb{P}\bigg( N_t^{(O)}\le 4G\mid N_{t-1}\in[n-3G,n-G), r(n,t)\le 1-\frac{1}{\log n}
\bigg)\\
&\le&\mathbb{P}\bigg(n-\Big(L_{t}^-+(n-N_{t-1})\Big)\le4G\mid N_{t-1}\in[n-3G,n-G), r(n,t)\le 1-\frac{1}{\log n}
\bigg)\\
&=&\mathbb{P}\bigg(L_{t}^-\ge N_{t-1}-4G\mid N_{t-1}\in[n-3G,n-G), r(n,t)\le 1-\frac{1}{\log n}
\bigg)\\
&\le&\mathbb{P}\bigg(L_{t}^-\ge n-7G\mid N_{t-1}\in[n-3G,n-G), r(n,t)\le 1-\frac{1}{\log n} \bigg)
\end{eqnarray*}

On one hand, if $r(n,t)>\frac{1}{4e}$, then we further have
\begin{eqnarray}
\nonumber&&\mathbb{P}\bigg(N_t^{(O)}\in\mathbb{L}_2\mid N_{t-1}\in\mathbb{A}_2\cup\mathbb{A}_1, \frac{1}{4e}<r(n,t)\le 1-\frac{1}{\log n}
\bigg)\\
\nonumber&\le&\mathbb{P}\bigg(L_{t}^-\ge (1+\rho_1)n\Big(1-\frac{1}{\log n}\Big)>(1+\rho_1)\hat{\mathbb{E}}[L_{t}^-]\mid N_{t-1}\in[n-3G,n-G), \frac{1}{4e}<r(n,t)\le 1-\frac{1}{\log n}
\bigg)\\
\nonumber&<&\mathbb{P}\bigg(L_{t}^->(1+\rho_1)\hat{\mathbb{E}}[L_{t}^-]\mid N_{t-1}\in[n-3G,n-G), \frac{1}{4e}<r(n,t)\le 1-\frac{1}{\log n}
\bigg)\\
\label{equation:r_large}&<&e^{-\hat{\mathbb{E}}[L_{t}^-]\rho_1^2/4}\prec\frac{1}{SuperPoly(n)},
\end{eqnarray}
where $\rho_1=(n-7G)/(n-n/\log n)-1=\Theta(1/\log n)$ (since $G=\gamma^{4/7}\log n$ and $\gamma\le n/\log n$), $\hat{\mathbb{E}}[L_{t}^-]=\mathbb{E}[L_{t}^-\mid N_{t-1}\in[n-3G,n-G), 1/4e<r(n,t)\le 1-1/\log n]=\Theta(n)$, and the last two inequalities is obtained by Chernoff bound.

On the other hand, let us consider the case $r(n,t)\le\frac{1}{4e}$. By Chernoff bound, we have
\begin{eqnarray}
\nonumber&&\mathbb{P}\bigg(N_t^{(O)}\in\mathbb{L}_2\mid N_{t-1}\in\mathbb{A}_2\cup\mathbb{A}_1, r(n,t)\le \frac{1}{4e}
\bigg)\\
\nonumber&=&\mathbb{P}\bigg(L_{t}^-\ge n-7G=(1+\rho_2)\bar{\mathbb{E}}[L_{t}^-]\mid N_{t-1}\in[n-3G,n-G), r(n,t)\le \frac{1}{4e}
\bigg)\\
\label{equation:r_small}&<&\Big(\frac{e}{1+\rho_2}\Big)^{(1+\rho_2)\bar{\mathbb{E}}[L_{t}^-]} =\Big(\frac{e}{1+\rho_2}\Big)^{n-7G}\prec\frac{1}{SuperPoly(n)},
\end{eqnarray}
where $\bar{\mathbb{E}}[L_{t}^-]=\mathbb{E}[L_{t}^-\mid N_{t-1}\in[n-3G,n-G), r(n,t)\le 1/4e]\le n/4e$ and $\rho_2=(n-7G)/\bar{\mathbb{E}}[L_{t}^-]-1>(n/2)/(n/4e)-1\ge 2e-1$ (since $G=\gamma^{4/7}\log n$ and $\gamma\le n/\log n$). Combining Eqs. \ref{equation:r_large} and \ref{equation:r_small} together, we obtain
\begin{eqnarray*}
&&\mathbb{P}\bigg(N_t^{(O)}\in\mathbb{L}_2\mid N_{t-1}\in\mathbb{A}_2\cup\mathbb{A}_1, r(n,t)\le 1-\frac{1}{\log n} \bigg)\prec\frac{1}{SuperPoly(n)}.
\end{eqnarray*}
Combining the above inequality with Eq. \ref{equation:parent_fail_to_longjump}, we know that the probability of reaching $\mathbb{L}_2$ from $\mathbb{A}_2\cup\mathbb{A}_1$ by one-generation transition is super-polynomially close to $0$. The rest part of the proof of Proposition B\ref{theorem:generalization_low_sigma}.3b is similar to that of Proposition B\ref{theorem:generalization_low_sigma}.3a.

(\emph{For \textsc{BitMatching}$_D$}) According to Chernoff bounds, with an overwhelming probability there are at most $U_1$ flipped bits among the total $n$ bits after the DOP change (which implies that the number of matching bits can decrease or increase by at most $U_1$ after DOP change):
\begin{eqnarray}
\nonumber&&\mathbb{P}\bigg(|N_t^{(P)}-N_{t-1}|\ge U_1\mid \sigma<\frac{\delta\log n}{n}\bigg)<\bigg(\frac{e}{\gamma^{1/7}}\bigg)^{U_1}=\bigg(\frac{e}{\omega(1)}\bigg)^{\omega(\log n)}\prec \frac{1}{SuperPoly(n)}.
\end{eqnarray}
Noting that the range between $\mathbb{L}_2$ and $\mathbb{A}_2\cup\mathbb{A}_1$ is much larger than $U_1$ ($n-7G>U_1$), we know that, given the condition that $\sigma<\delta\log n/n$, the probability of reaching $\mathbb{L}_2$ from $\mathbb{A}_2\cup\mathbb{A}_1$ \emph{by the help of the DOP change at the beginning of the $t^{th}$ generation} is super-polynomially close to $0$.

Moreover, since the selection operator of the EA always preserves the better individual between the parent and offspring (For \textsc{BitMatching}$_D$, $N_t=\max\{N_t^{(P)},N_t^{(O)}\}\ge N_t^{(P)}$ holds according to Eqs. \ref{equation:N_t} and \ref{equation:N_t_range}), the above inequality also implies that $N_t\notin \mathbb{L}_2$ holds with an overwhelming probability. In other words, the probability of reaching $\mathbb{L}_2$ from $\mathbb{A}_2\cup\mathbb{A}_1$ by one-generation transition is super-polynomially close to $0$. The rest part of the proof of Step \ref{theorem:px_low_sigma}.3b is similar to that of Step \ref{theorem:px_low_sigma}.3a.

 \textbf{Analysis of Proposition B\ref{theorem:generalization_low_sigma}.4}.
Let $U_2=\gamma^{1/7}$.
  According to the condition
$N_{t-1}\in\mathbb{A}_2\cup\mathbb{A}_1\cup\mathbb{F}_2$, there are at most $3G$ non-matching bits at the end of the $(t-1)^{th}$ generation. On the other hand, since the number of flipped non-matching bits is always no smaller than the final increment of the number of matching bits (after DOP change or/and mutation), we know that, to increase the number of matching bits by at least $U_2$, the number of flipped non-matching bits must be larger than $U_2$. Hence, concerning the number of matching bits of the parent after the DOP change at the $t^{th}$ generation, we obtain
\begin{eqnarray}
\nonumber&&\mathbb{P}\bigg(N_t^{(P)}-N_{t-1}\ge U_2\mid
N_{t-1}\in\mathbb{A}_2\cup\mathbb{A}_1\cup\mathbb{F}_2,\sigma<\frac{\delta\log n}{n}\bigg)\\
\nonumber&<&{3G \choose U_2}\sigma^{U_2}(1-\sigma)^{n-U_2}<
{3G \choose \gamma^{1/7}}\sigma^{\gamma^{1/7}}\\
\label{equation:apply_def_G}&<&(3G)^{\gamma^{1/7}}\sigma^{\gamma^{1/7}}<\Big(\frac{3\delta G\log n}{n}\Big)^{\gamma^{1/7}}=\Big(\frac{3\delta
\gamma^{4/7}\log^2 n}{n}\Big)^{\gamma^{1/7}}\\
\label{equation:apply_def_gamma}&<&\Big(\frac{3\delta n^{4/7}\log^2 n}{n}\Big)^{\gamma^{1/7}}\prec \frac{1}{SuperPoly(n)},
\end{eqnarray}
where we obtain Eqs. \ref{equation:apply_def_G} and \ref{equation:apply_def_gamma} by applying Eqs. \ref{equation:def_gamma} and \ref{equation:def_G} respectively.

Next, concerning the number of matching bits of the offspring at the $t^{th}$ generation, we must consider two different cases. The EA adopts SMR at the $t^{th}$ generation in the first case while it adopts LMR at the $t^{th}$ generation in the second case.

For the former case, we estimate the following inequality by noting the condition $N_{t-1}\in\mathbb{A}_2\cup\mathbb{A}_1\cup\mathbb{F}_2$:
\begin{eqnarray}
\nonumber&&\mathbb{P}\bigg(N_t^{(O)}-N_{t-1}\ge U_2\mid N_{t-1}\in\mathbb{A}_2\cup\mathbb{A}_1\cup\mathbb{F}_2,P_m(n,t)<
\frac{\gamma^{1/14}\log n}{n},\sigma<\frac{\delta\log n}{n}\bigg)\\
\nonumber&<&{3G \choose U_2}r(n,t)^{U_2}\Big(1-r(n,t)\Big)^{n-U_2}<
{3G \choose \gamma^{1/7}}r(n,t)^{\gamma^{1/7}}\\
\nonumber&<&(3G)^{\gamma^{1/7}}r(n,t) ^{\gamma^{1/7}}<\Big(\frac{6 G\gamma^{1/14}\log n}{n}\Big)^{\gamma^{1/7}}=\Big(\frac{6
\gamma^{9/14}\log^2 n}{n}\Big)^{\gamma^{1/7}}\\
\label{equation:SMR_within3G}&<&\Big(\frac{6 n^{9/14}\log^2 n}{n}\Big)^{\gamma^{1/7}}\prec \frac{1}{SuperPoly(n)},
\end{eqnarray}
where we utilize the fact that the composite bitwise mapping rate $r(n,t)$ satisfies that $r(n,t)=(1-\sigma)P_m(n,t)+(1-P_m(n,t))\sigma=P_m(n,t)+\sigma-2P_m(n,t)\sigma<2\gamma^{1/14}\log n/n$ (since $P_m(n,t)<\gamma^{1/14}\log n/n$ and $\sigma<\delta\log n/n<\gamma^{1/14}\log n/n$ holds).

On the other hand, let us investigate the latter case in which the EA adopts LMR at the $t^{th}$ generation. According to the condition $N_{t-1}\in\mathbb{A}_2\cup\mathbb{A}_1\cup\mathbb{F}_2$, $N_{t-1}=n-o(n)$ holds. By applying Lemma \ref{lemma:_adaptive_prob_sigmapm_any_eta}.3, we obtain
\begin{eqnarray}
\label{equation:LMR_within3G}\mathbb{P}\bigg(N_t^{(O)}>N_{t-1}\mid N_{t-1}\in\mathbb{A}_2\cup\mathbb{A}_1\cup\mathbb{F}_2,P_m(n,t)\ge \frac{\gamma^{1/14}\log n}{n},\sigma<\frac{\delta\log n}{n}\bigg)\prec \frac{1}{SuperPoly(n)}.
\end{eqnarray}
Combining Eqs. \ref{equation:SMR_within3G} and \ref{equation:LMR_within3G} together we have:
\begin{eqnarray}
\mathbb{P}\bigg(N_t^{(O)}-N_{t-1}\ge U_2\mid N_{t-1}\in\mathbb{A}_2\cup\mathbb{A}_1\cup\mathbb{F}_2,\sigma<\frac{\delta\log n}{n}\bigg)\prec \frac{1}{SuperPoly(n)},
\end{eqnarray}

Noting that $N_t\in\{N_t^{(P)},N_t^{(O)}\}$, we obtain the following result by combining Eqs. \ref{equation:apply_def_gamma} with the above inequality:
\begin{eqnarray*}
&&\mathbb{P}\bigg(N_t-N_{t-1}\ge U_2\mid
N_{t-1}\in\mathbb{A}_2\cup\mathbb{A}_1\cup\mathbb{F}_2,\sigma<\frac{\delta\log n}{n}\bigg)\\
&<&\mathbb{P}\bigg(N_t^{(P)}-N_{t-1}\ge U_2\mid
N_{t-1}\in\mathbb{A}_2\cup\mathbb{A}_1\cup\mathbb{F}_2,\sigma<\frac{\delta\log n}{n}\bigg)\\
&&\quad\quad+\mathbb{P}\bigg(N_t^{(O)}-N_{t-1}\ge U_2\mid
N_{t-1}\in\mathbb{A}_2\cup\mathbb{A}_1\cup\mathbb{F}_2,\sigma<\frac{\delta\log n}{n}\bigg)\\
&\prec&\frac{1}{SuperPoly(n)}.
\end{eqnarray*}
Consequently,
\begin{eqnarray}\label{equation:upper_bound_one_step}
\mathbb{P}\bigg(N_t-N_{t-1}< U_2\mid N_{t-1}\in\mathbb{A}_2\cup\mathbb{A}_1\cup\mathbb{F}_2,\sigma<\frac{\delta\log n}{n}\bigg)\succ 1-\frac{1}{SuperPoly(n)}.
\end{eqnarray}
Hence, we know that once the EA is in $\mathbb{A}_2\cup\mathbb{A}_1\cup\mathbb{F}_2$, the number of matching bits
 can only increase by at most $U_2$ in one generation
with an overwhelming probability no matter which mutation rate the EA adopts.

\textbf{Analysis of Proposition B\ref{theorem:generalization_low_sigma}.5}. As we have shown in Proposition B\ref{theorem:generalization_low_sigma}.3, so far we have not excluded the possibility\footnote{Here ``excluding the possibility'' of an event is referred to proving that the probability of the event is super-polynomially close to $0$.} of the following two events:
\begin{enumerate}
\item The EA reaches $\mathbb{F}_2$ via $\mathbb{B}_2$ and then via $\mathbb{L}_2$ by
multiple-generation transition (it is possible that the EA reaches $\mathbb{F}_2$ from $\mathbb{L}_2$ by large enough LMR, e.g., the mutation rate $1$).

\item The EA reaches $\mathbb{F}_2$ via $\mathbb{A}_2$ (without reaching some intermediate points in $\mathbb{L}_2$) by
multiple-generation transition.
\end{enumerate}
Hence, the proof of Proposition B\ref{theorem:generalization_low_sigma}.5 contains two parts. First, we need to prove that, if the EA has already reached $\mathbb{B}_2$, then it cannot travel through $\mathbb{B}_1$ and reach $\mathbb{L}_2$ with an overwhelming probability. Second, we need to prove that, if the EA has already reached $\mathbb{A}_2$, then it cannot travel through $\mathbb{A}_1$ and reach $\mathbb{F}_2$ with an overwhelming probability. If the above results have been proven, then we can only hope some events with super-polynomially small probability (e.g., the EA reaches $\mathbb{F}_2$ from $\mathbb{A}_2$ by one-generation transition) to happen, which will lead to a super-polynomial first hitting time with an overwhelming probability.

Since the ideas of two parts are quite similar, we only provide the details of the second part (which will lead to the final conclusion of the theorem) for the sake of brevity. Assume that we have proven the first part, that is, if the EA has already reached $\mathbb{B}_2$, then it cannot travel through $\mathbb{B}_1$ and finally reach $\mathbb{L}_2$ with an overwhelming probability. According to Proposition B\ref{theorem:generalization_low_sigma}.4,  to reach $\mathbb{F}_2$, the EA must reach $\mathbb{L}_2$ or $\mathbb{A}_2$ first, we know that the only choice to reach $\mathbb{F}_2$ is via $\mathbb{A}_2$. To reach $\mathbb{F}_2$ via $\mathbb{A}_2$, the EA must travel through $\mathbb{A}_1$. The reason is given by Proposition B\ref{theorem:generalization_low_sigma}.4 and the fact that $\mathbb{A}_1$ is with the length of $G>U_2$.

Next, we will provide the proof of for the aforementioned proposition: if the EA has already reached $\mathbb{A}_2$, then it cannot travel through $\mathbb{A}_1$ and reach $\mathbb{F}_2$ with an overwhelming probability. (as we have mentioned, by the same technique we can prove the similar result for $\mathbb{L}_2$). For $\forall t\in\mathbb{N}^+$, given the conditions that $N_t\in \mathbb{A}_2\cup\mathbb{A}_1$ and $N_{t-1}=i$, we let the probabilities of decreasing and increasing the number of matching bits be $p^-(n,i,t)$ and $p^+(n,i,t)$, respectively, i.e., \begin{eqnarray*} &&p^-(n,i,t)=\mathbb{P}\bigg(N_t<N_{t-1}\mid N_{t-1}=i\in
\mathbb{A}_2\cup\mathbb{A}_1,\sigma<\frac{\delta\log n}{n}\bigg),\\
&&p^+(n,i,t)=\mathbb{P}\bigg(N_t>N_{t-1}\mid N_{t-1}=i\in \mathbb{A}_2\cup\mathbb{A}_1,\sigma<\frac{\delta\log n}{n}\bigg).
\end{eqnarray*}

 Next we prove that $\forall
t\in\mathbb{N}^+$ that satisfies $N_{t-1}=i\in \mathbb{A}_2\cup\mathbb{A}_1$, the following two inequalities holds:
\begin{eqnarray}
\label{equation:n_i_t_decrease}&&p^-(n,i,t)>p_i^-(n)=\Theta\Big(\frac{\gamma\log n}{n}\Big),\\
\label{equation:n_i_t_increase}&&p^+(n,i,t)<p_i^+(n)=\Theta\Big(\frac{\gamma^{4/7}\log n}{n}\Big),
\end{eqnarray}
where $p_i^-=p_i^-(n)$ is a general lower bound of $p^-(n,i,t)$, and $p_i^+=p_i^+(n)$ is a general upper bound of $p^-(n,i,t)$.

To prove the bound for $p^-(n,i,t)$, we need to consider two cases. In the first case, the EA adopts SMR at the $t^{th}$ generation; In the second case, the EA adopts LMR at the $t^{th}$ generation. Concerning the first case, we estimate the following probability:
\begin{eqnarray*}
&&\mathbb{P}\bigg(N_t<N_{t-1}\mid N_{t-1}=i\in \mathbb{A}_2\cup\mathbb{A}_1, P_m(n,t)< \frac{\gamma^{1/14}\log
n}{n},\sigma<\frac{\delta\log n}{n}\bigg)\\
&\ge&\Bigg(\sum_{k=1}^{i}{i \choose k}\sigma^k(1-\sigma)^{i-k}\big(1-P_m(n,t)^k\big)\Bigg)(1-\sigma)^{n-i}\big(1-P_m(n,t)\big)^{n-i}\\
&>&\Bigg(\big(1-P_m(n,t)\big)\sum_{k=1}^{i}{i \choose k}\sigma^k(1-\sigma)^{i-k}\Bigg)(1-\sigma)^{n-i}\big(1-P_m(n,t)\big)^{n-i}\\
&=&\big(1-(1-\sigma)^i\big)(1-\sigma)^{n-i}\big(1-P_m(n,t)\big)^{n-i+1},
\end{eqnarray*}
where we consider the case in which all the non-matching bits are not flipped during the DOP change and mutation
 (this event is with the probability of $(1-\sigma)^{n-i}(1-P_m(n,t))^{n-i}$) while
some of the matching bits are flipped by the DOP change and at least one of these flipped matching bits is not flipped again by mutation (this event is with the probability of $\sum_{k=1}^{i}{i \choose k}\sigma^k(1-\sigma)^{i-k}\big(1-P_m(n,t)^k\big)$. According to the value of $\sigma$, let us consider two subcases:
\begin{enumerate}
\item[1. If] $\sigma\le \frac{1}{n}$, according to Eq. \ref{equation:def_gamma}, we know $\gamma=\sigma n^2/\log n$.
 Since $n-i+1<3G+1<n/(\gamma^{1/7}\log n)$ holds, we have
\begin{eqnarray*}
&&\mathbb{P}\bigg(N_t<N_{t-1}\mid N_{t-1}=i\in \mathbb{A}_2\cup\mathbb{A}_1, P_m(n,t)< \frac{\gamma^{1/14}\log
n}{n},\sigma\le\frac{1}{n}\bigg)\\
&\ge& \big(1-(1-\sigma)^i\big)(1-\sigma)^{n}\big(1-P_m(n,t)\big)^{n-i+1}\ge c_1\cdot i\sigma\Big(1-\frac{\gamma^{1/14}\log n}{n}\Big)^{\frac{n}{\gamma^{1/7}\log n}}\ge \frac{c_1}{2}\cdot \frac{\gamma\log n}{n},
\end{eqnarray*}
where $c_1$ is a positive constant.
\item[2. If] $\frac{1}{n}<\sigma<\delta\log
n/n$, according to Eq. \ref{equation:def_gamma}, we know $\gamma=n/\log n$. Since $3G<n/(\gamma^{1/7}\log n)$ and $i>n-3G>n/2$ hold, we have
\begin{eqnarray*}
&&\mathbb{P}\bigg(N_t<N_{t-1}\mid N_{t-1}=i\in \mathbb{A}_2\cup\mathbb{A}_1, P_m(n,t)< \frac{\gamma^{1/14}\log
n}{n},\frac{1}{n}<\sigma<\frac{\delta\log n}{n}\bigg)\\
&>&\bigg(1-\Big(1-\frac{1}{n}\Big)^i\bigg)(1-\sigma)^{3G}\big(1-P_m(n,t)\big)^{3G}
>c_2\cdot (1-\sigma)^{3G}\big(1-P_m(n,t)\big)^{3G}\\
&>&c_2\cdot \Big(1-\frac{\delta\log n}{n}\Big)^{\frac{n}{\gamma^{1/7}\log n}}\Big(1-\frac{\gamma^{1/14}\log n}{n}\Big)^{\frac{n}{\gamma^{1/7}\log n}}= c_3\cdot \frac{\gamma\log n}{n},
\end{eqnarray*}
 where $c_2$ and $c_3$ are positive constants.
\end{enumerate}
Combining the above two cases together, we know that
\begin{eqnarray}\label{equation:reduce_rate}
\mathbb{P}\bigg(N_t<N_{t-1}\mid N_{t-1}=i\in \mathbb{A}_2\cup\mathbb{A}_1, P_m(n,t)< \frac{\gamma^{1/14}\log n}{n},\sigma<\frac{\delta\log n}{n}\bigg)>\max\Big\{\frac{c_1}{2},c_3\Big\}\cdot\frac{\gamma\log n}{n}
\end{eqnarray}
holds.

We now consider the second case in which the EA adopts LMR at the $t^{th}$ generation. We estimate the following probability:
\begin{eqnarray*}
\mathbb{P}\bigg(N_t^{(P)}<N_{t-1}\mid N_{t-1}=i\in \mathbb{A}_2\cup\mathbb{A}_1, \sigma<\frac{\delta\log n}{n}\bigg)\ge\sum_{k=1}^{i}{i \choose k}\sigma^k(1-\sigma)^{n-k}>\big(1-(1-\sigma)^i\big)(1-\sigma)^{n-i}.
\end{eqnarray*}
Moreover, by applying Lemma \ref{lemma:_adaptive_prob_sigmapm_any_eta}.3, we have
\begin{eqnarray*}
\mathbb{P}\bigg(N_t^{(O)}\ge N_{t-1}\mid N_{t-1}\in\mathbb{A}_2\cup\mathbb{A}_1,P_m(n,t)\ge \frac{\gamma^{1/14}\log n}{n},\sigma<\frac{\delta\log n}{n}\bigg)\prec \frac{1}{SuperPoly(n)}.
\end{eqnarray*}
Combining the above two inequalities together and noting the fact that the selection operator always preserve the better one between the parent and offspring, we have
\begin{eqnarray*}
&&\mathbb{P}\bigg(N_t<N_{t-1}\mid N_{t-1}=i\in\mathbb{A}_2\cup\mathbb{A}_1,P_m(n,t)\ge
\frac{\gamma^{1/14}\log n}{n},\sigma<\frac{\delta\log n}{n}\bigg)\\
&\ge&\mathbb{P}\bigg(N_t^{(P)}< N_{t-1},N_t^{(O)}<N_{t-1}\mid N_{t-1}=i\in\mathbb{A}_2\cup\mathbb{A}_1,P_m(n,t)\ge \frac{\gamma^{1/14}\log n}{n},\sigma<\frac{\delta\log
n}{n}\bigg)\\
&=&\mathbb{P}\bigg(N_t^{(P)}< N_{t-1}\mid N_{t-1}=i\in\mathbb{A}_2\cup\mathbb{A}_1,P_m(n,t)\ge \frac{\gamma^{1/14}\log n}{n},\sigma<\frac{\delta\log
n}{n}\bigg)\\
&&-\mathbb{P}\bigg(N_t^{(P)}< N_{t-1},N_t^{(O)}\ge N_{t-1}\mid N_{t-1}=i\in\mathbb{A}_2\cup\mathbb{A}_1,P_m(n,t)\ge \frac{\gamma^{1/14}\log n}{n},\sigma<\frac{\delta\log
n}{n}\bigg)\\
&>&\mathbb{P}\bigg(N_t^{(P)}<N_{t-1}\mid N_{t-1}=i\in \mathbb{A}_2\cup\mathbb{A}_1, \sigma<\frac{\delta\log
n}{n}\bigg)\\
&&-\mathbb{P}\bigg(N_t^{(O)}\ge N_{t-1}\mid N_{t-1}\in\mathbb{A}_2\cup\mathbb{A}_1,P_m(n,t)\ge \frac{\gamma^{1/14}\log n}{n},\sigma<\frac{\delta\log
n}{n}\bigg) \\
&>&\big(1-(1-\sigma)^i\big)(1-\sigma)^{n-i}-\frac{1}{SuperPoly(n)}> \frac{1}{2}\cdot\big(1-(1-\sigma)^i\big)(1-\sigma)^{n-i}.
\end{eqnarray*}
According to the value of $\sigma$, let us further consider two subcases:
\begin{enumerate}
\item[1. If] $\sigma\le \frac{1}{n}$, according to Eq. \ref{equation:def_gamma}, we know $\gamma=\sigma n^2/\log n$. We have
\begin{eqnarray*}
&&\mathbb{P}\bigg(N_t<N_{t-1}\mid N_{t-1}=i\in \mathbb{A}_2\cup\mathbb{A}_1, P_m(n,t)\ge \frac{\gamma^{1/14}\log
n}{n},\sigma\le\frac{1}{n}\bigg)\\
&>& \frac{1}{2}\cdot\big(1-(1-\sigma)^i\big)(1-\sigma)^{n-i}>\frac{1}{2}\cdot\big(1-(1-\sigma)^i\big)(1-\sigma)^{n}\ge c_4\cdot i\sigma> \frac{c_4}{2}\cdot \frac{\gamma\log n}{n},
\end{eqnarray*}
where $c_4$ is a positive constant.
\item[2. If] $\frac{1}{n}<\sigma<\delta\log
n/n$, according to Eq. \ref{equation:def_gamma}, we know $\gamma=n/\log n$. Since $3G<n/(\delta\log n)$ and $i>n-3G>n/2$ hold, we have
\begin{eqnarray*}
&&\mathbb{P}\bigg(N_t<N_{t-1}\mid N_{t-1}=i\in \mathbb{A}_2\cup\mathbb{A}_1, P_m(n,t)\ge \frac{\gamma^{1/14}\log
n}{n},\frac{1}{n}<\sigma<\frac{\delta\log n}{n}\bigg)\\
&>&\frac{1}{2}\cdot\bigg(1-\Big(1-\frac{1}{n}\Big)^i\bigg)(1-\sigma)^{3G}
>c_5\cdot (1-\sigma)^{3G}>c_5\cdot \Big(1-\frac{\delta\log n}{n}\Big)^{\frac{n}{\delta\log
n}}= c_6\cdot \frac{\gamma\log n}{n},
\end{eqnarray*}
 where $c_5$ and $c_6$ are positive constants.
\end{enumerate}
Combining the above two cases together, we know that
\begin{eqnarray*}
\mathbb{P}\bigg(N_t<N_{t-1}\mid N_{t-1}=i\in \mathbb{A}_2\cup\mathbb{A}_1, P_m(n,t)\ge \frac{\gamma^{1/14}\log n}{n},\sigma<\frac{\delta\log n}{n}\bigg)>\max\Big\{\frac{c_4}{2},c_6\Big\}\cdot\frac{\gamma\log n}{n}
\end{eqnarray*}
holds. Combining Eq. \ref{equation:reduce_rate} with the above inequality, we have proven that
\begin{eqnarray*}
p^-(n,i,t)>p_i^-(n)=\Theta\Big(\frac{\gamma\log n}{n}\Big).
\end{eqnarray*}

Now let us prove the upper bound of $p^+(n,i,t)$. To increase the number of matching bits by DOP change and/or mutation, the number of flipped non-matching bits must be larger than the number of flipped matching bits after DOP change and/or mutation. According to Lemma \ref{lemma:droste}, concerning the relation between $N_t^{(P)}$ and $N_{t-1}$, we have
\begin{eqnarray}\label{equation:par_increase}
&&\mathbb{P}\bigg(N_t^{(P)}>N_{t-1}\mid N_{t-1}=i\in \mathbb{A}_2\cup\mathbb{A}_1,\sigma<\frac{\delta\log n}{n}\bigg)\le \frac{3G}{n-3G}\le\frac{3G}{n/2}\le \frac{6\gamma^{4/7}\log n}{n}.
\end{eqnarray}

For the offspring at the $t^{th}$ generation, we still need to consider two cases. In the first case, the EA adopts SMR at the $t^{th}$ generation; In the second case, the EA adopts LMR at the $t^{th}$ generation.

We now consider the first case in which the EA adopts SMR. By applying Lemma \ref{lemma:droste} we have
\begin{eqnarray}\label{equation:off_increase}
&&\mathbb{P}\bigg(N_t^{(O)}>N_{t-1}\mid N_{t-1}=i\in \mathbb{A}_2\cup\mathbb{A}_1, P_m(n,t)< \frac{\gamma^{1/14}\log n}{n},\sigma<\frac{\delta\log n}{n}\bigg)\le \frac{3G}{n-3G}\le\frac{3G}{n/2}\le \frac{6\gamma^{4/7}\log n}{n},
\end{eqnarray}
where we utilize the fact that the overall impact of DOP change and mutation to the offspring individual at the $t^{th}$ generation can be represented by a overall mapping with bitwise mapping rate $r(n,t)=(1-\sigma)P_m(n,t)+(1-P_m(n,t))\sigma=P_m(n,t)+\sigma-2P_m(n,t)\sigma$.

Noting that $N_t\in\{N_t^{(P)},N_t^{(O)}\}$, we obtain the following result by combining Eqs. \ref{equation:par_increase} and \ref{equation:off_increase} together:
\begin{eqnarray}
\nonumber&&\mathbb{P}\bigg(N_t>N_{t-1}\mid N_{t-1}=i\in \mathbb{A}_2\cup\mathbb{A}_1, P_m(n,t)< \frac{\gamma^{1/14}\log
n}{n},\sigma<\frac{\delta\log n}{n}\bigg)\\
\nonumber&<& \mathbb{P}\bigg(N_t^{(P)}>N_{t-1}\mid N_{t-1}=i\in
\mathbb{A}_2\cup\mathbb{A}_1,\sigma<\frac{\delta\log n}{n}\bigg)\\
\nonumber&&+\mathbb{P}\bigg(N_t^{(O)}>N_{t-1}\mid N_{t-1}=i\in \mathbb{A}_2\cup\mathbb{A}_1, P_m(n,t)<
\frac{\gamma^{1/14}\log n}{n},\sigma<\frac{\delta\log n}{n}\bigg)\\
\label{equation:increase_rate_SMR}&\le&\frac{12\gamma^{4/7}\log n}{n}.
\end{eqnarray}

On the other hand, we consider the case in which LMR is adopted by the EA. By applying Lemma \ref{lemma:_adaptive_prob_sigmapm_any_eta}.3, we know
\begin{eqnarray*}
\mathbb{P}\bigg(N_t^{(O)}>N_{t-1}\mid N_{t-1}\in\mathbb{A}_2\cup\mathbb{A}_1\cup\mathbb{F}_2,P_m(n,t)\ge \frac{\gamma^{1/14}\log n}{n},\sigma<\frac{\delta\log n}{n}\bigg)\prec \frac{1}{SuperPoly(n)}.
\end{eqnarray*}
In other words, if the LMR is used, the number of matching bits found by the offspring at the $t^{th}$ generation will not be larger than $N_{t-1}$. Similar to Eq. \ref{equation:increase_rate_SMR}, we obtain
\begin{eqnarray*}
&&\mathbb{P}\bigg(N_t>N_{t-1}\mid N_{t-1}=i\in \mathbb{A}_2\cup\mathbb{A}_1, P_m(n,t)\ge \frac{\gamma^{1/14}\log
n}{n},\sigma<\frac{\delta\log n}{n}\bigg)\\
 &<& \mathbb{P}\bigg(N_t^{(P)}>N_{t-1}\mid N_{t-1}=i\in
\mathbb{A}_2\cup\mathbb{A}_1,\sigma<\frac{\delta\log n}{n}\bigg)\\
&&+\mathbb{P}\bigg(N_t^{(O)}>N_{t-1}\mid N_{t-1}=i\in \mathbb{A}_2\cup\mathbb{A}_1, P_m(n,t)\ge
\frac{\gamma^{1/14}\log n}{n},\sigma<\frac{\delta\log n}{n}\bigg)\\
&\le&\frac{6\gamma^{4/7}\log n}{n}+\frac{1}{SuperPoly(n)}<\frac{12\gamma^{4/7}\log n}{n}.
\end{eqnarray*}
Combining Eq. \ref{equation:increase_rate_SMR} with the above inequality, we obtain
\begin{eqnarray*}
&&\mathbb{P}\bigg(N_t>N_{t-1}\mid N_{t-1}=i\in \mathbb{A}_2\cup\mathbb{A}_1,\sigma<\frac{\delta\log n}{n}\bigg)<\frac{12\gamma^{4/7}\log n}{n}.
\end{eqnarray*}
In other words, we have proven Eq. \ref{equation:n_i_t_increase}.

Droste utilized the idea of ``effective mutation'' to prove Theorem 2 of \cite{Droste03}. Intuitively speaking, this technique estimates the lower bound for the number of effective mutation (which can be interpreted as the number of generations in which the number of matching bits changes) for reaching the target, and it also estimates the upper bound for effective mutation that can be provided by the EA. If the former one is significantly smaller than the later one, then the EA cannot reach the target. Following this intuitive idea, we then provide the formal proof of the final conclusion of the theorem. By Eqs. \ref{equation:n_i_t_decrease} and \ref{equation:n_i_t_increase}, we obtain the upper bound of the probability of $N_t>N_{t-1}$, conditional on the event that $N_t\neq N_{t-1}$:
\begin{eqnarray}
\nonumber&&\mathbb{P}\bigg(N_t>N_{t-1}\mid N_t\neq N_{t-1}=i\in
\mathbb{A}_2\cup\mathbb{A}_1, \sigma<\frac{\delta\log n}{n}\bigg)\\
\label{equation:efective_prob_increase}&=&\frac{p^+(n,i,t)}{p^+(n,i,t)+p^-(n,i,t)}<\frac{p^+_i}{p^+_i+p^-_i}=\frac{\Theta\big(\gamma^{4/7}\log n/n\big)}{\Theta\big(\gamma^{4/7}\log n/n\big)+\Theta\big(\gamma\log n/n\big)}=\Theta\big(\gamma^{-3/7}\big),
\end{eqnarray}
where $\Theta$ is referred to the asymptotic order of the problem size $n$.

We now prove that the EA will spend super-polynomial number of generations to travel through $\mathbb{A}_1$ and reach $\mathbb{F}_2$. Let $T$ be defined formally as follows:
\begin{eqnarray}\label{equation:equation:eff_mutation}
T=\Big|\Big\{t\in\mathbb{N}^+\mid N_{t-1}\neq N_t,N_{t-1}\in \mathbb{A}_2\cup\mathbb{A}_1,\sigma<\frac{\delta\log n}{n}\Big\}\Big|.
\end{eqnarray}
Recall that in Proposition B\ref{theorem:generalization_low_sigma}.3, we have proven that the EA has to travel through the whole $\mathbb{A}_1$ with length of $G$; In Proposition B\ref{theorem:generalization_low_sigma}.4, we have proven that in one generation the number of matching bits cannot increase by more than $U_2$ with an overwhelming probability. According to Propositions B\ref{theorem:generalization_low_sigma}.3 and B\ref{theorem:generalization_low_sigma}.4, to travel through $\mathbb{A}_1$ and reach $\mathbb{F}_2$, $T$ is lower bounded by $G/U_2=\gamma^{3/7}\log n$ with an overwhelming probability. Formally we have
\begin{eqnarray}\label{equation:T_lowerbound}
\mathbb{P}\Big(T\ge\gamma^{3/7}\log n\Big)\succ 1-\frac{1}{SuperPoly(n)}.
\end{eqnarray}
Further, let $T^+$ and $T^-$ be defined as follows:
\begin{eqnarray}
&&T^+=\Big|\Big\{t\in\mathbb{N}^+\mid N_{t-1}< N_t,N_{t-1}\in
\mathbb{A}_2\cup\mathbb{A}_1,\sigma<\frac{\delta\log n}{n}\Big\}\Big|,\\
&&T^-=\Big|\Big\{t\in\mathbb{N}^+\mid N_{t-1}> N_t,N_{t-1}\in \mathbb{A}_2\cup\mathbb{A}_1,\sigma<\frac{\delta\log n}{n}\Big\}\Big|,
\end{eqnarray}
and the above definitions imply that $T=T^++T^-$.

Moreover, by Proposition B\ref{theorem:generalization_low_sigma}.4, given the condition that the EA is in $\mathbb{A}_2\cup\mathbb{A}_1$, in one generation the number of matching bits cannot increase by more than $U_2$ with an overwhelming probability. Hence, among the $T$ generations, the number of matching bits can increase by $T^+U_2-T^-$ at most with an overwhelming probability. To travel through $\mathbb{A}_1$, the following inequality must hold (a necessary condition):
\begin{eqnarray*}
T^+U_2-T^-\ge G.
\end{eqnarray*}
Noting that $T=T^++T^-$, it follows that
\begin{eqnarray*}
T^+\ge\frac{G+T}{U_2+1}.
\end{eqnarray*}
Recall the definitions of $G$, $T$ and $U_2$, we have
\begin{eqnarray*}
\frac{G+T}{U_2+1}>d\cdot\frac{T}{\gamma^{1/7}}
\end{eqnarray*}
where $d$ is a positive constant. Combining the above two inequalities together, we know that the following condition must be satisfied so as to travel through $\mathbb{A}_1$:
\begin{eqnarray}\label{equation:T_plus_lowerbound}
T^+>d\cdot\frac{T}{\gamma^{1/7}}.
\end{eqnarray}
Next, we only need to prove that the above condition cannot be satisfied with an overwhelming probability (thus the EA cannot travel through $\mathbb{A}_1$ with an overwhelming probability). It follows from Eqs. \ref{equation:efective_prob_increase} and \ref{equation:equation:eff_mutation} that
\begin{eqnarray*}
\mathbb{E}[T^+\mid T]=O\bigg(\frac{T}{\gamma^{3/7}}\bigg),
\end{eqnarray*}
where $O$ is referred to the asymptotic order of the problem size $n$. By Chernoff bounds, we estimate the probability of Eq. \ref{equation:T_plus_lowerbound}:
\begin{eqnarray*}
\mathbb{P}\bigg(T^+>d\cdot\frac{T}{\gamma^{1/7}}\mid T\bigg) <\bigg(\frac{e\gamma^{1/7}\mathbb{E}[T^+\mid T]}{dT}\bigg)^{dT/\gamma^{1/7}} =O\bigg(\frac{1}{\gamma^{2/7}}\bigg)^{\Theta(T/\gamma^{1/7})},
\end{eqnarray*}
where $d$ is a positive constant. Meanwhile, Eq. \ref{equation:T_lowerbound} implies that
\begin{eqnarray*}
\mathbb{P}\Big(T\ge\gamma^{3/7}\log n\Big)\succ 1-\frac{1}{SuperPoly(n)}.
\end{eqnarray*}
By the total probability theorem \cite{Papoulis84}, we obtain
\begin{eqnarray*}
\mathbb{P}\bigg(T^+>d\cdot\frac{T}{\gamma^{1/7}}\bigg) \prec \frac{1}{SuperPoly(n)}.
\end{eqnarray*}
In other words, the condition $T^+U_2-T^-\ge G$ does not hold with an overwhelming probability, which implies that the EA cannot travel through $\mathbb{A}_1$ and reach $\mathbb{F}_2$ with an overwhelming probability, given the condition that it has already reached $\mathbb{A}_2$. Consequently, we have proven that the EA cannot reach $\mathbb{F}_2$ by a polynomial first hitting time with an overwhelming probability. Formally, let $\tau_{\mathbb{F}_2}$ be defined as follows
\begin{eqnarray}
\tau_{\mathbb{F}_1}= \min\Big\{t\ge 0; (N^{(P)}_t\in \mathbb{F}_1)\vee(N^{(O)}_t\in \mathbb{F}_1)\Big\}.
\end{eqnarray}
We have proven that
\begin{eqnarray*}
\mathbb{P}(\tau_{\mathbb{F}_2}\prec Poly(n))\prec \frac{1}{SuperPoly(n)},
\end{eqnarray*}
which leads to Theorems \ref{theorem:generalization_low_sigma} and \ref{theorem:px_low_sigma} according to the definition of $\mathbb{F}_2$ in Definition \ref{definition:decomposition_I}. $\hfill\square$

\section*{Acknowledgements}
The authors would like to thank Dr. Yang Yu for his constructive comments to this paper. This work is partially supported by Natural Science Foundation of China grants (No. 61033009, No. 61028009, No. 61003064, and No. U0835002), National S\&T Major Project (Grant 2010ZX01036-001-002), and an Engineering and Physical Science Research Council grant in UK (No. EP/I010297/1).

%

\end{document}